\documentclass[twocolumn, superscriptaddress, floatfix, longbibliography, showkeys, aps, nofootinbib]{revtex4-1}  

\usepackage{graphicx} 
\usepackage{bm}	
\usepackage{color}                     
\usepackage{xcolor}
\usepackage{epsfig}
\usepackage{amsmath} 
\usepackage{amssymb} 
\usepackage{longtable} 
\usepackage{float}
\usepackage{mathtools}
\usepackage{xparse}
\usepackage{hyperref}
\usepackage{times}
\usepackage[normalem]{ulem}
\usepackage[ruled]{algorithm}
\usepackage{algpseudocode}
\usepackage{enumitem}
\usepackage[section]{placeins}
\usepackage{array}
\usepackage[caption=false]{subfig}
\usepackage{sidecap}
\sidecaptionvpos{figure}{t}

\newcommand{\figref}[1]{Figure {\ref{#1}}}
\newcommand{\secref}[1]{Section {\ref{#1}}}
\newcommand{\secsref}[1]{Sections {\ref{#1}}}
\newcommand{\tableref}[1]{Table {\ref{#1}}}
\newcommand{\appendixref}[1]{Appendix {\ref{#1}}}
\renewcommand{\eqref}[1]{Eq. ({\ref{#1}})}

\newcommand{\be}{\begin{equation}}
\newcommand{\ee}{\end{equation}}
\newcommand{\bd}{\begin{displaymath}}
\newcommand{\ed}{\end{displaymath}}
\newcommand{\BE}{\begin{eqnarray}}
\newcommand{\EE}{\end{eqnarray}}

\newcommand{\avg}[1]{\left\langle{#1}\right\rangle}

\definecolor{darkgreen}{rgb}{0.0, 0.5, 0.0}

\begin{document}

\setlist[enumerate,1]{label=\arabic*, start=0}

\title{Generalization Metrics for Practical Quantum Advantage in Generative Models}

\author{Kaitlin Gili}
\thanks{Both authors contributed equally to this work.}
\affiliation{University of Oxford, Oxford, OX1 2JD, United Kingdom}
\affiliation{Zapata Computing Canada Inc., 25 Adelaide St E, Toronto, ON, M5C 3A1, Canada}

\author{Marta Mauri}
\thanks{Both authors contributed equally to this work.}
\affiliation{Zapata Computing Canada Inc., 25 Adelaide St E, Toronto, ON, M5C 3A1, Canada}

\author{Alejandro Perdomo-Ortiz}
\email{alejandro@zapatacomputing.com}
\affiliation{Zapata Computing Canada Inc., 25 Adelaide St E, Toronto, ON, M5C 3A1, Canada}


\date{\today} 

\begin{abstract}
As the quantum computing community gravitates towards understanding the practical benefits of quantum computers, having a clear definition and evaluation scheme for assessing practical quantum advantage in the context of specific applications is paramount. Generative modeling, for example, is a widely accepted natural use case for quantum computers, and yet has lacked a concrete approach for quantifying success of quantum models over classical ones. 
In this work, we construct a simple and unambiguous approach to probe practical quantum advantage for generative modeling by measuring the algorithm's generalization performance. Using the sample-based approach proposed here, any generative model, from state-of-the-art classical generative models such as GANs to quantum models such as Quantum Circuit Born Machines, can be evaluated on the same ground on a concrete well-defined framework. In contrast to other sample-based metrics for probing practical generalization, we leverage constrained optimization problems (e.g., cardinality-constrained problems) and use these discrete datasets to define specific metrics capable of unambiguously measuring the quality of the samples and the model’s generalization capabilities for generating data beyond the training set but still within the valid solution space. Additionally, our metrics can diagnose trainability issues such as mode collapse and overfitting, as we illustrate when comparing GANs to quantum-inspired models built out of tensor networks. Our simulation results show that our quantum-inspired models have up to a $68 \times$ enhancement in generating unseen unique and valid samples compared to GANs, and a ratio of 61:2 for generating samples with better quality than those observed in the training set. We foresee these metrics as valuable tools for rigorously defining practical quantum advantage in the domain of generative modeling. 
\end{abstract}

\maketitle

\section*{Introduction}\label{s:intro}
Outstanding efforts have been made in recent decades in the search for quantum advantage, and reaching this milestone will have a profound impact on many areas of research and applications. Quantum advantage is generally intended as the capability of quantum computing devices to outperform classical computers, providing exponential speedups in solving a given task, which would otherwise be unsolvable, even using the best classical machine and algorithm~\cite{Google2019supremacy, wu2021strong, madsen2022quantum, Preskill2018, boixo2018characterizing, bouland2019complexity}. In recent years, a large part of the quantum computing community has been gravitating toward a more concrete definition of quantum advantage, namely \emph{practical quantum advantage} (PQA), also propelled by the growing interest from technology firms and companies in various application domains. Practical quantum advantage indicates the quest for quantum machines that can solve problems \emph{of practical interest} that are not tractable for traditional computers~\cite{daley2022practical,carena2021practical}. In other words, practical quantum advantage is the ability of a quantum system to perform a useful task faster or better than is possible with any existing classical system~\cite{alsing2022accelerating}. As long as the superiority is demonstrated in the real-world setting, under the real constrains and problem size of interest, one can waive the need for demonstrating an asymptotic scaling with problem size, which is the usual emphasis in algorithmic quantum speedup \cite{Ronnow25072014}. Our work focuses on further specifying and measuring practical quantum advantage in the context of generative models, which have been identified as promising candidates for harnessing the power of quantum computers~\cite{PerdomoOrtiz2017}. There have been several contributions that outline the potential benefits and limitations of using quantum generative models as alternative or enhancers to classical models~\cite{Coyle2019, kasture2022protocols, Vinci2020VAE, phillipson2020quantum, dunjko2018machine, liu2018differentiable, gao2017efficient, rudolph2020generation, hinsche2021learnability, sweke2020learnability, hinsche2022}. However, we still lack a unitary vision of what practical quantum advantage exactly means when it comes to generative models. 

We aim to provide such a vision and equip it with quantitative tools to evaluate progress toward its accomplishment. We suggest that generative models’ performance be assessed by their capability to \emph{generalize}, i.e., generate new high-scoring diverse solutions for the task of interest~\cite{nica2022evaluating, alcazar2021enhancing}. We highlight that our definition of generalization differs from the one outlined within the theoretical setting of computational learning theory~\cite{comp_learning_theory, sweke2020learnability, hinsche2021learnability, hinsche2022}, i.e., a model's ability to learn the ground truth probability distribution given a limited set of training data. Our approach follows closely the definitions and frameworks used by other ML practitioners (see e.g., \cite{zhao2018bias, nica2022evaluating}), which focus on scalable and practical methods to evaluate the performance of generative models. We believe that the two approaches complement each other in a practical context, and in \appendixref{app:supp_figs}, we provide a demonstration of the correlation between the approaches.

In this work, we present a unified framework to measure the generalization capabilities for both state-of-the-art classical and quantum generative models, and provide a first comparison between different models that highlights the superior performance of quantum-inspired methods over classical ones. We compare our practical vision of generalization to the computational learning theory standpoint (\secref{s:ev-approaches}) and to previously developed frameworks (\secref{app_bench}). In \secref{s:gen_define}, we propose our quantitative definition of generalization, while \secref{s:gen_task} illustrates our discrete-dataset-based framework to assess this capability. By leveraging discrete datasets relevant to many application domains~\cite{Ruiz-Torrubiano2010}, we can unequivocally measure the generalization capabilities of any generative models for practical tasks. In \secref{s:gen_metrics} we introduce robust sample-based metrics that allow one to conduct a comprehensive quantitative assessment of a model’s practical generalization capabilities and detect common pitfalls associated to the training process. Furthermore, in \secsref{exp_details} and \ref{s:results} we illustrate our approach by comparing models from two separate regimes, namely fully classical Generative Adversarial Networks (GANs) and quantum-inspired Tensor Network Born Machine (TNBM) architectures, for a specific task with relevance in financial asset management. 

To the best of our knowledge, this is the first proposal of an approach that combines a heuristic-based analysis with an application-based dataset to quantitatively evaluate generalization of unsupervised generative models and to directly compare classical and quantum-inspired models side by side in search for practical quantum advantage.

\section{Related Works}~\label{review}
Generative models are powerful and widespread algorithms, but the evaluation of their performance, especially on real-world datasets, is an open challenge. A huge variety of metrics and studies have been proposed to evaluate generative models, which can be found in two distinct sub-fields of machine learning (ML) research: computational learning theory~\cite{comp_learning_theory,briol2019statistical,power_gen_learning} and models' performance benchmarking~\cite{zhao2018bias, borji2019GANmetrics,alaa2021faithful, thanhtung2021generalization,nica2022evaluating}. First, we aim to give a brief overview of these two areas of research, and to draw a clear distinction between them, pointing to the advantages and limitations of each for evaluating unsupervised generative models. Subsequently, as this work predominantly contributes to the models' performance benchmarking sub-field, we focus on providing an overview of the main evaluation strategies that exist in this literature domain, pointing to Ref.~\cite{borji2019GANmetrics, borji2021pros} for a thorough review.

\subsection{Two Evaluation Approaches}\label{s:ev-approaches}
The language utilized in the sub-fields of computational learning theory and models' performance benchmarking varies greatly when discussing the evaluation approaches of unsupervised learning algorithms. There is a common goal of finding the best model (i.e., the one that `generalizes' best); however, the optimal criterion and the generalization definition differ in the two perspectives. 

In the context of computational learning theory, the optimal model is the one that has best approximately learned the ground truth probability distribution from the available training data~\cite{pac}. Thus, generalization coincides with good inference capability. Upon taking this to be the definition of generalization, the model is able to achieve high-quality performance if its output distribution post-training is sufficiently close to the (unknown) ground truth. By using the Probably Approximately Correct (PAC) approach~\cite{pac}, one can derive worst-case generalization error bounds for a very broad range of models. These insights are incredibly useful for identifying clear cases in which models will not provide value, especially in the search for circumstances where quantum algorithms might exhibit an advantage over classical ones~\cite{sweke2020learnability, hinsche2022, power_gen_learning}. On real-world datasets, this definition of generalization can be extended to evaluating the difference between the trained model distribution and the empirical approximation of the ground truth, using a quantitative distance metric of choice. 

However, we note that this is where the definition of generalization in the context of computational learning theory diverges from that of the models' performance benchmarking domain. For many practical problems, indeed, the optimal generative model is the one that can generate unseen high-quality data points that are solutions to a specific task, i.e., samples drawn from the ground truth distribution, but that did not exist in the empirical distribution used for training~\cite{alcazar2021enhancing,li2021quantum,nica2022evaluating}. This implies that the emphasis is on the model being able to produce samples that come from the unseen part of the ground truth distribution: this capability of generating novel, diverse and good solutions is what is defined as generalization in this practical context~\cite{nica2022evaluating}. Hence, if a model is provided with the complete set of solutions in the training process, it cannot generalize. Instead, since all the samples from the support of the ground truth distribution are given, the model would be restricted to exhibiting a behavior that we describe as \emph{memorization}, in even the best training scenario. In computational learning theory, this behavior would still be seen as a form of high-quality \emph{generalization} performance, as long as the model learned the right features of the distribution. This is usually a case of interest in density estimation tasks; however, in our practical context, this behavior is distinct from generalization such that it can be detected when it is not useful for specific real-world applications, where the generative model is trained with the purpose of generating novel samples from the ground truth distribution. 

In summary, the main difference between the two approaches is that in the models' performance benchmarking domain, the goal is to capture the model’s generalization performance as a novelsamples generator (\emph{``efficient generator''}), not as a ground truth learning algorithm(\emph{``efficient learner''}), as it is the case in computational learning theory. We highlight that an \emph{``efficient learner''} does not always imply an \emph{``efficient generator''} for a practical task at hand, and vice versa. The exact relation between the two approaches, especially its rigorous proof, is out of the scope of this paper (despite a first empirical demonstration in \tableref{table:valid_learning_theory}), but it is certainly an exciting avenue to bridge the gap between the two communities. We believe that the practical evaluation schemes, further described in \secref{app_bench}, can augment our understanding of models' performance by providing a detailed picture, based on evaluating specific desired features of generated data,  as well  as by highlighting their tendency to exhibit training failures. However, we recognize that this practical evaluation does not provide the same insights with regard to scaling complexity as those in computational learning theory. Therefore, we strongly emphasize that both research sub-fields are necessary to fully evaluate generative models, and that when possible, results from both realms should be included. For the purposes of PQA, we adopt and build on the more practical performance benchmarking approaches to generalization, which are meaningful enough to industrial real-world generative applications.

As we have seen the definition of generalization to take on slightly different meanings depending on the research domain, we now formally distinguish this practical generalization from the one defined in computational learning theory by providing the names \emph{validity-based} generalization and \emph{quality-based} generalization when defining our framework. 

\subsection{Models' Performance Benchmarking}\label{app_bench}
A common approach to evaluate generative models uses statistical divergences, such as the Kullback-Leibler divergence~\cite{Gili2022} and the Total Variation Distance~\cite{hinsche2022}. Unfortunately, the sample complexity of such quantities scales poorly with the dimensionality of the distribution under examination, proving them inadequate in high-dimensional spaces. To overcome this limitation, alternative evaluation metrics with polynomial sample complexity have been proposed, such as Inception Score (IS)~\cite{salimans2016improved}, Frech\'et Inception Distance (FID)~\cite{heusel2018gans}, and Kernel Inception Distance (KID)~\cite{binkowski2021demystifying}. Additional strategies include utilizing kernel methods such as measuring the Maximum Mean Discrepancy (MMD) ~\cite{briol2019statistical}, or neural networks to estimate statistical divergences ~\cite{gulrajani2020towards}.

The main limitation affecting divergence-based metrics lies in that a single number summary is used to score a model, thus being unable to distinguish its different modes of failure. In light of this consideration, Ref.~\cite{sajjadi2018assessing} introduced precision and recall as metrics to evaluate generative models, hence proposing a 2D evaluation to disentangle the various scenarios that can arise after training. Follow-up contributions have attempted to extend this idea from discrete to arbitrary probability distributions~\cite{simon2019revisiting}, and to improve precision and recall definitions and computation~\cite{naeem2020reliable,kynkaanniemi2019improved}.

This plethora of methods suggests how challenging it is to evaluate generative models. Evaluating the evaluation metrics themselves is an even more complicated task, despite the paramount importance of choosing the right metric for drawing the right conclusions~\cite{theis2016note}. Ref.~\cite{xu2018empirical} addresses such a problem, identifying a few necessary conditions that a metric should satisfy in order to qualify as a good performance estimator. One of these conditions is the ability of a metric to detect overfitting. As highlighted by Ref.~\cite{webster2019detecting}, overfitting is basically equivalent to memorization, i.e., \emph{anti-generalization}, and it is not always well defined, despite its importance. 

While being well established in the context of image classification, notions of generalization are less standardized for generative models. Initial studies on this topic in the context of generative models can be found in Refs.~\cite{meehan2020non, gulrajani2020towards}. Nonetheless, none of the available metrics is specifically tailored to assessing generalization capabilities, or, in other words, to detect overfitting upon occurrence~\cite{thanhtung2021generalization}. So far, very few contributions have been proposed to address the interesting problem of studying and quantifying generalization from a real-world application perspective for generative models. This knowledge gap becomes exceedingly evident when looking at the recent literature contributions to the field of quantum generative modeling. Several of these works have hinted at the concept of generalization, but have ultimately restricted their results to replicating a given target probability distribution~\cite{han2018unsupervised, bradley2020modeling, stokes2019probabilistic, miller2020tensor, alcazar2021enhancing}. Leaving such a question for future research indicates the difficulty in benchmarking both classical and quantum models on real-world datasets for their generalization capabilities. Our work aims at filling this gap: we propose a well-defined approach to practical generalization, deepening insights gathered from Ref.~\cite{zhao2018bias}, and adequate metrics to quantify such capability, following up on the authenticity metric proposed in Ref.~\cite{alaa2021faithful}.

Ref.~\cite{zhao2018bias} proposed a strategy to analyze generalization in generative models, which consists in probing the input-output behaviour of generative models by projecting data onto carefully chosen low-dimension feature spaces. By comparing the training and the generated distribution in these spaces, it is possible to assess whether a model can generate out-of-training samples. However, this contribution focuses only on spotting unseen (i.e., non-memorized) samples, without questioning whether these new samples are meaningful data for the task being solved, or useless noise. Ref.~\cite{xuan2019anomalous} hints at this limitation, referring to some of the results in Ref.~\cite{zhao2018bias} as \emph{anomalous generalization} behaviour, where the generated distribution differs significantly from the training distribution. The approach we propose in this work takes off from these two contributions. It goes deeper into the formal definition of generalization, identifying different regimes that allow us to assess if a generative model can generate samples that are new high-quality solutions to the problem at hand. Our approach is able to discriminate between anomalous generalization and generalization to valid and good samples. Inspired by the numerosity feature map proposed in Ref.~\cite{zhao2018bias}, we focus our work on discrete probability distributions. This choice allows us to avoid the introduction of complicated embeddings, which are instead required for most of the evaluation metrics proposed so far, and it is also more in line with our interest in extending the generalization study to quantum models in search for practical quantum advantage. 

In addition to defining the approach, we introduce several quantifiable measures of the practical generalization concepts we formalize. Ref.~\cite{alaa2021faithful}'s proposal of the \emph{authenticity} metric to identify data-copied samples paved the way for our generalization metrics. We share their starting point that precision and recall are independent of generalization capabilities, as the latter is not properly assessed by the former. Additionally, we share their point on the importance of the novelty feature of the samples generated by a model. The metrics we propose, though, go beyond the authenticity metric in that they aim at equipping the ``novelty space" with estimators that quantify important features, i.e., fidelity, rate and coverage of such an unseen space. The focus of our evaluation metrics revolves around the out-of-training generated samples, disregarding the known data.

To better contextualize our metrics with respect to previous works, we highlight that we share the starting point of Ref.~\cite{sajjadi2018assessing}. Hence, we propose multiple generalization metrics to disentangle different features and modes of failure. Additionally, our metrics satisfy the conditions expressed in Ref.~\cite{xu2018empirical}: they are able to detect overfitting and mode collapse. The generalization metrics proposed in this work aim at starting a new thread in comparing classical and quantum generative models on real-world applications, focused on assessing if they are able to generate new valid and valuable data. We see this approach as a necessary step forward in the models' performance benchmarking domain for demonstrating practical quantum advantage, not necessarily to be used in isolation to determine overall quality, but rather alongside other evaluation metrics and insights obtained from computational learning theory to provide a comprehensive assessment of these powerful data generators.

\section{Generalization}\label{s:gen_define}

Unsupervised generative models aim at capturing implicit correlations among unlabeled training data in order to generate samples with the same underlying features.
In this work, we focus on binary encodings of datasets with discrete values, and therefore, discrete probability distributions. This is needed to facilitate the comparison of quantum and classical generative models, and to allow for a more accurate and unambiguous evaluation of generalization as opposed to the continuous case, as further clarified in \secref{ss:quality-based}.

More concretely, given a dataset $\mathcal{D}_{\text{Train}} = \{{x_{1}, x_{2},...,x_{T}\}}$, where each sample $x_{t}$ is an $N$-dimensional binary vector such that $x_{t} \in \{0,1\}^N$ with $t=1,2,\dots,T$, we can train a generative model to resemble the unknown probability distribution $P(x)$ from which the samples in $\mathcal{D}_{\text{Train}}$ were drawn. We denote these samples as $\mathcal{D}_{\text{Gen}}=\{x_{1}, x_{2},..., x_{G}\}$, where each $x_{g}$ is again an $N$-dimensional binary bitstring, with $g=1,2,\dots,G$. As it will be shown later, the only requirement for the data distribution $P(x)$ is to have a support, which is a ``valid" sector, and a complement, which is a set of noise or undesirable features. Many real-world datasets can be represented this way: for example, portfolio optimization as demonstrated in our work, as well as molecular design problems~\cite{gao2021quantumclassical}. Remarkably, the notion of a constraint that defines valid and invalid spaces arises naturally within the context of combinatorial optimization as the constraint is usually part of the problem definition~\cite{Ruiz-Torrubiano2010, Lopez-Piqueres2022}. 

Since the goal of the present work is to compare the generalization performance of models for measuring practical quantum advantage, we introduce formal definitions and metrics in \secref{s:gen_metrics} to quantify different aspects of the practical behaviours that arise when we sample from the generative model. To further distinguish these definitions from those in computational learning theory, we provide contextual names: validity-based and quality-based generalization. Here, we provide a brief high level introduction of them, presenting the essential concepts for studying various flavours of generalization. 

\subsection{Pre-Generalization}
We refer to \emph{pre-generalization} as the generative model's ability to go beyond the training set $\mathcal{D}_{\text{Train}}$ by producing unseen outputs. More precisely, for any level of generalization to occur it is necessary - but not sufficient - that there exist some points $x_g$ such that
\begin{equation}\label{eq:pre-gen}
    {x}_g \in \mathcal{D}_{\text{Gen}} \wedge {x}_g \notin \mathcal{D}_{\text{Train}}.
\end{equation}
However, these outputs may not be samples distributed according to $P(x)$; for example, they may just be meaningless noise instead. In other words, pre-generalization is the model's ability to generate any new output - whether it is distributed according to $P(x)$ or not (\figref{fig:gen-def}). 
Note that we consider this behaviour to be a prerequisite for a model to be able to generalize, and not generalization in and of itself. As mentioned above and further specified below, to have any kind of generalization, a model must first be able to generate data beyond the training set, and the generalization potential is higher if the amount of unseen data is maximized. This implies that the training set  cannot be exhaustive, i.e. the number of unique\footnote{Bitstrings = \{00, 00, 11\}, unique bitstrings = \{00, 11\}.} training bitstrings  must be less than the number of unique bitstrings that can be sampled from $P(x)$. To discover new data, the training dataset should not consist of all of the bitstrings that could be sampled from the original distribution (i.e. its support). 

The pre-generalization behaviour can be verified with our exploration metric $E$, defined in \secref{sec:pre-gen}, that quantifies how many generated samples were not included in the training set. We note that this quantity has a similar definition to the \emph{authenticity} metric in Ref.~\cite{alaa2021faithful}, that captures sample novelty. However, our exploration metric is computed directly from samples rather than requiring an embedding scheme and a separate classification network. This quantity allows one to investigate the general questions: \emph{“Can the model reach out-of-training data points? And with which frequency?”}.

\subsection{Validity-Based Generalization}\label{sec:const-gen}
We refer to \emph{validity-based generalization} as the generative model's ability to go beyond the training set $\mathcal{D}_{\text{Train}}$ and effectively produce new bitstrings living in a given solution space with the underlying distribution $P(x)$ (\figref{fig:gen-def}). In other words, the model is able to learn a fixed particular feature about bitstrings drawn from $P(x)$ and produce new samples with the same feature, where this feature is specified via a constraint on the bitstrings. More precisely, the generative model outputs samples $x_{g}$ such that 
\begin{equation}
x_g \notin \mathcal{D}_{\text{Train}} \wedge x_g \in \text{support of } P(x).
\label{eq:constr-gen}
\end{equation}

We remark here that this approach for validity-based generalization is task-independent, as the metrics are exclusively sample-based and agnostic to the specific use case, or more specifically, independent of the quality associated to each bitstring. In \secref{s:gen_task} we highlight the essential conditions one needs to meet when defining an appropriate task to study validity-based generalization.

We evaluate the validity-based generalization behaviour introducing the three metrics of \emph{fidelity} $F$, \emph{rate} $R$, and \emph{coverage} $C$. In a nutshell, $F$ quantifies the probability that a model generates unseen samples that are valid results rather than unwanted noise. $R$ quantifies the frequency at which a model produces unseen and valid results. $C$ quantifies the fraction of unseen and valid results retrieved among all the potential valid and unseen samples.
These metrics allow one to answer the following general questions, respectively linked to the three generalization estimators presented above:
\begin{itemize}
\item{$F$: “How effectively can the model distinguish between noisy and valid unseen results?”}
\item{$R$: “How efficiently can the model reach unseen and valid results?”}
\item{$C$: “How effectively can the model reach all unseen and valid results?”}
\end{itemize}

\subsection{Quality-Based Generalization}\label{ss:quality-based}
We refer to \emph{quality-based generalization} as the generative model's ability to go beyond the training set $\mathcal{D}_{\text{Train}}$ and effectively produce bitstrings living in a given solution space with underlying distribution $P(x)$, where the new bitstrings can be mapped to a real number indicating their quality. While there can be many examples of functional maps that one could use to assign each bitstring a score to be maximized, we emphasize optimization as a natural choice for assigning such a value to each sample, as proposed by Refs.~\cite{alcazar2021enhancing, Bengio2021, nica2022evaluating}. In this case, the score is quantified by a cost to be minimized. In other words, optimization provides a natural framework to introduce quantitative estimators of generalization, as a generative task can be equipped with a well-defined cost function, indicating the quality of samples.  The framework presented here combines generalization and optimization as a promising strategy towards the definition of quantitative metrics. We highlight that if one uses a generative model as an optimizer, the success of the algorithm depends on the generation of high-quality solution candidates, rather than inferring the ground truth data distribution as it is the case in computational learning theory.

\begin{figure}[!hbtp]
\includegraphics[width=\linewidth, scale=1]{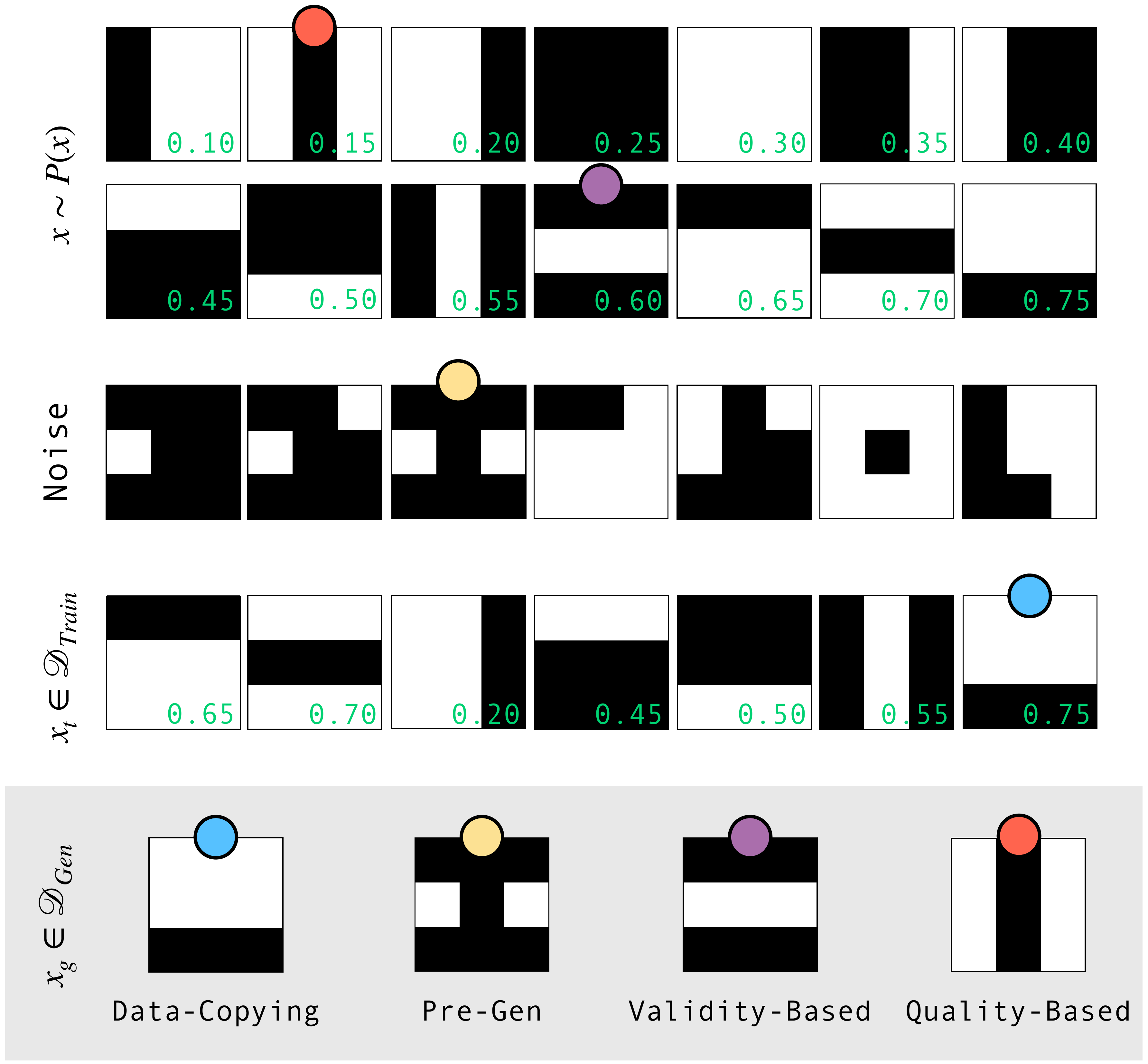}
\caption{\textbf{A visual representation of generalization-related concepts.} The figure shows the different behaviours a model can exhibit when generating data, using a 3x3 Bars and Stripes dataset as an example. The top two rows display a set of samples $x$ distributed according to the data distribution $P(x)$; note that only a subset of the 3x3 Bars and Stripes dataset is displayed, rather than the full set of patterns. The third row contains samples that do not belong to this dataset (Noise). The fourth row contains a subset of samples $x_t \in \mathcal{D}_{\text{Train}}$ used for training and distributed according to $P_{\text{Train}}(x_t)$, while the bottom row shows a new set of samples $x_g$ produced by the model and living in $\mathcal{D}_{\text{Gen}}$.
Note that each sample contains an associated toy-score that corresponds to the samples' associated cost. In this toy example, the samples are assigned a real-valued score in $(0, 1)$, except for noisy samples that don't have an associated cost as they are not part of the valid solution space. The bottom row displays four samples from the generated queries, each of which is tagged with a different model behaviour: memorizing data from $\mathcal{D}_{\text{Train}}$ (blue dot), producing data outside of $\mathcal{D}_{\text{Train}}$ that may be noise (yellow dot), generalizing to new data distributed according to $P(x)$ (purple dot), and generalizing to new data distributed according to $P(x)$ that contains a minimum value to an associated cost function (red dot).}
\label{fig:gen-def}
\end{figure}

When focusing on quality-based generalization, one is interested in generating samples that satisfy a validity criterion, but also have associated costs that minimize a given objective function (\figref{fig:gen-def}). When considering continuous data distributions (e.g., in image generation tasks), assessing the quality of samples is particularly challenging, as embedding and non trivial transformations are needed in order to utilize the available metrics~(see, e.g., Refs.~\cite{zhao2018bias, alaa2021faithful}). Hence, on purpose we limit the scope of this work to discrete datasets, since this setting provides a more accurate and unambiguous evaluation of the generalization capabilities.

A generative model thus exhibits quality-based generalization if it is able to produce at least some unseen and valid samples that have on average similarly low (or lower) cost values than the ones associated to at least some of the training samples. More precisely,
\begin{equation}
x_{g} \text{  satisfies  \eqref{eq:constr-gen}} \wedge f(\mathcal{D}_{\text{Gen}}, c(x)) < f(\mathcal{D}_{\text{Train} }, c(x)), 
\label{eq:vb-gen}
\end{equation}
for a given suitable function $f$ (e.g., the minimum sample cost $c(x)$ in each sample set) that depends on how strict the cost minimization requirements are for the problem under examination (see \secref{s:vb-gen}).

Developing metrics for assessing quality-based generalization is a task-dependent challenge as it allows one to evaluate the model’s \emph{sample quality}, according to a specific task and measured by its associated cost function. 

In \secref{s:vb-gen}, we introduce two versions of the sample quality metric, induced by a different choice of $f$: the first one evaluates the model's ability to generate a minimum cost value that is lower than anything in the training set, whereas the second accounts for a diversity of samples whose cost is below a user-defined percentile threshold. Even though the former could seem more adequate to quantify the generator’s ability to go beyond the sample quality available in the training set, it may be the case that producing the lowest cost value is not the only desired behaviour of the task. For instance, it may be that the desired behaviour is to generate diversity of new samples with a cost comparable to the lowest values found in the training set. In this scenario, the latter version allows one to reward alternative solutions without restricting the model only toward values below the training threshold.
Since for many practical optimization tasks one cares about reaching a diverse pool of high-quality solutions, we also see value in considering the number of unique samples with a lower cost value than a user-defined threshold in the training set (e.g., the minimum value in the training set).

The quality-based generalization metrics allow us to investigate the general question: \emph{“Can the model reach unseen and valid results that are more or just as valuable than the best in the training set?”}.

\section{Generalization Task Definition}\label{s:gen_task}
In order to properly assess generalization from the practical perspective, the generative model's task must meet some essential requirements. Such assumptions do not limit the scope of our approach as they simply provide a robust definition of the task at hand.

As previously specified, we focus our analysis on binary encodings of discrete datasets $\mathcal{D}_{\text{Train}} = \{{x_{1}, x_{2},...,x_{T}\}}$, with $x_{t} \in \{0,1\}^N$. We can thus identify a search space $\mathcal{U}$ of size $2^N$, that contains all possible $N$-dimensional bitstrings. For validity-based generalization, there must exist a subspace of $\mathcal{U}$ containing the set of bitstrings we would like our trained model to generate. We refer to this as the valid solution space $\mathcal{S}$, that includes all the samples that exhibit a given desired feature. Hence, the model aims to approximate the underlying unknown \emph{data distribution}, defined as: 
\begin{equation}\label{data_prob}
P(x) = \frac {1} {|\mathcal{S}|}, \forall x \in \mathcal{S}. 
\end{equation}

We highlight that the notion of validity produces a non-trivial distribution of valid samples across the overall search space $\mathcal{U}$, adding complexity to the problem despite the data distribution being uniform over the solution space $\mathcal{S}$. 
We emphasize that this general solution space $\mathcal{S}$ will contain different bitstrings for various representational datasets of interest. For instance, \figref{fig:gen-def} displays samples from the well-known Bars and Stripes dataset~\cite{MacKay-book-2002}: in this case, the solution space $\mathcal{S}$ would contain all valid bar and stripe patterns, some of which are shown in the top row of the figure. Alternative datasets could focus on solution spaces defined by a parity constraint, by a cardinality constraint or by any other property of interest. We highlight that the solution space must have a well defined notion of validity that can be evaluated for each of the bitstrings in $\mathcal{U}$ to verify whether or not they live in its subset $\mathcal{S}$. 

The model’s task is therefore to generate novel samples in $\mathcal{S}$, after a learning process involving a limited number $T$ of unique training samples, i.e., $T = \epsilon|\mathcal{S}|$, where the seen portion $\epsilon \ll 1$ is a small parameter quantifying the percentage of $\mathcal{S}$ that gets seen during training. Note that this is a necessary requirement for generalization because it guarantees that the training set is not exhaustive. 

With $T$ training samples, the model has access only to an approximated version of the data distribution, that we denote as the \emph{training distribution}: 
\begin{equation}\label{train_prob}
P_{\text{Train}}(x) = \frac {1} {T}, \forall x \in \mathcal{D}_{\text{Train}}. 
\end{equation}

For quality-based generalization, there is an additional requirement as this behaviour depends not only on the  validity of the bitstrings, but also on the value associated to each pattern, according to a cost function $c(x)$. As such, in order to assess quality-based generalization, it is necessary for the task of interest to have a well-defined objective function that indicates the cost of each bitstring, in search for minimum values.

As we would like for our model to learn the valid bitstring patterns as well as to generate patterns with low-cost values, it is integral to re-weight the dataset distribution in \eqref{data_prob}. Here we use a \emph{softmax} function in order to introduce cost-related information in the training data set. In this scenario, the training samples approximate the following \emph{re-weighted training distribution}: 
\begin{equation}\label{biased_prob}
P^{\text{(w)}}_{\text{Train}}(x)= \dfrac{e^{-\beta_m c(x)}}{\sum_{i = 1}^{T} e^{-\beta_mc(x)}}, \forall x \in \mathcal{D}_{\text{Train}}.
\end{equation}

Following Ref.~\cite{alcazar2021enhancing}, $\frac{1}{\beta_m}$ was chosen to be the standard deviation of the costs in the training data, whereas $c(x)$ is the cost of each sample bitstring.

In summary, the two main essentials for evaluating respectively validity-based and quality-based generalization are the following: 

\begin{itemize}
\item There exists a well-defined solution space $\mathcal{S}$, containing bitstring patterns that are valid according to easy to specify and verify constraints.
\item There exists a well-defined cost function $c(x)$ that can be computed to assess the generalization for all valid bitstring patterns.
\end{itemize}

\section{Metrics for Evaluating Practical Generalization}\label{s:gen_metrics}

\begin{figure}[htp]
\includegraphics[width=\linewidth, scale=1]{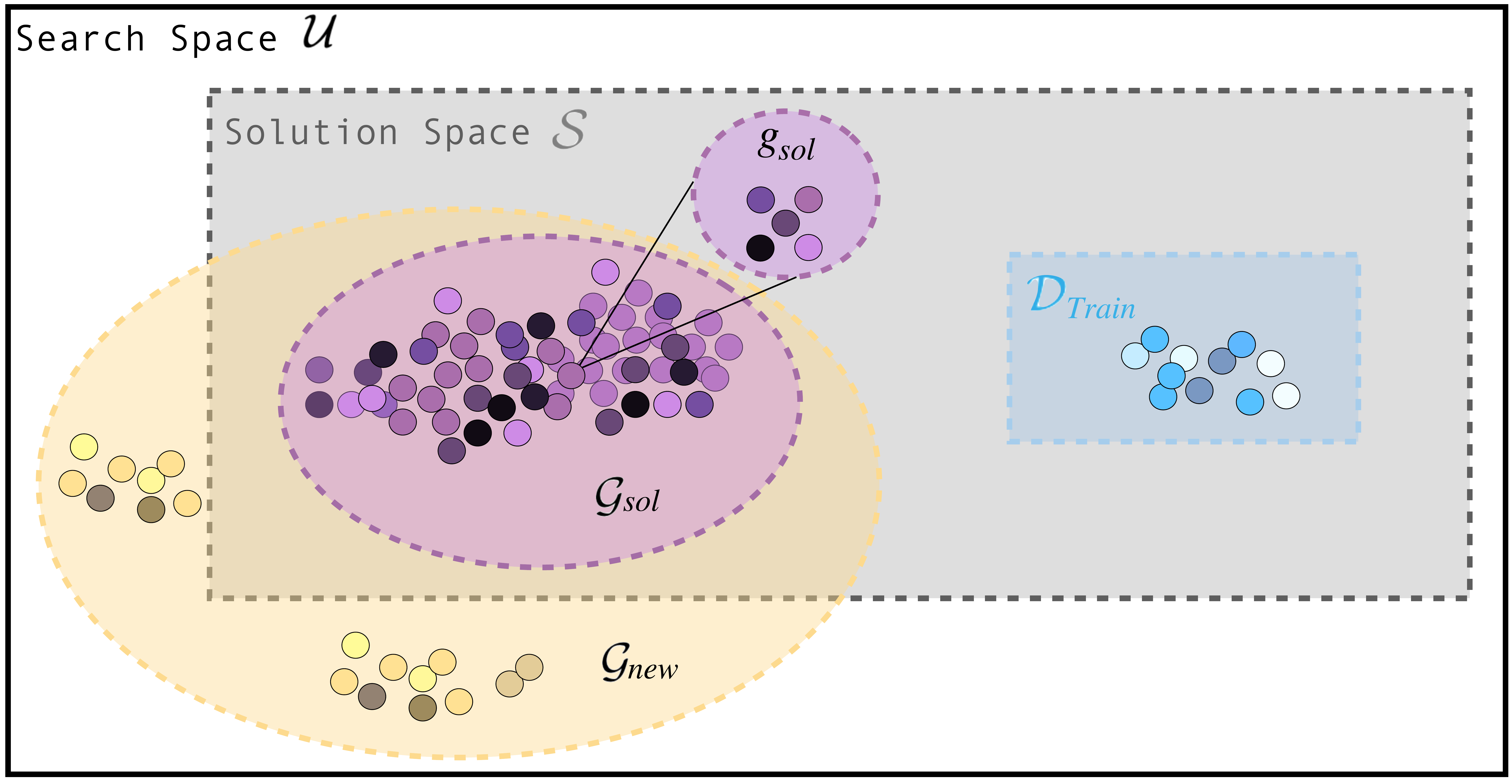}
\caption{\textbf{A visual representation of all possible spaces where a generated query might be located.} Each query is represented by a color-coded dot, where the color-code is the same as in \figref{fig:gen-def} (Data-Copying: blue, Pre-Generalization: yellow, Validity-Based Generalization: purple) and the color-shade represents a unique bitstring sample. We take all non-unique queries outside of the training set to be in the multi-subset $\mathcal{G}_{\text{new}}$ (inside the yellow oval), whether they are in the solution space $\mathcal{S}$ or not. Furthermore, we take $\mathcal{G}_{\text{sol}}$ to be all non-unique queries that exist in the solution space (inside the pink oval) and $g_{\text{sol}}$ to be all of the unique queries among $\mathcal{G}_{\text{sol}}$ (zoomed-in). Lastly, if a query exists in $\mathcal{D}_{\text{Train}}$, it is a memorized count from the training set. We note that the quality-based queries (not shown) must exist inside of the solution space.}
\label{fig:sets-def}
\end{figure}

As described in \secref{s:gen_define} and \secref{s:gen_task}, practical generalization occurs when a model generates novel samples that display desired features and belong to the support of some underlying distribution. To give a quantitative definition of the validity- and quality-based generalization metrics, we first need to clarify the nomenclature of all the spaces involved. We have already defined the collection of all queries generated by a trained generative model as $\mathcal{D}_{\text{Gen}}$, where $|\mathcal{D}_{\text{Gen}}| = Q$.
We then call {$\mathcal{G}_{\text{sol}}$ the multi-set of all valid and unseen queries, which reflect the model’s validity-based generalization capability. We further define a subset of $\mathcal{G}_{\text{sol}}$ that contains all its unique bitstring solutions as $g_{\text{sol}}$, thus the only difference between $\mathcal{G}_{\text{sol}}$ and $g_{\text{sol}}$ is that in the latter each bitstring appears only once, whereas in the former there can be many occurrences of the same sample. Lastly, we define the multi-subset of unseen queries as $\mathcal{G}_{\text{new}}$, where some of these queries might be unwanted noise and hence reflect the model’s exploration capability. Note that we use uppercase variables for multi-sets and lowercase variables for unique sets, and a visual representation of the sets in play can be found in \figref{fig:sets-def}.}

Having clarified the nomenclature of the spaces involved in the task, we can now proceed to the definition of the generalization metrics.

\subsection{Evaluating Pre-Generalization}\label{sec:pre-gen}
While a model’s capability to generate unseen samples that are not valid or valuable solutions to the task at hand is not considered generalization behaviour in and of itself, it is an important prerequisite for generalization from the practical perspective. If the model is not able to go beyond the training set, even just to produce noisy outputs, then the model is not passing the first requirement for generalization - the ability to produce novel data points. To conduct a pre-generalization evaluation prior to assessing for any kind of validity-based or quality-based generalization, we introduce the \emph{exploration} metric $E$, that quantifies the fraction of generated queries that are new data points, namely: 
\begin{equation}\label{exploration}
E = \frac {|\mathcal{G}_{\text{new}}|} {Q}.
\end{equation}

If $E \approx 0$, the model will not pass the first required check for practical generalization. This may be due to an intrinsic property of the model, i.e., the inability to generate novel data, or it can be an artifact of the training set being (almost) exhaustive, because nothing new can be generated if the training data covers (almost) all the entire valid space. 

\subsection{Evaluating Validity-Based Generalization}\label{sec:constrained-gen}

We introduce three sample-based metrics that describe each model’s validity-based generalization behaviour after training: fidelity $F$, rate $R$, and coverage $C$.

Fidelity describes the model’s ability to distinguish an unseen and valid sample in $\mathcal{S}$ from a meaningless output (i.e., noise) and it quantifies the fraction of unseen queries that fall into the unseen solution space. It is defined as follows:
\begin{equation}\label{fidelity}
F = \frac {|\mathcal{G}_{\text{sol}}|}{|\mathcal{G}_{\text{new}}|}.
\end{equation}

Rate describes the model’s ability to efficiently produce unseen and valid samples and it quantifies the fraction of all queries that fall into the unseen solution space, namely:
\begin{equation}\label{rate}
R = \frac {|\mathcal{G}_{\text{sol}}|} {Q} .
\end{equation}

Coverage describes the model’s ability to recover all unique unseen and valid samples and it quantifies how much of the solution space that was unexplored gets covered by the generative model's queries. It is defined as follows, where we highlight that the ratio does not take into account the queries' frequencies, as a single occurrence has the same weight as a one that appears multiple times:
\begin{equation}\label{coverage}
C = \frac {|g_{\text{sol}}|} {|\mathcal{S}| - T} .
\end{equation}
We highlight that one should expect the value of these metrics to depend on the number of queries $Q$ that are retrieved from the trained model. For example, to have a quality coverage of a space, i.e., $C \rightarrow 1$, one should have enough samples that fall in the entire unexplored space. However, this dependency does not constitute a limitation for drawing a comparison between models, as we can fix the number of queries for all the models under investigation, and evaluate and fairly compare their generalization performance at the given number of queries. Moreover, in \secref{sec:metric_robustness}, we further showcase the values of $C$ as we increase the number of queries toward and beyond the size of the solution space. We see a clear trend towards the metric ideal limit $C \rightarrow 1$ as we increase the number of queries. Conversely, in \appendixref{app:metric_trends} we demonstrate that fidelity and rate are not dependent on the number of generated samples, despite being sample-based metrics. 

We note that the different metrics are not completely independent, as there are mutual relations between them. For instance, it can be noted that rate and fidelity are correlated, as $R = EF$. Rate is the same as fidelity whenever a model generates exclusively unseen queries, which only holds in the case of perfect generalization (or in pathological cases such as mode collapse to unseen and valid queries). Another example of mutual relation between the metrics is that $C \leq \frac{EQ}{|\mathcal{S}-T|}$, which implies that $C<E$ for large solution spaces and limited queries budget.

To further clarify the expected metrics' values for a well-generalizing model, we highlight that these metrics will be exactly 1 when evaluated for a model that exhibits the highest validity-based generalization. However, in a practical sense, this might be difficult to achieve; we are then equipped with a theoretical upper bound of 1 for all metrics, with the understanding that one should aim to reach this limit to obtain a robust model for generalization.


Lastly, we note that the pre-generalization condition in \eqref{eq:pre-gen} impacts the validity metrics; hence, exploration $E$ is directly related to $(F, R, C)$. For $F$, the pre-generalization condition in \eqref{eq:pre-gen} must be met in order for the metric to be well-defined. When the condition is not met, $F$ will be null, and $C, R = 0$. Therefore, our metrics rely on the model's ability to go beyond the training set, and will indicate if the model is only data-copying. Other properties from the model can be inferred from these metrics as demonstrated in \tableref{table:cheat-sheet} in \appendixref{app:cheat_sheet}. For example, a metric which measures the degree of data-copying could be defined as $D = 1 - E$, hence perfect memorization would mean $E = 0$. We highlight that, in this framework, one can additionally use our proposed metrics to detect alternative and complementary behaviours to generalization and define additional metrics that are tailored towards specific properties one would like to investigate.

In conclusion, we propose to utilize the metrics $(F, R, C)$ to introduce a 3D quantitative investigation of the generalization capabilities mentioned in \secref{sec:const-gen}, that we report here for convenience:
\begin{itemize}
    \item Fidelity, $F$, evaluates how effectively the model can distinguish between unseen valid and invalid bitstrings.
    \item Rate, $R$, evaluates how efficiently the model can produce unseen and valid bitstrings.
    \item Coverage, $C$, evaluates how effectively the model can retrieve all unseen and valid patterns.
\end{itemize}

\subsection{Evaluating Quality-Based Generalization}\label{s:vb-gen}
To quantify the quality-based generalization properties of a generative model, we propose adequate metrics addressing the \emph{sample quality} of the generated samples, which speaks to how many of the queries are more valuable results in the context of a specific application domain, i.e., how many bitstrings have a low enough associated cost. Since the quality of a result depends on a given cost function, this metric is task-specific, as opposed to the validity-based generalization case that only requires the notion of validity of a query, according to a well-defined hard constraint.

More precisely, we introduce different nuances of this \emph{sample quality} metric for our quality-based generalization assessment, proposing two different versions with slightly different implementations of $f$ in the right-hand side condition of \eqref{eq:vb-gen}.

Firstly, we consider the Minimum Value ($MV$) of the costs associated to the queries generated by the model as a relevant evaluation metric, since in many optimization applications the main goal is to find the solution that minimizes the cost, or equivalently, the sample with the best quality. This corresponds to choosing $f =\min$, so that the condition of \eqref{eq:vb-gen} becomes:
\begin{equation}\label{lowest_risk}
x_{g} \text{  satisfies  \eqref{eq:constr-gen}} \wedge
\min\limits_{x_g \in \mathcal{D}_{\text{Gen}}} c(x_g) < \min\limits_{x_t \in \mathcal{D}_{\text{Train}}} c(x_t).
\end{equation}
Despite its practical impact, this punctual metric can be highly unstable if it is not supported by enough statistics as the metric relies on generating one specific value, the lowest. Since generating the query with the lowest cost is highly dependent on the selected batch $b$ of queries, we define this metric as an average across $B$ batches of queries to avoid biasing the results due to an anomalous batch. In other words, for each generative model evaluated, we define: $$MV = \frac{1}{B}\sum_{b = 1}^{B}\min\limits_{x_g \in \mathcal{G}_{\text{sol}}^b} c(x_g).$$ For the results presented in this work, we fixed $B=5$. Including such average in the definition of the $MV$ metric itself contributes to alleviate its intrinsic instability, thus making it more robust for quality-based generalization evaluation.

Secondly, we define the Utility $U$ as the average cost of a user-defined set $P_{t}$ of unseen and valid samples from the generative model. Specifically, $P_{t} (\mathcal{D})$ is the set obtained from taking the $t\%$ of samples with the best quality (lowest costs) in $\mathcal{D}$. Setting $t=5$, this corresponds to choosing $f=\avg{\cdot}$ on the set ${P_5}$, and the condition of \eqref{eq:vb-gen} reads:

\begin{equation}\label{utility}
\begin{split}
&\quad\quad\quad\quad x_{g} \text{  satisfies  \eqref{eq:constr-gen}  } \wedge \\[1.1ex]
&\avg{c(x_g)}_{x_g \in P_5(\mathcal{G}_{\text{sol}})} < \avg{c(x_t)}_{x_t \in P_5(\mathcal{D}_{\text{Train}})}.
\end{split}
\end{equation}

Given its set-based definition, this metric is much more stable than the previous one. 

Lastly, we note that it is possible to give another definition of \emph{sample quality}, which simply consists in counting the number of unseen and valid queries whose cost is lower than a specific critical cost value $c'(x)$ in the training set. For example, one could take $c'(x)$ to be the lowest cost value in the training set i.e., $c'(x_t)=\min\limits_{x_t \in \mathcal{D}_{\text{Train}}} c(x_t)$. When utilizing this estimator, one is interested in verifying the following condition:
\begin{equation}\label{harsh_utility}
\Big|{\{x_g \text{ s.t. } c(x_g) < c'(x_t)\}}\Big|>0 \text{, for } x_t \in \mathcal{D}_{\text{Train}},
\end{equation}
where clearly a higher value of the left-hand side implies a better sample quality. Even though this quantity can carry interesting information, we don't include it among our quality-based generalization metrics as it is a harsh restriction to impose and may only be important for optimization tasks that are looking for many potential $MV$ bitstrings. We highlight that our framework is not limited to the metrics proposed so far, but allows one to define several other figures of merit which can be relevant for specific applications at hand.

We use these metrics to introduce insights into a model’s quality-based generalization capabilities, and determine which models are able to generate the most value for task-specific challenges. We emphasize again that this approach can be utilized beyond cost minimization problems, as long as there is a quantitative quality scale associated to each bitstring in the valid subspace.

\section{Approach demonstration}\label{exp_details}
To present the robustness of our approach in evaluating and comparing generative models, we choose a well-defined task and two families of models: classical Generative Adversarial Networks (GANs) and quantum-inspired Tensor Network Born Machines (TNBMs). The following sections outline the specific use case (\secref{sec:Practical Task}) and the generative models (\secref{sec:models}) selected for our experimental demonstrations.

\subsection{Use Case}\label{sec:Practical Task}
To demonstrate a practical application of our approach, we choose an important use case in the finance sector that addresses the challenge of cardinality-constrained portfolio optimization.
The goal of such task is to minimize the risk $\sigma$ associated to a collection of assets, randomly selected from the S\&P500 market index, for a fixed desired return $\rho$. Below, we highlight how this task is amenable to the  framework and requirements described in \secref{s:gen_task}.

Given a fixed size $N$ of the asset universe, a portfolio candidate can be encoded into a bitstring of length $N$, where each bit corresponds to an asset either being selected in the portfolio (1) or left out of the portfolio (0). Therefore, the search space $\mathcal{U}$ of all possible portfolios grows exponentially with the asset universe size, i.e. $|\mathcal{U}|=2^N$.  

To assess validity-based generalization within this task, we define the solution space $\mathcal{S}$ to be comprised of all bitstrings containing a fixed number $k=N/2$ of selected assets, i.e., a candidate solution must be a bitstring with a fixed Hamming weight equal to $k$.

With such $k$-cardinality constraint, the problem solution set $\mathcal{S}$ contains all possible portfolio bitstrings $x$ that fit this constraint. Thus, its cardinality is:

\begin{equation}\label{binom}
|\mathcal{S}| = \binom{N}{k}.
\end{equation}

To further assess quality-based generalization, we define an objective function that encodes the quality of each bitstring, namely the financial risk $\sigma$ associated to each portfolio, which in the case of the Mean-Variance Markowitz model~\cite{Markowitz52} can be efficiently computed by means of Mixed Integer Quadratic Programming (MIQP)~\cite{Alcazar2020ClvsQuant}. Unlike when investigating validity-based generalization, we use $\sigma$ to re-weight the training dataset with the softmax function described in \eqref{biased_prob}.

As such, this task satisfies both the previously introduced conditions necessary to evaluate validity-based and quality-based generalization. We again emphasize that our framework can be applied to any task that meets the essential requirements in \secref{s:gen_task}, and is not limited to this financial application. 

\subsection{Generative Models}\label{sec:models}
We focus our investigation on Generative Adversarial Networks (GANs) and Tensor Network Born Machines (TNBMs). This choice is motivated by several reasons. On the one hand, GANs constitute one of the most popular and top utilized classical generative models, notwithstanding the challenges that plague their training such as mode collapse~\cite{che2016mode}, convergence issues~\cite{roth2017stabilizing}, and vanishing gradients~\cite{arjovsky2017towards}. Moreover, they are made up of several components that can be independently and successfully promoted to a quantum model~\cite{rudolph2020generation}, thus paving the way to the study of hybrid quantum-classical generative models. On the other hand, recent results for training TNBM architectures show that such models are promising candidates to exhibit both validity-based and quality-based generalization behaviours~\cite{alcazar2021enhancing}. 
We started our generalization study choosing these two models, but our approach can be leveraged to characterize any other state-of-the-art generative model of interest, and we do hope other interesting works will spin out from this initial proposal to evaluate quantitatively their generalization power. Future work can include an analysis of fully quantum models, even trained on hardware, once current limitations in training large and deep circuits are overcome. 
\subsubsection{Generative Adversarial Network (GAN)}\label{sec:GAN}
Our classical model consists of a Generative Adversarial Network (GAN) architecture with a normal prior distribution, and we conduct the training as typically described in the literature~\cite{gui2020review, ruthotto2021introduction, goodfellow2014generative}. GANs are trained as two neural networks, a discriminator $D$ and a generator $G$, competing against one another for optimal performance in an adversarial game. Samples from a prior distribution $q(z)$ are fed into the generator’s  input layer, and throughout training the generator attempts to produce new data $x$ that can fool the discriminator into classifying $x$ as a real rather than an artificially created data point. The goal of training is to maximize the generator’s score and minimize the discriminator’s score as described by the loss function:

\begin{equation}\label{gan_cost}
\mathcal{L}_{\text{GAN}} = \min_{G} \max_{D}  [\mathbf{E}_{x \sim P_{\text{Train}}}(x)[\log D(x)] \\
\end{equation}
 \[+ \mathbf{E}_{z \sim q(z)}[\log (1 - D(G(z)))]].\]

For both the generator and the discriminator, we utilize a feed forward architecture with fully connected linear layers (details are listed in \tableref{table:GAN-params} in \appendixref{app:training-details}).

\subsubsection{Tensor Network Born Machine (TNBM)}\label{sec:TNBM}
Our quantum-inspired generative model is a Tensor Network Born Machine (TNBM), whose underlying architecture is chosen to be a Matrix Product State (MPS), a well-known 1D tensor network characterized by a low level of entanglement~\cite{han2018unsupervised}. A TNBM takes unlabelled $N$-dimensional training bitstrings from the dataset $\{x_t\}^{T}_{t = 1}$, and aims to encode the underlying probability distribution in a quantum wavefunction $\psi$, expressing the correlations between samples in the amplitude of a quantum state, namely: 

\begin{equation}\label{product_state}
|\psi\rangle = \sum_{\{s\}}\sum_{\{\alpha\}} A_{\alpha_{1}}^{s_{1}}A_{\alpha_{1} \alpha_{2}}^{s_{2}} ... A_{\alpha_{N}}^{s_{N}}|s_{1}s_{2}... s_{N}\rangle.
\end{equation} 

To motivate this representation, we note that an $N$-dimensional bitstring can be interpreted as a possible realization of the spin state (0,1) of $N$ particles $|s_{1}s_{2}... s_{N}\rangle$, and therefore the full quantum state can be written as a superposition of all the possible spin states. Rather than using the exact coefficient matrix to build $|\psi\rangle$, we approximate it by the product of smaller parametrized single-particle matrices $A^{s_i}$, where the dimensions $\{\alpha\}$ are known as bond dimensions. The summation across $\alpha$ determines the probability amplitude for each superposition state of individual sites; thus, the bond dimensions controls the expressivity of the TNBM.

We use a similar training method as described in Ref.~\cite{han2018unsupervised}, where models are trained via a DMRG-like algorithm with the log-likelihood cost function: 
\begin{equation}\label{TNBM_cost}
\mathcal{L}(\theta) = - \frac{1}{T} \sum_{t} \log(p_{\theta}(x_t)).
\end{equation}

During training, samples are generated from the wavefunction according to the Born Rule:
\begin{equation}\label{born_rule}
p_{\theta}(x_{t}) = |\langle x_{t} | \psi \rangle|^2, 
\end{equation}
and the goal of the learning process is to find an optimal TNBM parametrization $\theta$ such that $p_{\theta}(x_{t}) \rightarrow P_{\text{Train}}(x_t)$.

A TNBM is known as a quantum-inspired technique as it builds upon fundamental concepts and formalism of the quantum-mechanical theory, but it is executed entirely on a classical platform.

\section{Results and Discussion}~\label{s:results}
Having defined several quantitative metrics that allow one to conduct a generalization analysis of generative models from a practical perspective, we use them to investigate the performance of TNBM and GAN architectures. We present the results of our simulations, whose details are specified in \secref{sec:sim-det}. We demonstrate the robustness of our proposed metrics (\secref{sec:metric_robustness}), show their ability to spot common pitfalls in model training (\secref{sec:pitfalls}), and introduce insights into the validity-based and quality-based generalization capabilities of each model (\secref{sec:model_evaluation}). 

\subsection{Simulation Details}\label{sec:sim-det}
For our experiments, we consider a specific instance of a cardinality constrained portfolio optimization task, where we aim at minimizing the associated risk $\sigma$ for a given target return $\rho=0.002$, such that the asset universe from which one can pick to build a new candidate portfolio has size $N=20$. Here, assets are randomly selected from the S$\&$P500 index, as previously done in Refs.~\cite{alcazar2021enhancing,Alcazar2020ClvsQuant}, and the return level $\rho$ is the same as used in previous studies. We impose the cardinality constraint that each portfolio must have a fixed Hamming weight $k = N/2 = 10$. As previously stated, such an essential restriction creates a subset of the search space $\mathcal{U}$, of size $2^{N} \sim O(10^6)$, defining a solution space $\mathcal{S}$ of size $\binom{N}{k} \sim O(10^5)$. The choice of these values allows for a big enough space so that generalization capabilities can be probed.

Given the solution space of portfolio candidates, the data distribution $P(x)$ given in \eqref{data_prob} used to assess validity-based generalization is automatically defined. To build a non-exhaustive $P_{\text{Train}}(x)$ as in \eqref{train_prob}, only a fixed number $T = \epsilon |\mathcal{S}|$ of training samples are randomly selected from the solution space, thus making the task of learning the distribution $P(x)$ highly non-trivial (despite it being defined as a uniform distribution over the valid bitstrings). Specifically, all generative models are trained for a fixed number of epochs $n_{\text{epochs}}=100$ with a fixed value of $T$ that equals 1\% of the solution space (i.e., $\epsilon=0.01$) , leaving the  remaining 99\% of the space available for testing generalization capabilities. Several values of this hyperparameter have been investigated, and we found this particular percentage to be a good choice as it gives the models many chances of generalizing, while providing enough samples $T\sim O(10^3)$ for the learning process to be successful. 
In order to assess quality-based generalization, we conduct the same process outlined above, with the addition of a pre-processing step that uses a softmax function to introduce risk-based information in the training dataset, so that low-risk portfolios are assigned a higher probability, and sampled with higher frequency. 

We investigate the generalization behaviours of different versions of the TNBM and GAN architectures, using various hyperparameter sets. In the case of the TNBM, we consider different values for the bond dimension $\alpha$, as this is the main parameter that affects the model quality. For GANs, the choice of hyperparameters is significantly more challenging~\cite{lucic2018gans}. Therefore, in addition to identifying hyperparameters via a trial-and-error procedure, we investigate whether automated hyperparameter optimization using Optuna~\cite{optuna_2019} could significantly improve the performance. We propose three different GANs that only differ in their hyperparameters as per \tableref{table:GAN-params} in \appendixref{app:training-details}, and show generalization behaviours for all of them. From here onward, we refer to a GAN that has mode collapsed onto one seen and valid bitstring as GAN-MC and to the Optuna enhanced GAN as GAN+.

As mentioned above, all models have been trained for a fixed number of epochs and the associated generalization metrics have been computed based on a fixed number $Q=10^5$ of queries retrieved from the trained model returned after the last epoch. Other strategies can be employed, such as considering the set of weights associated to the lowest loss function during training, or including more advanced training techniques such as early stopping. We decided to leverage a simple training scheme to avoid introducing any training bias and allow for the fairest comparison of the two models under examination. We also chose to sample this high magnitude of queries since this was not a limitation for the problem size considered here. However, in \appendixref{app:metric_trends} we present the behaviour of our sample-based metrics as a function of the number of queries. All of the numerical experiments in this work were carried out with Orquestra$^\text{\textregistered}$ \footnote{https://www.orquestra.io/} for workflow and data management. 

\subsection{Metric Robustness}\label{sec:metric_robustness}

The first step to validate our approach consists in showing the robustness of our sample-based metrics. To verify this, we conduct a statistical analysis of the generalization metrics' values and investigate the statistical errors associated to them. In addition, we propose an initial numerical investigation of the relationship between the values of our sample-based metrics and the distance measure from the model's distribution to the ground truth data distribution, in order to understand how the models' performance benchmarking approach connects with that of computational learning theory. 

We focus the robustness analysis on one instance of each of the two generative models presented in \secref{exp_details}. Specifically, we consider a TNBM model with fixed bond dimension $\alpha=7$, which has proven to be a good choice for generalization purposes as will be explained in \secref{sec:pitfalls}. For GAN, we consider the set of hyperparameters displayed in the first column of \tableref{table:GAN-params} in \appendixref{app:training-details}, which were selected as reasonable values via a trial and error procedure (i.e., without leveraging automated hyperparameter optimization). The analysis can be extended to other instances to further strengthen the evidence of the robustness of our metrics.

\begin{figure}[!t]
    \includegraphics[scale=1.0, width=\linewidth]{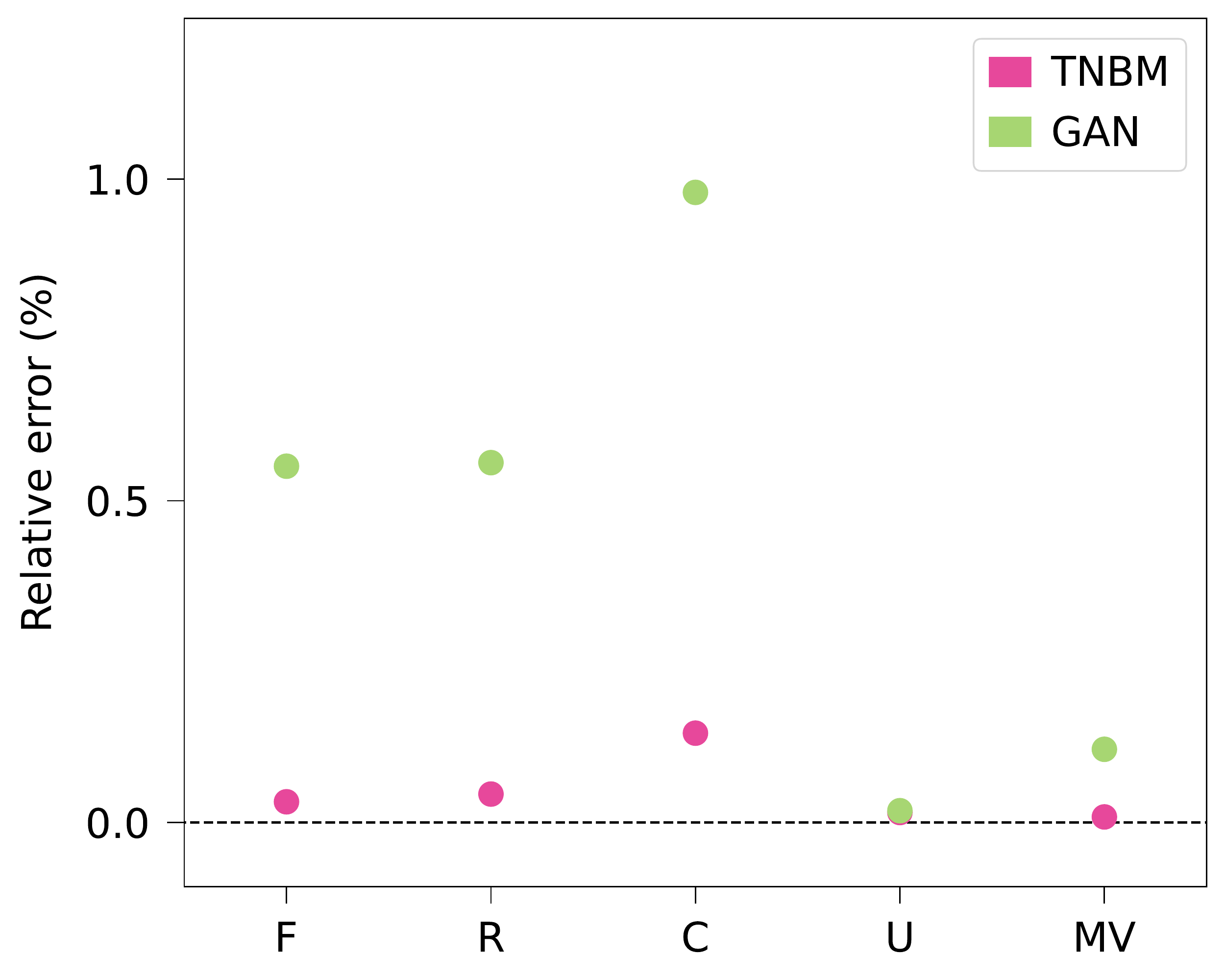}
\caption{\textbf{Robustness of the generalization metrics.} The plot shows the relative percentage error associated to each of the generalization metrics proposed in \secref{s:gen_metrics}, listed on the $x$ axis. The errors are estimated as the relative standard deviation of independent metric values computed on 30 sets of queries generated by trained TNBM (pink) and GAN (green) models. The proposed metrics show their statistical robustness: the associated error is small, suggesting that our approach is sample-based but not sample-dependent. Henceforth, new independent same-size query batches from the trained model will produce similar metric results.}
\label{f:robust-error}
\end{figure}

After training these two model instances using gradient-based optimizers (see \tableref{table:GAN-params}), we perform 30 independent query retrievals and compute our generalization metrics on these distinct sample sets. We then evaluate the relative percentage error\footnote{Relative percentage error is defined as the standard deviation of the metric values over their average.} associated to each of the metrics to assess their statistical robustness. For each of the two models, the error values for both validity-based and quality-based metrics are shown in \figref{f:robust-error}.

The errors associated to the different metrics assume similar values for the TNBM and GAN: this supports our claim that our metrics are model-agnostic and can be used to evaluate generalization capabilities for any generative model of interest. Furthermore, we can see in \figref{f:robust-error} that the relative errors are less than 1\%, thus suggesting that our metrics show significant robustness when computed on different sets of queries. Hence, we can affirm that the metrics proposed in this work are sample-based but not sample-dependent across different query batches of the same size. 

\begin{figure}[htp]
\includegraphics[width=\linewidth]{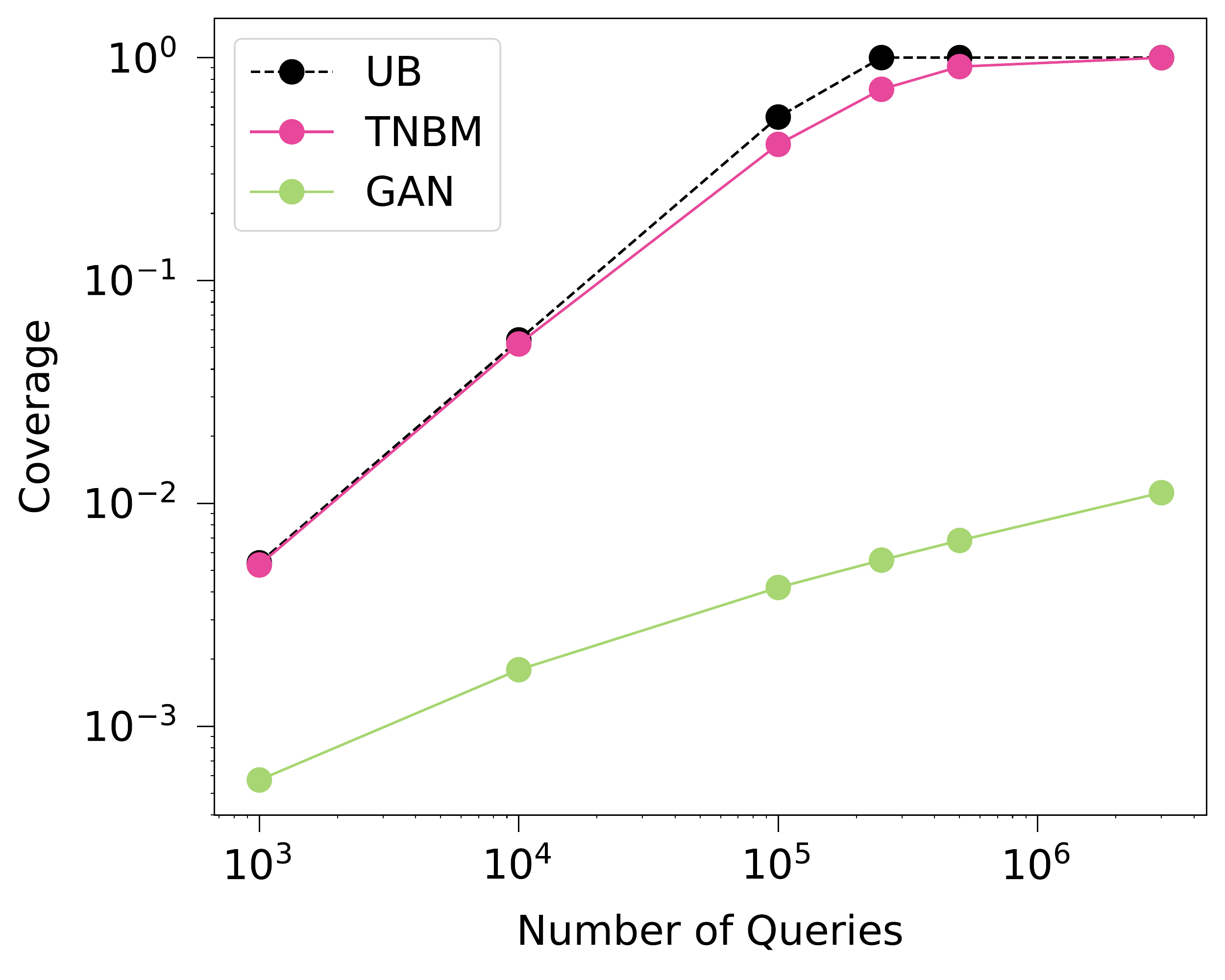}
\caption{\textbf{Coverage trends for increasing number of queries.} The plot displays the behaviours of the coverage metric for both TNBM (pink) and GAN (green) as we increase the number of queries $Q$ retrieved from the trained models. The dashed black line shows the upper bound $UB$ for each number of queries selected - i.e., the number of queries selected over the total size of the solution space. In the case of the TNBM, we observe that the coverage value follows the $UB$ curve and saturates to the ideal value of 1 for large numbers of $Q$, corresponding to the scenario in which the trained model is able to generate all unseen and valid samples. In the case of the GAN, we still observe that the coverage value gets closer to $UB$ and the ideal threshold of 1 when more and more queries are drawn from the model. However, it remains further from $UB$ and never reaches the desired threshold, suggesting that our GAN requires more queries than the TNBM to be able to reach all the unseen samples in the solution space.}
\label{fig:cov-trend}
\end{figure}

The latter statement requires further clarification in the case of the coverage metric in \eqref{coverage}. In this case, even though the coverage does not depend on the set of queries, it does depend on the number of queries that are retrieved from the trained model, as suggested in \secref{sec:constrained-gen}. The ideal coverage value of 1 is reached in the limit of a large number of queries, when the trained model has the opportunity to generate enough samples to cover most of the solution space. However, we note that, given a query budget $Q$, the effective upper bound $UB$ to the coverage value is set by $$UB=\frac{\min({Q, |\mathcal{S}|})}{|\mathcal{S}|} \leq 1,$$ thus implying that the ideal value of 1 can be reached only with a sufficiently high number of queries, i.e., $Q \geq |\mathcal{S}|$. We investigated if the models considered so far show this trend as we increase the number of queries retrieved after training from $10^4$ to $3\cdot 10^6$. The results of the simulations are displayed in \figref{fig:cov-trend}; in \appendixref{app:random-baseline} we compare them with the baseline given by random sampling from the search space $\mathcal{U}$. Results for how the other metrics vary with the number of queries $Q$ are shown in \appendixref{app:metric_trends}.

The data shows that the TNBM coverage closely resembles the $UB$ trend for any given value of $Q$ and saturates to the ideal value of 1 for a large enough number of queries, implying that this model is able to achieve excellent coverage. Conversely, the GAN coverage is further from the $UB$ and slowly increases without getting to the desired threshold, thus suggesting that significantly more queries would need to be taken to achieve a perfect coverage of all the unseen and valid patterns. Since there is no guarantee that the desired threshold is reached with a finite number of queries, this result might as well indicate that the model is quite poor at generalizing due to a high number of unreachable patterns. 
This is particularly relevant in the case of very large solution spaces $\mathcal{S}$. In this circumstance, the coverage metric has an intrinsic limitation: its low value might indicate that the number of generated queries is insufficient ($Q << |\mathcal{S}|-T$), rather than being due to poor generalization ($|g_{\text{sol}}| \approx 0$). Therefore, in order to mitigate the above issue when evaluating single models in the case of large problem sizes, we envision the denominator in $C$ to be replaced by the number of queries $Q$. This solution will slightly distort the meaning of coverage in \eqref{coverage} to a new metric quantifying the rate at which the model generates unique unseen and valid samples. When extending to large problem sizes, we see this as a more relevant evaluation metric as one cares more about the diversity of unique unseen and valid samples the model can reach rather than reaching all of them, which would be impossible without the number of queries being at least the size of the solution space. However, as our experiments are conducted with a mid-sized problem space, we stick to the definition in \eqref{coverage} for our evaluation.

Even though the coverage metric is dependent on the number of queries and its interpretation in terms of generalization is affected by the size of the solution space, we can draw a fair comparison between the coverage of different models. Indeed, we can compare TNBM and GAN models if we keep the number of queries generated from each fixed, as reported in \secref{sec:model_evaluation}, where it will be shown that the quantum-inspired model outperforms this GAN model when given the same sample budget.

Lastly, we put forth an initial investigation on the correlation between our metrics and the model's ability to infer the ground truth, as is the goal in computational learning theory discussed in \secref{review}. In \tableref{table:valid_learning_theory}, we report the average values of $(F, R, C)$ that result from five independent trainings of TNBMs with $\alpha = 7$. To take into account the fact that we span over a few $\epsilon$ values, we also show a normalized version of the rate value, given by $\tilde{R} = R/(1 - \epsilon)$. Alongside the $(F, R, C)$ values, we record two versions of the KL divergence: the quantity $KL_{\text{Train}}$, computed as usual between the model's output distribution and the training distribution in \eqref{train_prob}, and the quantity $KL_{\text{Target}}$, computed between the model's output distribution and the uniform ground truth data distribution in \eqref{data_prob}. Note that the latter is not usually available in real-world scenarios, since the ground truth is unknown; however, we find it relevant to analyze this quantity to validate our practical approach to generalization by relating it to computational learning theory. We see that with access to very little data $(\epsilon = 0.01)$, the model yields high $(F, R, C)$ values and gets closer to the data distribution than the training distribution -  as $ KL_{\text{Target}} < KL_{\text{Train}}$. When we increase $\epsilon$ to half of the solution space, we see that the $(F, R, C)$ metrics increase and the model is also able to approximate the ground truth more closely, since $KL_{\text{Target}}$ decreases. Hence, we see a promising correlation between our metrics' values and the model's ability to infer the ground truth in both of these data regimes. 

The main discrepancy between the two approaches occurs when the model is provided all of the data during training $(\epsilon=1)$. In this case that, we see that $ KL_{\text{Target}} = KL_{\text{Train}}$, and thus there is no room for generalization to occur, as defined in \secref{s:gen_define}. Therefore, the metrics' values are either zero or undefined (nan) in this instance. Despite this, we still see that the model is able to learn the ground truth well, as indicated by a low KL value. The ability to assess this \emph{memorization} behaviour is the main distinction between our practical approach in evaluating generalization and the one utilized in computational learning theory. From a practical standpoint, being able to identify this behaviour is highly relevant, thus supporting the need for a more practical approach to generalization to be considered in parallel to the theoretical one. In \appendixref{app:supp_figs}, we show multiple plots that report our metrics' values throughout the entire training alongside the $KL_{\text{Train}}, KL_{\text{Target}}$ values for a more complete analysis. Remarkably, the different panels in \figref{f:stat_approach} demonstrate excellent correlations between the theoretical and practical approaches, while also highlighting the value of having a multidimensional evaluation perspective, which provides enhanced explainability when assessing strengths and weaknesses of generative models. We note that while this example indicates a good correlation between our metrics' values and the ground truth inference ability, more investigations are necessary to strengthen the understanding of this relationship, potentially including theoretical proofs that establish precise connections between the two approaches. 

\begin{table}[htb]
\centering
\renewcommand{\arraystretch}{1.3}
\resizebox{\linewidth}{!}{
\begin{tabular}{||>{\centering}p{3cm} | >{\centering}p{1.5cm} | >{\centering}p{1.5cm} | >{\centering}p{1.5cm} ||}
\hline
\multicolumn{1}{||c|}{\textbf{Metric}} & \multicolumn{1}{|c|}{$\mathbf{\epsilon = 0.01}$} & \multicolumn{1}{|c|}{$\mathbf{\epsilon = 0.5}$} & \multicolumn{1}{|c||}{$\mathbf{\epsilon = 1.0}$}  \\
\hline
\multicolumn{1}{||c|}{$F$} & \multicolumn{1}{|c|}{$0.979(0.38\%$)} &\multicolumn{1}{|c|}{$0.986 (0.06\%)$} &\multicolumn{1}{|c||}{$0.0$} \\
\hline
\multicolumn{1}{||c|}{$R$} & \multicolumn{1}{|c|}{$0.969 (0.39\%)$} &\multicolumn{1}{|c|}{$0.497 (0.28\%)$} &\multicolumn{1}{|c||}{$0.0$}  \\
\hline
\multicolumn{1}{||c|}{$\tilde{R}$} & \multicolumn{1}{|c|}{$0.979 (0.39\%)$} &\multicolumn{1}{|c|}{$0.993 (0.28\%)$} &\multicolumn{1}{|c||}{nan}   \\
\hline
\multicolumn{1}{||c|}{$C$}& \multicolumn{1}{|c|}{$0.405 (0.52\%)$} &\multicolumn{1}{|c|}{$0.416 (0.32\%)$} &\multicolumn{1}{|c||}{nan}  \\
\hline
\hline
\multicolumn{1}{||c|}{$KL_\text{Train}$} & \multicolumn{1}{|c|}{$4.575 (0.07\%)$} &\multicolumn{1}{|c|}{$0.702 (0.01\%)$} &\multicolumn{1}{|c||}{$0.009 (0.36\%)$}  \\
\hline
\multicolumn{1}{||c|}{$KL_\text{Target}$} & \multicolumn{1}{|c|}{$0.074 (13.28\%)$} &\multicolumn{1}{|c|}{$0.009 (0.56\%)$} &\multicolumn{1}{|c||}{$0.009 (0.36\%)$}  \\
\hline
\end{tabular}}
\caption{\textbf{The relationship between the validity-based metrics and learning the ground truth for the TNBM}. Down each column, we record the final average $(F, R, C)$ metrics' values (including the normalized rate $\tilde{R}$) along with the average KL divergences of the model output distribution relative to the training distribution, denoted as $KL_{\text{Train}}$, and to the data distribution, denoted as $KL_{\text{Target}}$. We see that there is a good correlation between the high-scoring metrics' values and learning the ground truth distribution, even in multiple data regimes. We see that the largest discrepancy between the two frameworks exists when $\epsilon=1$, where $KL_{\text{Target}} = KL_{\text{Train}}$ reaches a low value, but the other metrics are either zero or undefined. This is a case of \emph{memorization}, where the model still scores high in the context of learning the ground truth, while demonstrating poor performance from a practical generalization standpoint. This is expected from a practical perspective: the generative model cannot add value in terms of generating novel samples, since all of them were given as part of the training set. All relative percentages errors are computed across five independent trainings.}
\label{table:valid_learning_theory}
\end{table}

\subsection{Spotting Pitfalls in Generative Model Training }\label{sec:pitfalls}
We further demonstrate that we can use our metrics to detect common pitfalls that are known to affect the training of the TNBM and GAN models. This result strengthens the validity of our approach, which turns out not only to be a good framework for quantifying generalization of generative models, but also to enhance the study of their trainability. In the following sections, we show an example of this study for each of the models. For the TNBM, we analyze the relation between the bond dimension $\alpha$, our generalization metrics, and the trainability of the model. Conversely, for the GAN, we investigate the relation between our metrics and mode collapse. Additional results to compare the training stability of the two classes of models are shown in \figref{fig:training-stability} in \appendixref{app:training-details}.

\subsubsection{TNBM Bond Dimension and Trainability}\label{sec:pitfalls-mps}
In the TNBM architecture, the bond dimension $\alpha$ of the MPS plays an important role in the model’s ability to generate good quality samples as it is directly correlated with the expressive power of the model. Typically, increasing the bond dimension leads to a better model approximation. We take this one step further and directly connect bond dimension to the model's generalization behaviour and trainability.
\begin{figure}[h]
\includegraphics[width=\linewidth]{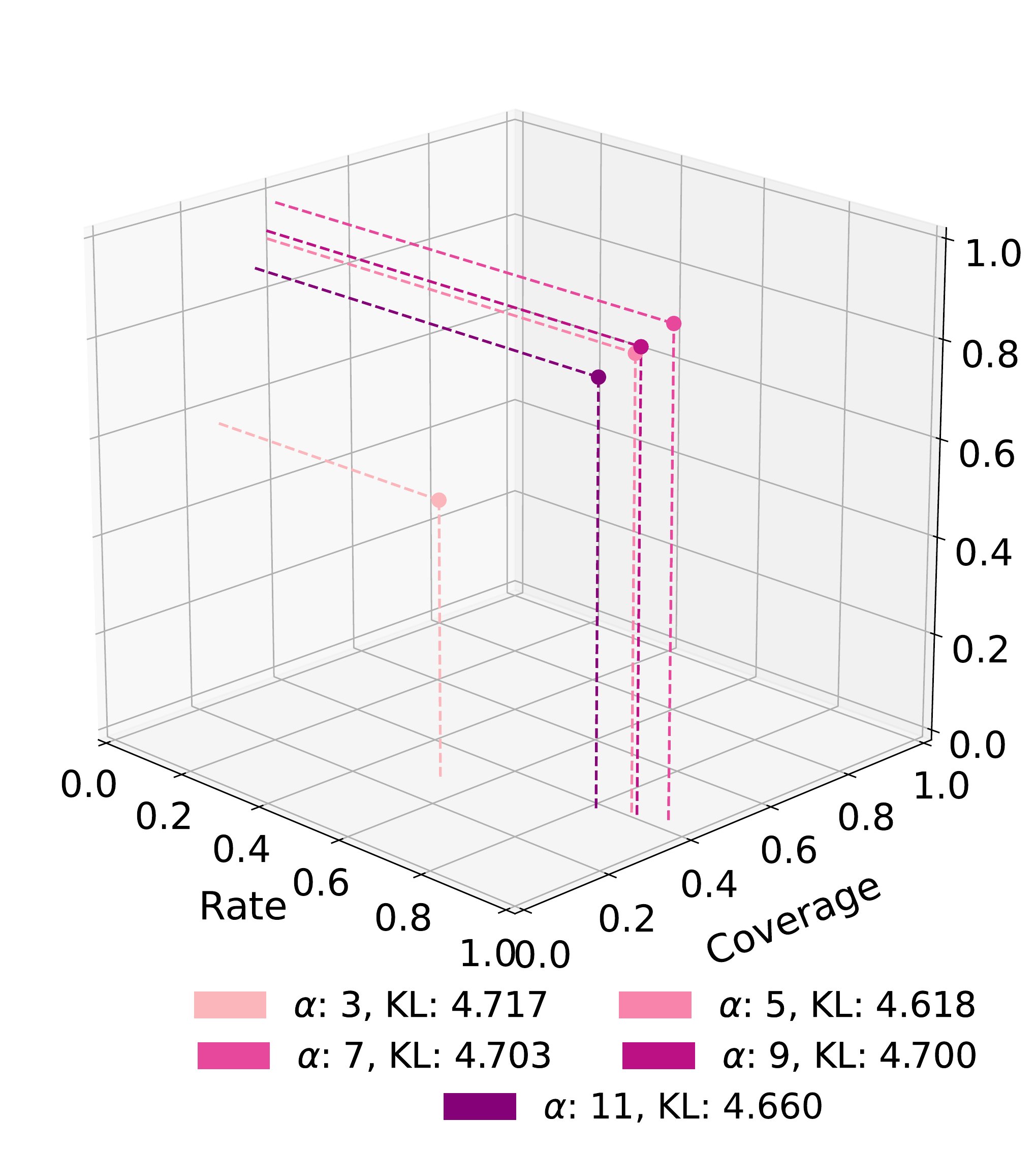}
\caption{\textbf{Training and generalization behaviours of the TNBM with different bond dimensions.} The plot displays the 3D evaluation of the validity-based generalization capabilities of the TNBM models with various $\alpha \in \{3, 5, 7, 9, 11\}$. Each data point corresponds to the average metrics' values, whose associated error is too small to be visible on the plot.  The legend connects each $\alpha$ to the last KL divergence value in the training after 100 epochs. The plot demonstrates that for various $\alpha$ values, there is a connection between KL divergence values of the model distribution to the training distribution, thus establishing a link between this capability and trainability properties of generative models.}
\label{fig:TNBM-training}
\end{figure}

In light of this goal, we train five different instances of the TNBM architecture on a fixed training dataset with various bond dimensions $\alpha \in \{3, 5, 7, 9, 11\}$. For a given $\alpha$ value, we select a typical\footnote{A typical training instance is identified as the resulting model from the median value of the loss function (i.e., KL divergence) at the last epoch, out of 30 independent trainings.} training and build a model with the last set of parameters retrieved after the learning process. We then generate 15 independent query batches from the trained model and compute our validity-based generalization metrics $(F, R, C)$. We show the results in \figref{fig:TNBM-training}, where we display the average metric evaluations for each bond dimension $\alpha$. In the plot legend, we report the last loss function value during training (complete training loss curves can be found in \appendixref{app:training-details}).

From \figref{fig:TNBM-training}, it can be seen that the median value of the KL divergence occurs for $\alpha =7$: this result motivates the usage of such value in \secref{sec:metric_robustness}, as it suggests that the training is most typical for this choice of the hyperparameter value. It is not surprising that the lowest value of the loss function, obtained for $\alpha=5$, does not correspond to the best validity-based generalization performance, as shown in \figref{fig:TNBM-training}, because this loss is relative to the training rather than the data distribution. If the model was to perfectly fit the training distribution, we would see data-copying rather than generalization behavior - which is a form of overfitting. We expect that our metrics will be able to identify similar overfitting behaviours when associated to an extremely successful training curve (\tableref{table:cheat-sheet}).

As the bond dimension grows, we see an increase in $(F, R, C)$ up to $\alpha = 7$, and then the metrics' values begin to decrease. Thus, it seems that we are hitting a trainability \textit{Goldilocks region} around $\alpha \approx 7$, with $\alpha < 7$ leading to underperforming models and $\alpha > 7$ being too expressive for the model to be able to generalize successfully. 
These results demonstrate that we can use our metrics to identify thresholds in hyperparameter tuning and to get insights on the trainability of the model as it relates to generalization. 

\subsubsection{Mode collapse in GAN}\label{MC-GANS}
\begin{figure*}[bht]
\subfloat[\label{sfig:gan-training-a}]{%
  \includegraphics[width=0.45\linewidth]{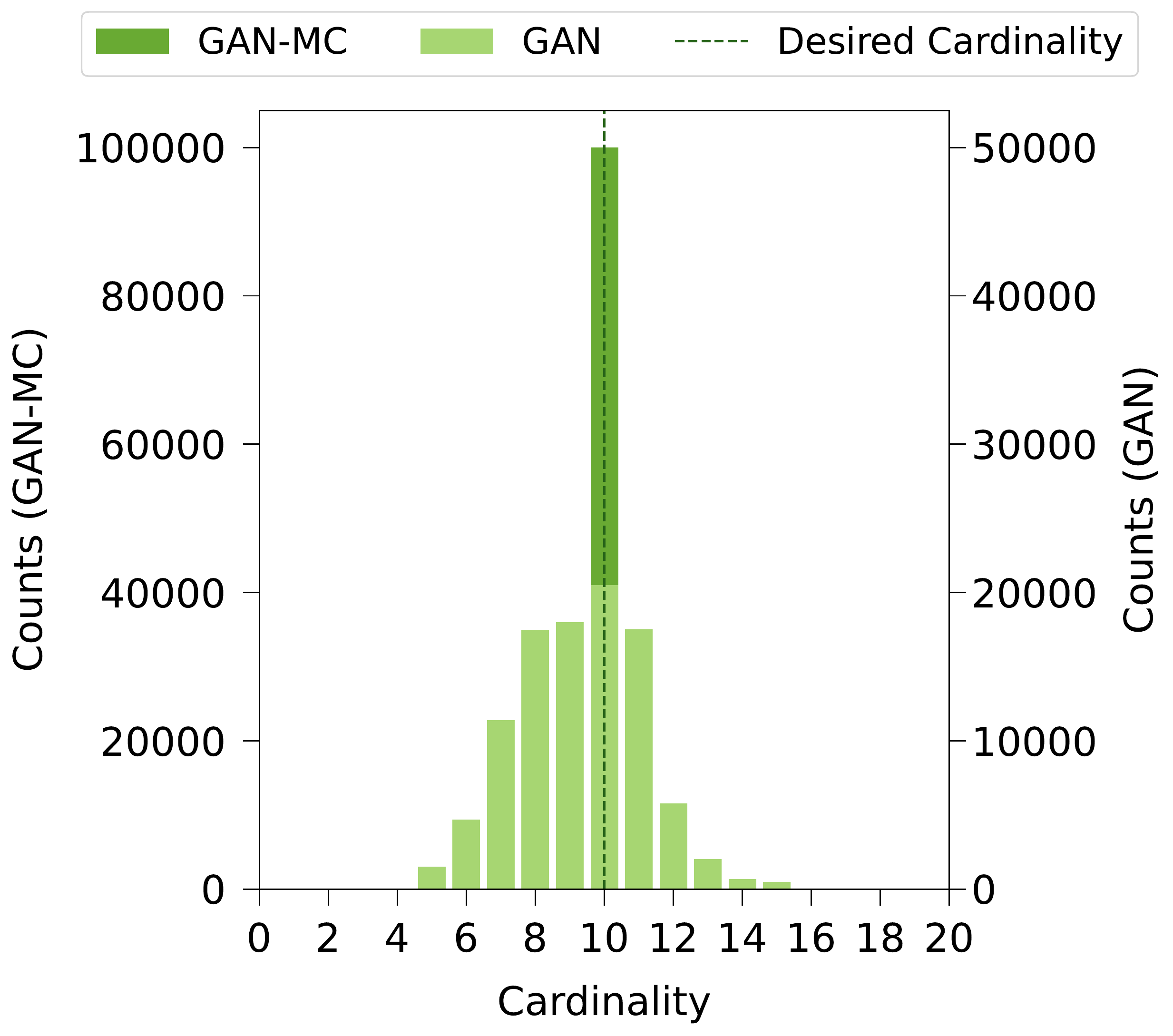}%
}\hfill
\subfloat[\label{sfig:gan-training-b}]{%
  \includegraphics[width=0.54\linewidth]{diversity.pdf}%
}
\caption{\textbf{Visualization of mode collapse in GAN training.} \figref{sfig:gan-training-a} shows the cardinality distribution of generated queries for GAN and GAN-MC, indicating that GAN-MC produces only samples with the desired cardinality (dashed line), whereas the GAN queries populate a larger subset of the cardinality domain. Hence, GAN-MC is associated to perfect fidelity $F=1$ and rate $R=1$. However, in \figref{sfig:gan-training-b} the queries' diversity is displayed, where the $x$ axis represents the set of distinct generated bitstrings (for readability, bitstrings labels are not shown, and only bitstrings with counts $>$ 50 have been included in the histogram). We can see that GAN-MC always generates the same unseen and valid query (mode collapse phenomenon), as opposed to GAN, which is able to cover a significantly larger portion of the solution space, as reflected by the metrics' value in \tableref{table:comparison}. Note the different scales for the $y$ axes in both \figref{sfig:gan-training-a} and \figref{sfig:gan-training-b}.}
\label{fig:gan-training}
\end{figure*}

One of the major issues that affects GAN training is the so-called mode collapse behaviour~\cite{che2016mode}. This undesired phenomenon occurs when the generator learns to produce a very limited number (sometimes only one) of highly plausible outputs, thus affecting the ability of the generative model to  further explore the solution space. Since mode collapse is a well-known pitfall, several strategies have been proposed to mitigate this issue in the context of GANs, among which a promising algorithm is the Wasserstein GAN~\cite{arjovsky2017wasserstein, gulrajani2017improved}.

We propose an example of how our metrics are able to detect mode collapse, when it occurs. We fine tune our hyperparameters such that the GAN exhibits mode collapse behaviour (see details in \tableref{table:GAN-params} in \appendixref{app:training-details}) for a fixed training dataset. We run a typical\footnote{A typical training instance is identified among 30 independent trainings as the one whose mode collapse shows the correct cardinality and whose last associated Hausdorff distance \cite{Huttenlocher1993ComparingIU} during training is the median. We highlight that the training is performed via an adversarial strategy, hence we use the Hausdorff distance only as a figure of merit to monitor the training.} training of this GAN-MC architecture, and then sample 15 query batches from the trained model to compute our generalization metrics $(F, R, C)$.

We display the validity-based metrics for the GAN and GAN-MC in \tableref{table:comparison}. For the GAN-MC, we see that fidelity and rate are the ideal value of 1, thus suggesting that the model generates exclusively unseen samples with the desired cardinality. However, the coverage value is close to 0, thus it is far from its ideal threshold, since the model is only able to produce one single pattern and does not have the ability to explore the solution space and cover it as much as possible. Such anomaly in the validity-based generalization metrics' values is not present if the training of a GAN doesn't exhibit training pitfalls, as displayed by the GAN results in the same table. 

We note that these metrics' values only capture mode collapse behaviour for models that collapse onto an unseen and valid bitstring. {If the model were to collapse onto a seen bitstring (in-training mode collapse), $F$ would be not well-defined and both $C$ and $R$ would equal zero. These metrics' values would be indistinguishable from the perfect memorization regime. In order to avoid this, one should also compare the number of individual queries generated, $|d_{\text{gen}}|$, to the size of the training set $T$. This would provide the additional information necessary to detect any form of mode collapse. Expected metrics' values for various mode collapse behaviours along with other model training pitfalls are displayed in \tableref{table:cheat-sheet} in \appendixref{app:cheat_sheet}. In summary, our metrics reflect mode collapse upon occurrence and therefore they can provide insights on the training progress of generative models.

In order to better visualize the difference between the two aforementioned models and detect the mode collapse phenomenon, in \figref{sfig:gan-training-a} we display the cardinality distribution of the generated queries for the two GAN variants under examination: for GAN, the distribution is centered around the correct cardinality but shows a larger spread as compared to the case of GAN-MC, where all the queries satisfy the cardinality constraint. Nevertheless, \figref{sfig:gan-training-b} allows one to identify the occurrence of mode collapse onto an unseen and valid bitstring: the GAN-MC model generates always the same query, as opposed to the diversity of samples retrieved from GAN.

These results demonstrate that we can use our metrics to identify the occurrence of a very well-known pitfall affecting the learning process of GANs, thus providing an insightful tool for the challenging task of monitoring the training of generative models.

\subsection{Evaluating and Comparing Models}\label{sec:model_evaluation}
We use our quantitative metric-based approach to evaluate the validity-based and quality-based generalization capabilities across different generative models and compare their performance.

We run 30 independent trainings for a fixed training dataset and choose the best run, which we define as the run with the lowest loss function at the end of the trainings. Then, we generate 15 query batches from such trained model, for each of the generative models under examination. We note that while we use a fixed training dataset to compare models, this evaluation method holds across multiple training datasets that could be selected from a specific problem instance. Indeed, each dataset is characterized by the same asset universe, cardinality, and seen portion $\epsilon$, but different datasets can be built by simply uniformly drawing independent bitstring subsets from the support of $P(x)$. We perform this analysis in \appendixref{app:metric_trends}, showing that validity-based and quality-based generalization metrics for 15 different training datasets display similar values, thus showcasing the robustness of the models' behaviour, and the conclusions shown in this work.

For validity-based generalization, we construct $\mathcal{D}_{\text{Train}}$ by sampling from a $P(x)$ that is uniform over the solution space of cardinality-constrained bitstrings, whereas for quality-based generalization, $\mathcal{D}_{\text{Train}}$ is re-weighted with cost-related information, i.e.,  from $P^{\text{(w)}}_{\text{Train}}(x)$, as in \eqref{biased_prob}. As stated previously, we use one fixed dataset for our evaluation in \secref{ValidityB-Gen} and \secref{ValueB-Gen}. Post training, $Q=10^5$ queries are collected from each model for comparison.

\subsubsection{Validity-Based Generalization}\label{ValidityB-Gen}
We first show the validity-based generalization results for each type of model. While we present these results as both an evaluation and comparison of models, we would like to emphasize that our results do not speak for all GAN or TNBM models, as each type of model may contain various hyper-parameters, multi-layered architectures, and other variances that would lead to different results. Thus, we restrict our comparison to the specific models we trained, as described in \secref{exp_details} with GAN hyperparameters listed in \tableref{table:GAN-params} (\appendixref{app:training-details}). We choose to focus on using these models to demonstrate the robustness of our framework and metrics, such that when exploring various GAN, TNBM, or alternative model architectures, this approach can be replicated.

Results for $(F, R, C)$ are listed in \tableref{table:comparison}, along with the values of the exploration $E$; the corresponding results for the metrics' baseline given by random sampling from the search space are reported in \appendixref{app:random-baseline}.
Additionally, we visualize the average validity-based metrics in \figref{fig:model_comparison} through a 3D representation. Lastly, \figref{sfig:2D-Viz} in \appendixref{app:supp_figs} gives an intuition of how the two models perform and allows to visualize their different abilities in reconstructing the data distribution $P(x)$, showing the remarkable performance of the TNBM as reflected in the metrics’ values.

\begin{table}[htb]
\centering
\renewcommand{\arraystretch}{1.3}
\resizebox{\linewidth}{!}{
\begin{tabular}{||>{\centering}p{3cm} | >{\centering}p{1.5cm} | >{\centering}p{1.5cm} | >{\centering}p{1.5cm} || >{\centering}p{1.5cm} ||}
\hline
\multicolumn{1}{||c|}{\textbf{Metric}} & \multicolumn{1}{|c|}{\textbf{TNBM}} & \multicolumn{1}{|c|}{\textbf{GAN}} & \multicolumn{1}{|c|}{\textbf{GAN-MC}} & \multicolumn{1}{|c||}{\textbf{GAN+}}  \\
\hline
\multicolumn{1}{||c|}{$E$} & \multicolumn{1}{|c|}{$0.989 (0.02\%$)} &\multicolumn{1}{|c|}{$0.995 (0.02\%)$} &\multicolumn{1}{|c|}{$1.0$} & \multicolumn{1}{|c||}{$1.0 (0.003\%)$} \\
\hline
\multicolumn{1}{||c|}{$F$} & \multicolumn{1}{|c|}{$0.989 (0.03\%)$} &\multicolumn{1}{|c|}{$0.263 (0.6\%)$} &\multicolumn{1}{|c|}{$1.0$} & \multicolumn{1}{|c||}{$0.243 (0.4\%)$} \\
\hline
\multicolumn{1}{||c|}{$R$}& \multicolumn{1}{|c|}{$0.978 (0.03\%)$} &\multicolumn{1}{|c|}{$0.261 (0.6\%)$} &\multicolumn{1}{|c|}{$1.0$} & \multicolumn{1}{|c||}{$0.243 (0.4\%)$} \\
\hline
\multicolumn{1}{||c|}{$C$} & \multicolumn{1}{|c|}{$0.409 (0.15\%)$} &\multicolumn{1}{|c|}{$0.006 (1.7\%)$} &\multicolumn{1}{|c|}{$5.5 \cdot 10^{-6}$} & \multicolumn{1}{|c||}{$0.001 (2.5\%)$} \\
\hline
\hline
\multicolumn{1}{||c|}{$C / \overline C$} & \multicolumn{1}{|c|}{$0.971$} &\multicolumn{1}{|c|}{$0.014$} &\multicolumn{1}{|c|}{$1.0 \cdot 10^{-5}$} & \multicolumn{1}{|c||}{$0.002$} \\
\hline
\end{tabular}}
\caption{\textbf{Pre-Generalization and validity-based generalization metrics for all models.} We display the average exploration $E$ and the average $(F, R, C)$ values for each best model run with an average and the associated relative percentage error across 15 query batches.
All the models exhibit a high exploration rate, thus showing that data-copying is not occurring. We see that our TNBM model outperforms our GAN and GAN+ models by more than $70$ percentage points for $F$ and $R$. $C$ is about 68x larger for TNBM than the GAN models. We further include the ratio of the coverage $C$ to the ideal expected coverage $\overline C$ to highlight the large difference between the TNBM and the GAN's ability to successfully learn the underlying data distribution $P(x)$. Additionally, for GAN-MC, we see perfect $F$ and $R$ and a near zero $C$ value, indicating mode collapse behaviour. Note that no error is provided for the GAN-MC as all the models produce exactly the same values for the metric, except for the coverage whose associated error is negligible.}
\label{table:comparison}
\end{table}

In evaluating our models, we see that the TNBM is a clear winner with average values $(0.989, 0.978, 0.409)$. The model achieves near-perfect rate and fidelity. As the maximum coverage one can achieve is the number of queries over the size of the solution space ($UB=0.54$), the TNBM performs remarkably well. Indeed, the ratio of the average coverage to the upper bound $UB$ for the TNBM is high, i.e. $C/UB=76\%$. However, we note that the upper bound represents a scenario that would rarely happen in practice, thus representing a pessimistic reference value. A more realistic reference can be derived if one considers the ideal expected coverage $\overline C$ when sampling from the data distribution $P(x)$. By means of simple statistical considerations (see e.g.,~\cite{blogpost-simple, blogpost-full}), it can be shown that 
$$\overline C =  1 - (1 - \frac{1}{|\mathcal{S}|-T})^Q,$$ and this estimator indicates which coverage $C$ one should expect when the generative model has perfectly learned the data distribution and generates samples accordingly. 
When comparing the average TNBM coverage to this more realistic reference value, we obtain a surprisingly high value of $97\%$, which shows that the model has learned an extremely good approximation of the data distribution $P(x)$. In \tableref{table:comparison}, we include $C / \overline C$ values for all models in order to highlight how well each model's average coverage compares to the ideal expected coverage. The limit of $C / \overline C \rightarrow 1.0$ holds for models with perfect generalization.

As shown in \figref{fig:cov-trend}, the TNBM is able to achieve an improved coverage when sampling up to 3 million queries. The model has a high exploration rate of $98.9\%$, i.e. $E=0.989$, such that most of the generated samples were not fed to the model during training. The GAN has much poorer average $(F, R, C)$ values with a slightly higher exploration rate than the TNBM, thus showing that neither of them is performing mere data-copying. The GAN achieves metric values $(0.263, 0.261, 0.006)$, but $99.5\%$ of its generated samples are outside of the training set. One can conclude that while the GAN has the potential to produce novel samples, it requires improved optimization strategies in order to avoid generating noisy samples - i.e. samples that do not match the cardinality constraint - so that fidelity and rate can grow to larger values. The GAN is not able to learn the underlying features as well as the TNBM, and thus is not able to generalize as well. Lastly, we compute the TNBM-to-GAN ratios for the validity-based metrics, and see that the TNBM is $(3.76, 3.75, 68.2)\times$ better than the GAN, respectively across $(F, R, C)$ values. We would like to highlight that using metric ratios, rather than absolute values, allows one to have a clearer picture of the relation between different models, and this strategy is especially useful when considering the coverage, whose absolute value has been shown to be more heavily affected by the number of collected queries $Q$.


\begin{figure}[htp]
\includegraphics[width=\linewidth]{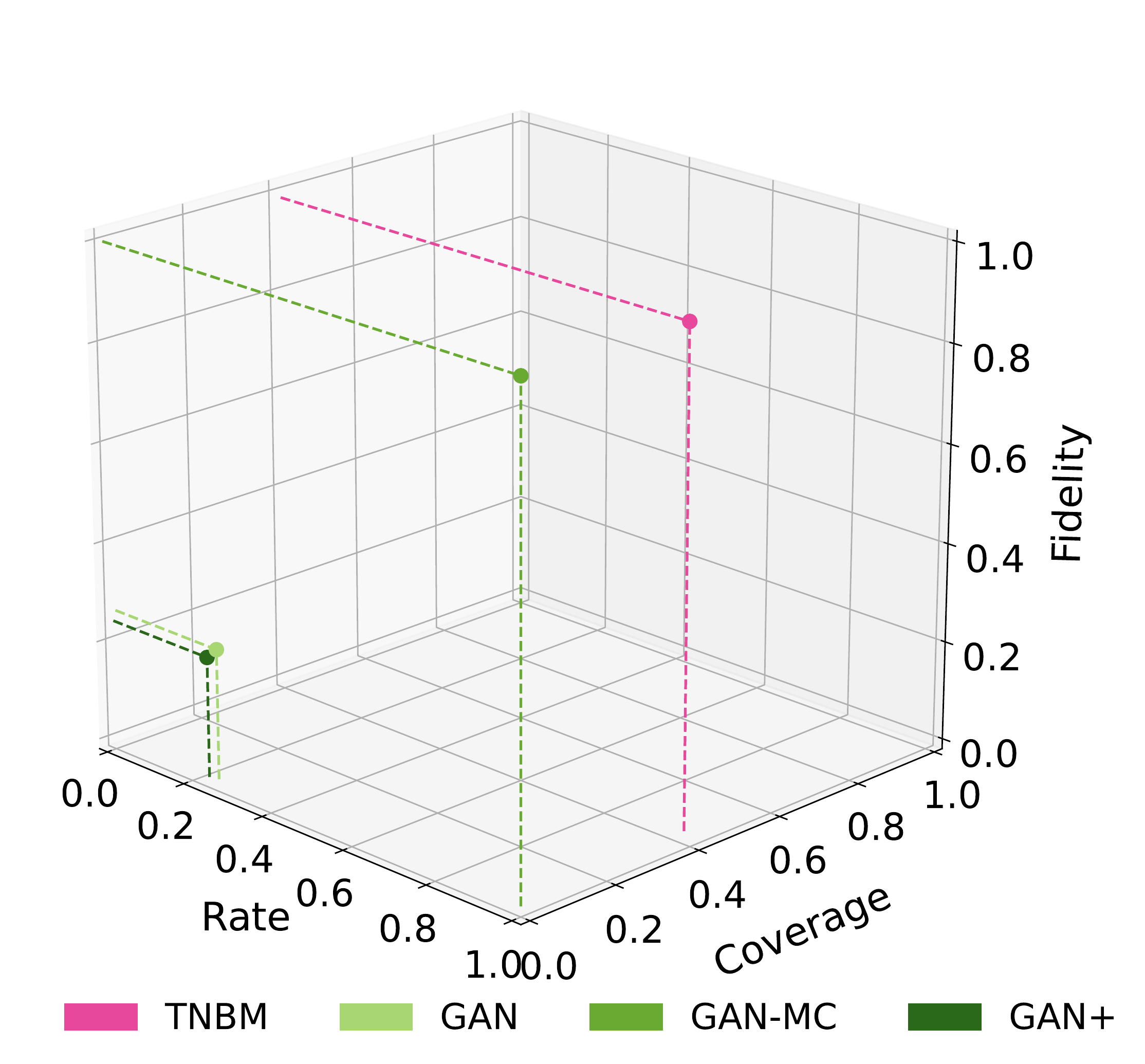}
\caption{\textbf{3D evaluation of validity-based generalization metrics for different generative models.} The plot displays results for four models, namely the TNBM with $\alpha = 7$ (pink), GAN (light green), GAN-MC (medium green) and GAN+ (dark green). The solid points show the average $(F, R, C)$ values across 15 query batches, whose associated error is too small to be visible in the plot. We see that our TNBM is the clear winner compared to our GAN models.}
\label{fig:model_comparison}
\end{figure}

As explained in \secref{MC-GANS}, we further show visually that our metrics detect mode collapse in GANs. The GAN-MC has an exploration rate of $100\%$ ($E=1$), demonstrating that the single generated sample was not introduced in the training set. Without the prior knowledge that the model exhibits mode collapse, we can use the average $(F, R, C)$ values $(1.0, 1.0, 5.5\text{e-}6)$ to detect this behaviour. If perfect fidelity and rate are achieved, with a coverage near zero, we can conclude that the model has focused in too closely on one or a few unseen and valid bitstrings. In general, whenever $C \rightarrow 0$ we can safely identify the behaviour as mode collapse. 

Then, we consider the $(F, R, C)$ values of the GAN+ and see that while the GAN+ is able to explore slightly more than the GAN, the $(F, R, C)$ values are very similar, namely $(0.243, 0.243, 0.001)$, showing that the optimization scheme with Optuna doesn't bring a significant improvement for our specific GAN model in terms of generalization.

Lastly, we note that $F$ and $R$ are highly correlated for each trained model. This is the case only because in all of the models studied here the exploration $E$ is quite high ($E \approx 1$). In this limit, and given that $R = E F$, then we have $R \approx F$. It is important to note that there is no reason to expect a value of $E$ to be similar across all models, as it happened for the GAN and TNBM explored here.

\subsubsection{Quality-Based Generalization}\label{ValueB-Gen}
We evaluate our generative models' ability to generate high quality samples using our quality-based approach and metrics. The models (TNBM and GAN) are evaluated across the two \emph{sample quality} metrics described in \secref{s:vb-gen}: Minimum Value ($MV$) and Utility ($U$). Note that for calculating the $MV$, as discussed in \secref{s:vb-gen}, five batches of $Q=10^5$ queries were used. Hence, the total number of query retrievals used to compute this metric is $5\times$ the number of query sets one would desire for gathering statistics (in our case, $15 \times 5 = 75$ query sets, but this can be adjusted according to the available sampling budget). 

When averaged over the 15 independent query retrievals, both the TNBM and the GAN meet the conditions in \eqref{lowest_risk} and in \eqref{utility}, as shown in \tableref{table:utility}.
\begin{table}[htb]
\centering
\renewcommand{\arraystretch}{1.3}
\begin{tabular}{||>{\centering}p{3cm} | >{\centering}p{1.5cm} | >{\centering}p{1.5cm}|>{\centering}p{1.5cm} ||}
\hline
\multicolumn{1}{||c|}{\textbf{Metric}} & \multicolumn{1}{|c|}{\textbf{TNBM}} & \multicolumn{1}{|c|}{\textbf{GAN}} & \multicolumn{1}{|c||}{\textbf{Threshold}}  \\
\hline
\multicolumn{1}{||c|}{$MV$} & \multicolumn{1}{|c|}{$0.1017 (0.01\%)$} &\multicolumn{1}{|c|}{$0.1024 (0.17\%)$} & \multicolumn{1}{|c||}{$0.1035$} \\
\hline
\multicolumn{1}{||c|}{$U$}& \multicolumn{1}{|c|}{$0.1049 (0.017\%)$} &\multicolumn{1}{|c|}{$0.1048 (0.02\%)$}  & \multicolumn{1}{|c||}{$0.1059$} \\
\hline

\end{tabular} 
\caption{\textbf{Quality-based generalization metrics for TNBM and GAN models.} The first column shows values obtained by averaging over all 15 query retrievals for the TNBM's sample quality performance, along with the associated relative percentage error. The second column displays the metrics' values and relative percentage error for the GAN model. The last column displays the training threshold, defined as the $MV$ and $U$ computed for the samples in $\mathcal{D}_{\text{Train}}$. We see that both the TNBM and the GAN meet the conditions in \eqref{lowest_risk} and \eqref{utility}.}
\label{table:utility}
\end{table}

We see that our TNBM exhibits a lower $MV$ than our GAN, even though both beat the training set on average. Thus, our TNBM model shows slightly enhanced performance when searching for a minimum value of the cost function $c(x)$, which is assumed to be the financial risk $\sigma(x)$ in the specific application we are considering. While this may be  relevant when one aims at finding the lowest possible minimum in an optimization task, it may not be the most important condition for alternative tasks that are simply looking for multiple low cost options - not necessarily the lowest. For example, when looking for a large frequency of low cost samples, the condition in \eqref{utility} may be more important and robust for comparing models. Note that the value of the utility threshold parameter $t$ can be set according to the task at hand. In our task, we take $t = 5\%$ as an appropriate threshold for demonstrating the model's ability to obtain the tail-end of the distribution over low-risk samples.

\begin{figure}[htp]
\centering
\includegraphics[width=\linewidth]{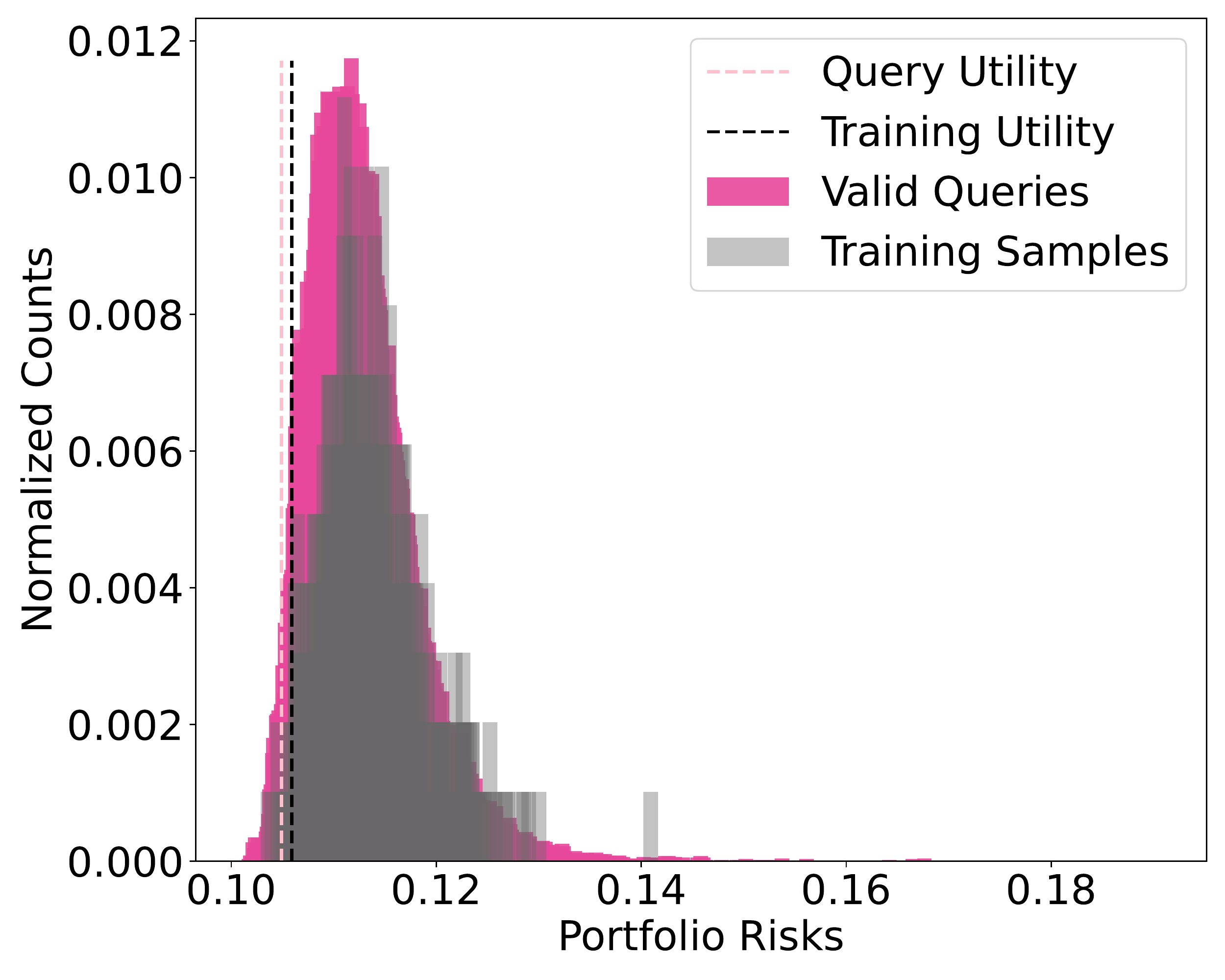}

\caption{\textbf{Visualization of quality-based metrics for TNBM-generated queries.} The plot displays the number of portfolio counts associated to given risk values. The pink spikes represent valid TNBM queries, whereas the gray spikes represent the samples from the training set. Note that for calculating our metrics, we used $Q=10^5$ queries, but the training distribution only contains $O(10^3)$ samples. We normalize the counts on the $y$ axis to provide a fair visual comparison between distributions, and we set the utility threshold to $t = 5\%$. Because the training distribution is re-weighted to favor lower risk values, the model distribution learns this feature in the dataset, and generates an even higher frequency of low risk values. The model queries have a lower utility (pink dashed) than the training set (black dashed), and the model is able to produce samples that have lower risks than those in the training set. We see that our TNBM model is able to effectively generalize to low-risk samples.}
\label{fig:TNBM_utility}
\end{figure}

From \tableref{table:utility}, we observe that the GAN and the TNBM have practically the same $U$, despite having such a large difference in $(F, R, C)$ values. We conclude that while both models generate new portfolios that happen to be similarly low in risk when taking the smallest $5\%$ of unseen and valid portfolio risks, the TNBM is simply able to generate more of them than the GAN (TNBM: $4556$, GAN: $843$, i.e.,  $5.4\times$). We display these utility samples for TNBM in \figref{fig:TNBM_utility}, demonstrating the comparison of $U$ relative to the training distribution $P_{\text{Train}}$. We include the same figure for the GAN in \appendixref{app:supp_figs}. 

Hence, the GAN is able to generalize to similarly low risk portfolios as the TNBM, but fewer in number and less diverse than those of the TNBM. Our $(F, R, C)$ metrics support that this generalization diversity is one of the largest differences between our TNBM and our GAN. Therefore, our TNBM model achieves superior performance when looking to produce a large diversified batch of new low-risk valid portfolios. We note that it remains an open question as to why the TNBM's performance is of such high quality. Investigating the nature of the model's inductive bias remains an ongoing research effort and opens an interesting opportunity to understand the power of quantum and quantum-inspired model when compared to their classical counterparts.

Lastly, we calculate the number of unique portfolios each model is able to produce that have a lower associated risk than a critical cost in the training set $c'(x)$. When this critical value is equivalent to the sample with the lowest risk in the training set, our TNBM on average is able to beat our GAN with a 61:4 ratio. In other words, our TNBM model is able to generalize to 61 unique portfolios that have a lower risk than the lowest risk in the training set, while the GAN can only produce 4 (i.e., $\sim 15\times$). We introduce this condition in \eqref{harsh_utility} on top of the other two metrics in order to have an additional layer to determine whether a model is suitable for generalization. Note that one could adjust this critical cost threshold $c'(x)$ to relax the restriction. 
For example, when $c'(x)$ is equivalent to the risk taken at cutoff of the lowest $5\%$ of samples in the training set, the TNBM-to-GAN ratio becomes 6709:345 on average (i.e., $\sim 19\times$). 

While the model might meet the \emph{sample quality} requirements \eqref{lowest_risk} and \eqref{utility}, it might be poor at finding many samples with lower cost than $c'(x)$, which is not ideal when one is not only concerned with the global minimum, but also with generating a large quantity of low-cost samples. Our GAN works well under these requirements. On the other hand, our TNBM model shows good quality performance for generalizing to both valid and quality-based portfolios with high diversity and frequency. 

\section*{Summary and Outlook}\label{s:conclusions}
In this work, we study the generalization performance of generative models in the context of measuring practical quantum advantage. We highlight that developing new approaches and frameworks to characterize the generalization capabilities of unsupervised generative models is still an ongoing research area in both the classical and quantum machine learning community~\cite{borji2021pros, sajjadi2018assessing, kynkaanniemi2019improved, naeem2020reliable,alaa2021faithful}. Thus, we first unify nomenclature for discussing practical generalization in generative models, clarifying our standpoint as compared to computational learning theory, next introduce a novel quantitative framework with metrics for identifying various behaviours with discrete datasets, and, finally, demonstrate the robustness of our approach by evaluating and comparing the generalization capabilities of two well-known generative models: classical GANs and quantum-inspired TNBMs. We highlight that the main goal of our work is to provide this fundamental tool that can be applied to a very difficult, open challenge in the field of quantum machine learning: a robust framework to assess classical and quantum generative models on the same ground. Additionally, to the best of our knowledge, this is the first work that quantitatively compares classical and quantum-inspired models for their generalization capabilities from a practical perspective.

In future work, we are looking to use this approach to evaluate and compare the practical generalization capabilities of alternative models. We see the value in further optimizing the hyperparameters of the GAN architecture, and potentially consider different types of networks such as Recurrent Neural Networks (RNNs) and Variational Autoencoders (VAEs), to push their generalization capabilities. As our framework is tailored towards discrete datasets, we are looking to use this approach in the near future on hybrid and fully quantum generative architectures as well. Previously, it has been a challenge to develop frameworks that can detect generalization in quantum circuits as we are capped with training small-depth circuits~\cite{Zhu2018}. With new meta-learning techniques~\cite{verdon2019learning,Wilson2021,sauvage2021flip} among other pre-training and initialization strategies~\cite{rudolph2022synergistic}, one may be able to train larger quantum circuits and use our approach to evaluate generalization. Additionally, demonstrating generalization capabilities on real quantum hardware would open up interesting questions as to how noise may impact the generalization capabilities of the quantum circuit models. Lastly, we can use this framework as a fair comparison between quantum models and their classical counterparts and we can look into further applications where generalization can deliver commercial value. 

In summary, the most prominent contribution of this work is to introduce and use a framework to unambiguously define and demonstrate generalization-based practical quantum advantage in the generative modeling domain. Generalization is the gold standard for measuring the quality of a machine learning model. With generative modeling having an edge over supervised models in the race for quantum advantage~\cite{PerdomoOrtiz2017}, we hope this work opens the possibility to start this race on a solid ground, and on datasets with commercial relevance~\cite{Ruiz-Torrubiano2010}. As shown here, training GANs and other state-of-the-art classical generative models can be challenging to the point that we report a superior performance from the quantum-inspired generative models used here. Although we expect potentially better results from other classical proposals, there is room as well to improve the quantum-inspired versions explored here. There are also exciting possibilities expected from purely quantum generative models such as Quantum Circuit Born Machines~\cite{Benedetti2019}, as we will be exploring in future work. We hope this work incites both quantum and classical ML experts to use this framework to enhance the performance and design of their models, in this now quantitative race towards demonstrating practical quantum advantage in generative modeling.

\begin{acknowledgments} 
The authors would like to acknowledge Manuel Rudolph, Vladimir Vargas-Calder\'on, Brian Dellabetta, and Dmitri Iouchtchenko for providing in-depth manuscript feedback. Additionally, the authors would like to thank Javier Alcazar, Luis Serrano, Luca Thiede, and Riley Hickman for providing ML expertise and advisement. Lastly, the authors would like to thank Chris Ballance for additional project support, and to recognize the Army Research Office (ARO) for providing funding through a QuaCGR PhD Fellowship. 
\end{acknowledgments}




\bibliography{quantum-ai}


\appendix

\section{Training Details}\label{app:training-details}
 
Here, we provide additional details on the training process for both the quantum-inspired and the classical model. 
The TNBM, whose underlying architecture is an MPS, is trained with a DMRG approach~\cite{han2018unsupervised} with the negative log-likelihood cost function \eqref{product_state}, and the optimization is performed via Stochastic Gradient Descent with learning rate $\eta = 1\text{e-}2$. The number of parameters for the worst case in the TNBM is 1864 for our specific model of $\alpha = 7$. As the bond dimensions for each site are adjusted throughout training, we see that the TNBM does not reach the worst case, and instead has a total number of 1152 parameters. The total number of parameters can be calculated by summing over the squared bond dimensions at each site, and multiplying by a factor of 2. 

In \figref{fig:mps_training_history}, we show the training curves for TNBM with several values of the bond dimension $\alpha$, reporting the KL divergence at each training epoch, that complete the data presented in \secref{sec:pitfalls-mps}. Once more, we stress that we can detect trainability issues with our metrics that are confirmed by the learning curves trends. However, if we consider models that are successfully trained, we expect that our metrics should be able to detect the overfitting and underfitting regime when varying the hyperparameters (e.g. the bond dimension $\alpha$ which controls the TNBM expressivity).

\begin{figure}[t]
\includegraphics[width=\linewidth]{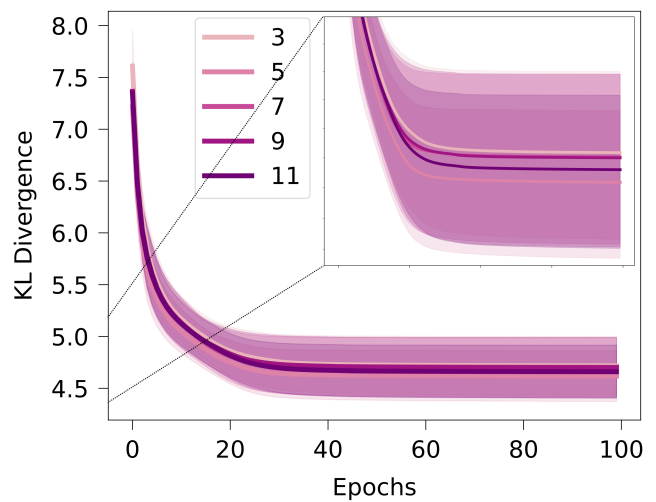}
\caption{\textbf{TNBM training curves for different bond dimensions.} We plot the KL divergence to monitor the training of the TNBM for bond dimensions $\alpha \in \{3, 5, 7, 9, 11\}$. The typical KL value is achieved for $\alpha = 7$ after 100 epochs, thus motivating our choice to utilize this value for further studies and model comparisons. The inset provides a more detailed view of the loss curves ordering. }
\label{fig:mps_training_history}
\end{figure}

In the case of the GAN, the architecture is set to be a feed-forward neural network with linear layers. The generator uses a Gaussian prior, ReLU activation function in the hidden layers and sigmoid cost function in the output layer. The discriminator uses Leaky ReLU activation function in all layers, along with a dropout operation before the final layer. The optimization is performed via the Adam algorithm~\cite{Kingma2014}. The values of the hyperparameters are shown in \tableref{table:GAN-params}. The number of total parameters in the GAN is the sum of the parameters in the discriminator and the generator. For our specific architecture, the number of parameters is  computed in each layer for the discriminator and generator, respectively. For our GAN with 1 hidden layer, we have a total of 4181 parameters.
\begin{table}[h]
\centering
\renewcommand{\arraystretch}{1.3}
\begin{tabular}{||>{\centering}p{3cm} | >{\centering}p{1cm} | >{\centering}p{1.5cm} | >{\centering}p{1cm} ||}
\hline
\multicolumn{1}{||c|}{\textbf{Hyperparameter}}
 & \multicolumn{1}{|c|}{\textbf{GAN}} & \multicolumn{1}{|c|}{\textbf{GAN-MC}} & \multicolumn{1}{|c||}{\textbf{GAN+}}\\
\hline
\multicolumn{1}{||l|}{Prior Size} & \multicolumn{1}{|c|}{20} & \multicolumn{1}{|c|}{8} & \multicolumn{1}{|c||}{12}\\
\hline 
\multicolumn{1}{||l|}{Hidden Size (G)} & \multicolumn{1}{|c|}{20} & \multicolumn{1}{|c|}{6} & \multicolumn{1}{|c||}{6}\\
\hline 
\multicolumn{1}{||l|}{Number of Layers (G)} & \multicolumn{1}{|c|}{1} & \multicolumn{1}{|c|}{4} & \multicolumn{1}{|c||}{1}\\
\hline
\multicolumn{1}{||l|}{Learning Rate (G)} & \multicolumn{1}{|c|}{0.02} & \multicolumn{1}{|c|}{0.051} & \multicolumn{1}{|c||}{0.001}\\
\hline
\multicolumn{1}{||l|}{Hidden Size (D)} & \multicolumn{1}{|c|}{20} & \multicolumn{1}{|c|}{9} & \multicolumn{1}{|c||}{9}\\
\hline
\multicolumn{1}{||l|}{Number of Layers (D)} & \multicolumn{1}{|c|}{1} & \multicolumn{1}{|c|}{3} & \multicolumn{1}{|c||}{1}\\
\hline
\multicolumn{1}{||l|}{Learning Rate (D)} & \multicolumn{1}{|c|}{0.02} & \multicolumn{1}{|c|}{0.008} & \multicolumn{1}{|c||}{0.006}\\
\hline
\multicolumn{1}{||l|}{Negative Slope (D)} & \multicolumn{1}{|c|}{0.02} & \multicolumn{1}{|c|}{0.007} & \multicolumn{1}{|c||}{0.010}\\
\hline
\multicolumn{1}{||l|}{Dropout (D)} & \multicolumn{1}{|c|}{$10^{-5}$} & \multicolumn{1}{|c|}{0.024} & \multicolumn{1}{|c||}{0.107}\\
\hline
\multicolumn{1}{||l|}{Batch Size} & \multicolumn{1}{|c|}{50} & \multicolumn{1}{|c|}{71} & \multicolumn{1}{|c||}{56}\\
\hline
\end{tabular} 
\caption{\textbf{GAN hyperparameter values.} The values labelled with $G$($D$) refer to the generator(discriminator). The hidden size indicates the number of nodes in each hidden layer within $G$ and $D$, approximated to the same significant digit.}
\label{table:GAN-params}
\end{table}

\section{Metrics and Model Behaviours}\label{app:cheat_sheet}
We provide a short guide to what one could expect to see in our metric values $E$ and $(F, R, C)$ when a model exhibits various training behaviours. This `cheat sheet' can be used to quickly check whether the model is perfectly overfitting/memorizing, perfectly generalizing, exhibiting mode collapse in different nuances, or generating too many novel but noisy samples (i.e., anomalous generalization).
\begin{table}[h]
\centering
\renewcommand{\arraystretch}{1.3}
\begin{tabular}{||>{\centering}p{3cm} | >{\centering}p{1cm} | >{\centering}p{1.5cm} | >{\centering}p{1cm} ||}
\hline
\multicolumn{1}{||c|}{\textbf{Model Behaviour}}
 & \multicolumn{1}{|c|}{\textbf{E}} & \multicolumn{1}{|c|}{\textbf{(F, R, C)}} & \multicolumn{1}{|c||}{\textbf{Extra Check}}\\
\hline
\multicolumn{1}{||l|}{Perfect Generalization} & \multicolumn{1}{|c|}{1} & \multicolumn{1}{|c|}{(1, 1, 1)} & \multicolumn{1}{|c||}{N/A}\\
\hline 
\multicolumn{1}{||l|}{Perfect Memorization} & \multicolumn{1}{|c|}{0} & \multicolumn{1}{|c|}{(null, 0, 0)} & \multicolumn{1}{|c||}{$|d_{\text{gen}}| \sim T$}\\
\hline 
\multicolumn{1}{||l|}{Anomalous Pre-Generalization} & \multicolumn{1}{|c|}{$\sim 1$} & \multicolumn{1}{|c|}{(0, 0, 0)} & \multicolumn{1}{|c||}{$|d_{\text{gen}}| \sim T$}\\
\hline
\multicolumn{1}{||l|}{MC (unseen and valid)} & \multicolumn{1}{|c|}{$ \sim 1$} & \multicolumn{1}{|c|}{$(1, 1, \sim 0)$} & \multicolumn{1}{|c||}{N/A}\\
\hline
\multicolumn{1}{||l|}{MC (unseen and invalid)} & \multicolumn{1}{|c|}{$\sim 1$} & \multicolumn{1}{|c|}{(0, 0, 0)} & \multicolumn{1}{|c||}{$|d_{\text{gen}}| << T$}\\
\hline
\multicolumn{1}{||l|}{MC (seen and (in)valid)} & \multicolumn{1}{|c|}{0} & \multicolumn{1}{|c|}{(null, 0, 0)} & \multicolumn{1}{|c||}{$|d_{\text{gen}}| << T$}\\
\hline
\end{tabular} 
\caption{\textbf{Metrics' values across various model behaviours.} The table displays the $E$ and $(F, R, C)$ values one obtains across different model behaviours such as perfect generalization, perfect memorization/overfitting, generating predominantly noise referred to as anomalous pre-generalization, and mode collapsing (MC) on various bitstring types. We see that $F$ will be null in the cases where the number of unseen generated samples is zero. Additionally, we provide an extra check allowing to distinguish between cases in which the generalization metrics yield the same results.}
\label{table:cheat-sheet}
\end{table}

\section{Random sampling as metrics' baseline}\label{app:random-baseline}
To better characterize the performance of the generative models under examination, we compare their generalization capabilities to a simple baseline: we sample randomly from the search space $\mathcal{U}$, thus collecting queries to compute the validity metrics, and we compare the results to the ones associated to the TNBM and the GAN.
The metrics' values are summarized in \tableref{table:baseline}: as expected, both generative models perform better than random sampling, which suggests that during the training process the models were indeed able to learn successfully, despite having different degrees of success. However, the coverage metric in the case of random sampling seems to be higher than the GAN, and this trend persists even considering different numbers of queries $Q$. What motivates this behaviour is the fact that the GAN suffers from mode collapse: its limited diversity impacts the coverage values, whereas the performance of random sampling is favoured by its higher diversity capabilities. However, \figref{fig:baseline-card} shows that the GAN (\figref{sfig:baseline-card-a}) is able to generate more samples in the valid space or its vicinity than the random sampler (\figref{sfig:baseline-card-b}), thus explaining the higher fidelity of the former as opposed to the latter.

\begin{table}[h]
\centering
\renewcommand{\arraystretch}{1.3}
\begin{tabular}{||>{\centering}p{3cm} | >{\centering}p{1.5cm} | >{\centering}p{1.5cm} | >{\centering}p{1.5cm} ||}
\hline
\multicolumn{1}{||c|}{\textbf{Metric}} & \multicolumn{1}{|c|}{\textbf{TNBM}} & \multicolumn{1}{|c|}{\textbf{GAN}} & \multicolumn{1}{|c|}{\textbf{Random}} \\
\hline
\multicolumn{1}{||c|}{$E$} & \multicolumn{1}{|c|}{$0.989 (0.02\%)$} &\multicolumn{1}{|c|}{$0.995 (0.02\%)$} & \multicolumn{1}{|c|}{$0.998 (0.013\%)$} \\
\hline
\multicolumn{1}{||c|}{$F$} & \multicolumn{1}{|c|}{$0.989 (0.03\%)$} &\multicolumn{1}{|c|}{$0.263 (0.6\%)$} & \multicolumn{1}{|c|}{$0.17 (0.50\%)$} \\
\hline
\multicolumn{1}{||c|}{$R$}& \multicolumn{1}{|c|}{$0.978 (0.03\%)$} &\multicolumn{1}{|c|}{$0.261 (0.6\%)$} & \multicolumn{1}{|c|}{$0.17 (0.50\%)$} \\
\hline
\multicolumn{1}{||c|}{$C$} & \multicolumn{1}{|c|}{$0.409 (0.15\%)$} &\multicolumn{1}{|c|}{$0.006 (1.7\%)$} & \multicolumn{1}{|c|}{$0.09 (0.48\%)$} \\
\hline
\end{tabular}
\caption{\textbf{Pre-generalization and validity-based generalization metrics.} We display the average exploration $E$ and the average $(F, R, C)$ values for each best model run with an average and the associated relative percentage error across 15 query batches. Both the TNBM and the GAN achieve better performance than the random sampler for all the different metrics, except for the GAN coverage as pointed out in the main text.}
\label{table:baseline}
\end{table}

\begin{figure}[!h]
\centering
\subfloat[\label{sfig:baseline-card-a}]{
  \includegraphics[width=\linewidth]{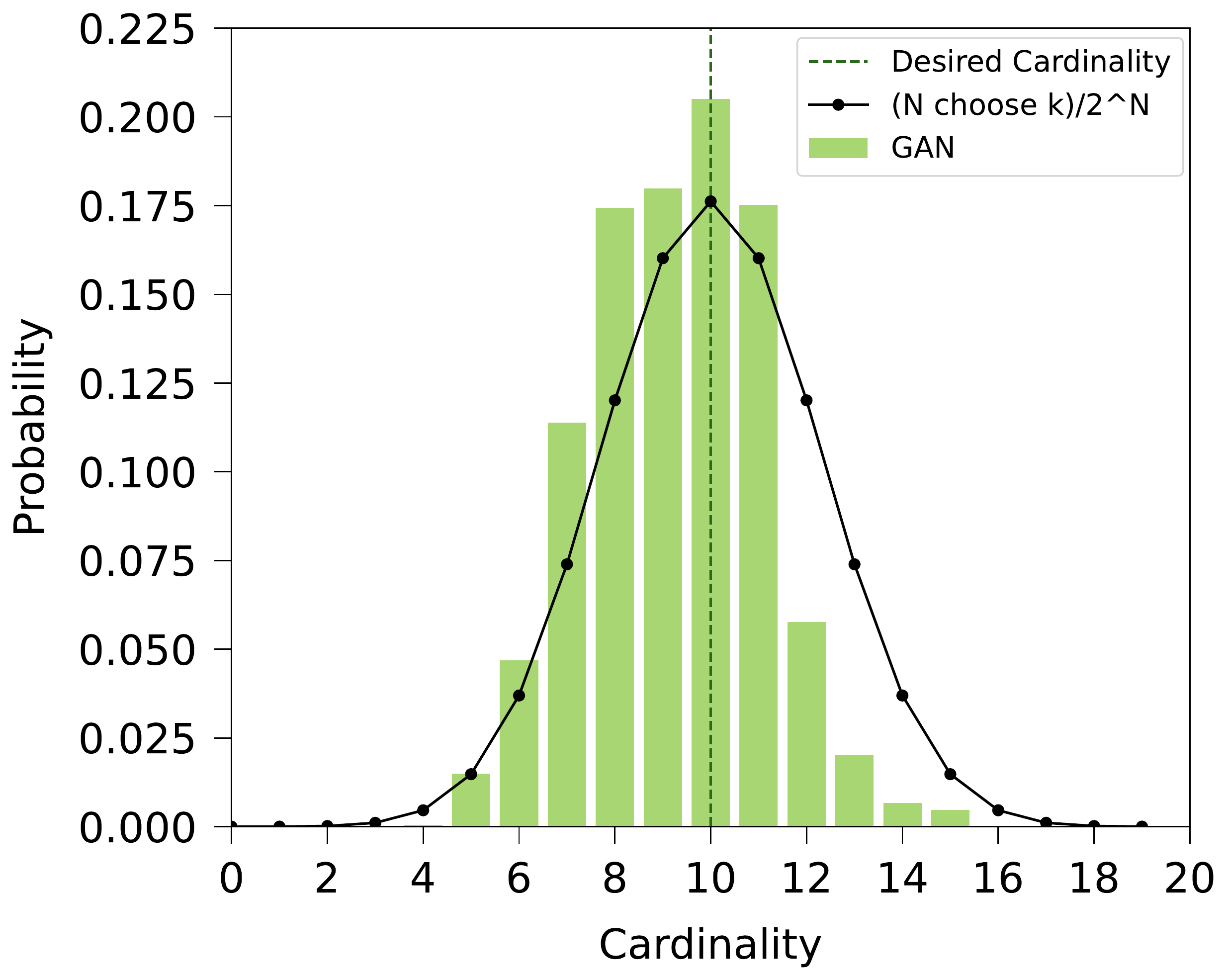}}
\\
\subfloat[\label{sfig:baseline-card-b}]{%
  \includegraphics[width=\linewidth]{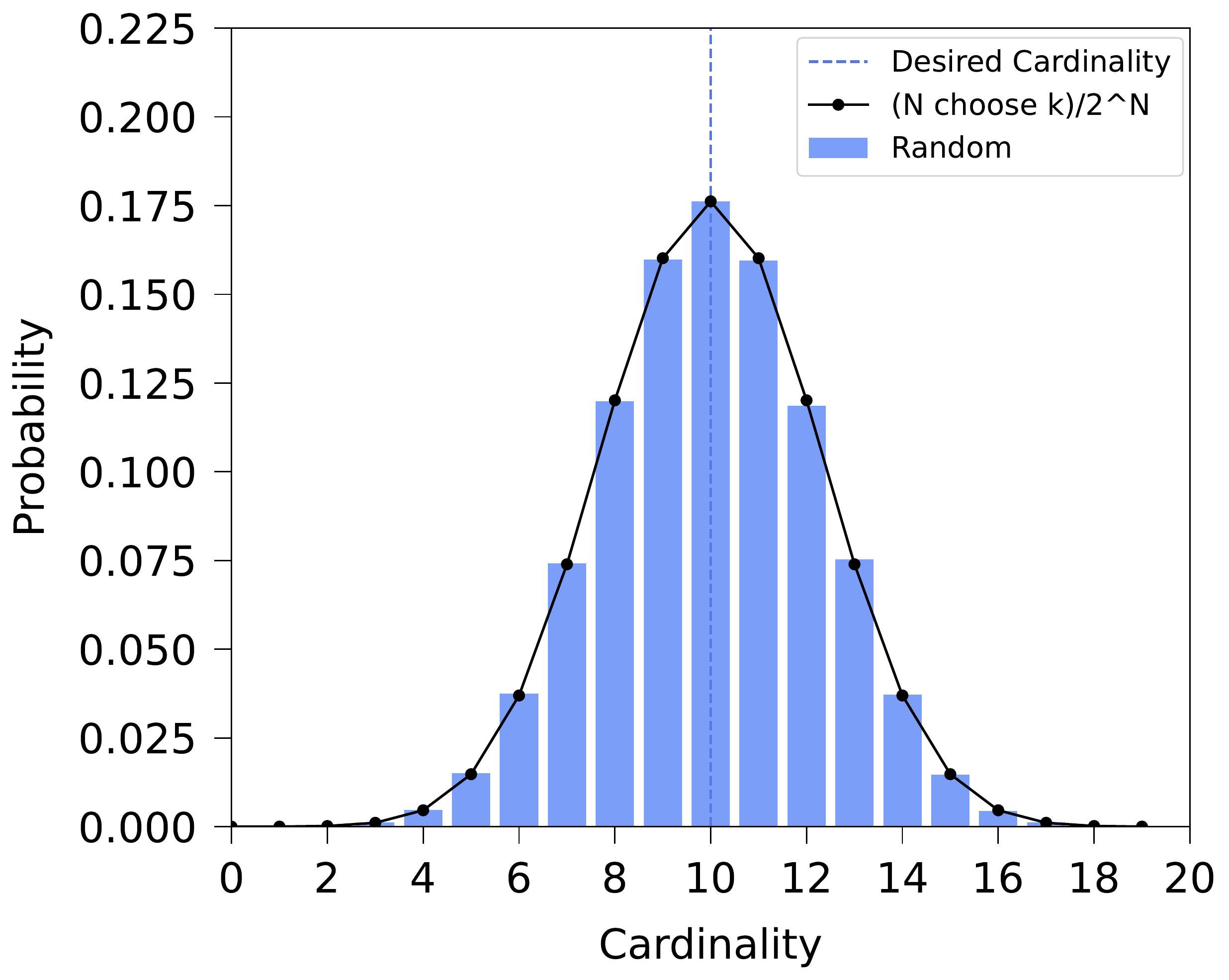}%
}
\caption{\textbf{Cardinality distribution for GAN and random sampler.} The plots show the percentage of queries with different cardinalities generated by the GAN (\figref{sfig:baseline-card-a}) and by the random sampler (\figref{sfig:baseline-card-b}). We notice that the GAN is able to produce a higher number of queries with the correct cardinality $k=10$ (or its vicinity), thus showing that the training process allowed the GAN to partially learn the validity pattern in the training dataset. The black line represents the probability to draw a query with a given cardinality when randomly sampling from the search space $\mathcal{U}$. }
\label{fig:baseline-card}
\end{figure}

\clearpage
\section{Metrics' Trends}\label{app:metric_trends}
\vspace{-2.5cm}
To further demonstrate the power and stability of our metrics, we provide additional details regarding how they scale as we vary the number of queries $Q$ generated from the trained model. Specifically, in Figures \ref{fig:fid-trend}-\ref{fig:uti-trend}, we plot the values of the validity-based and quality-based generalization metrics and show that most of them do not change with the number of queries - except for coverage, as already shown in \figref{fig:cov-trend}, and for the minimum value that is displayed in \figref{fig:mv-trend}. The validity-based trend plots display the constant behaviour of the metrics for both TNBM and GAN as $Q$ increases, along with a dashed black line indicating the ideal metrics value of 1. 
The quality-based trend plots display the constant behaviour of the utility metric for both TNBM and GAN as $Q$ increases, and a decreasing behaviour for the minimum value as $Q$ increases. The latter is the expected trend: with more queries one has a higher probability of reaching a sample with a lower cost value. For both of these plots, we include a dashed black line indicating the training threshold. This data supports our claim that while our metrics are sample-based, most of them are not dependent on the number of queries.
\vspace{-2.5cm}
\begin{figure}
\includegraphics[width=\linewidth]{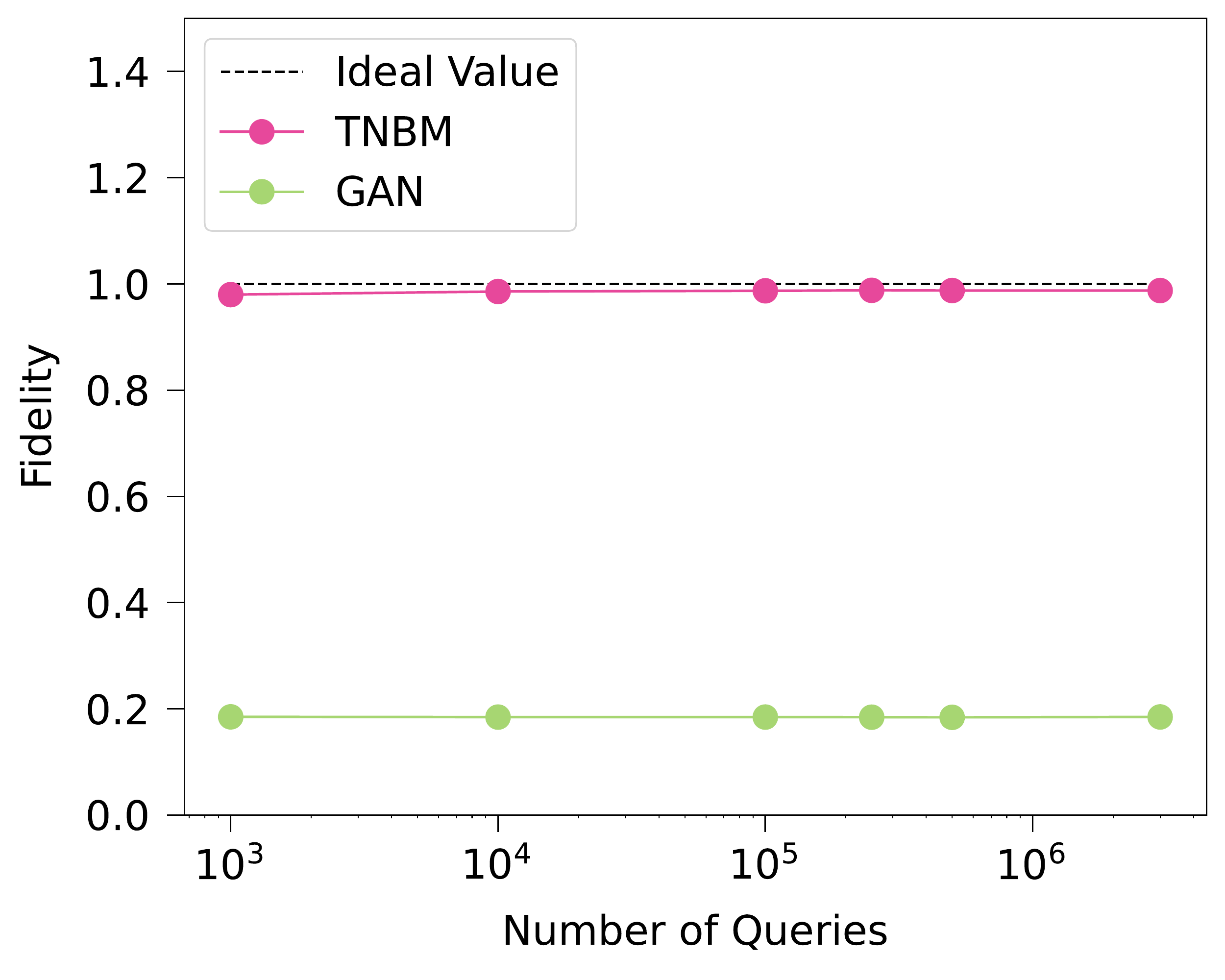}
\caption{\textbf{Fidelity trends for increasing number of queries.} The plot displays the constant behaviour of the fidelity $F$ for both TNBM (pink) and GAN (green) as we increase the number of queries $Q$ retrieved from the trained models. The dashed black line shows the ideal metric value of 1. In both models, the fidelity is independent of the number of generated queries.}
\label{fig:fid-trend}
\end{figure}

\begin{figure}
\includegraphics[width=\linewidth]{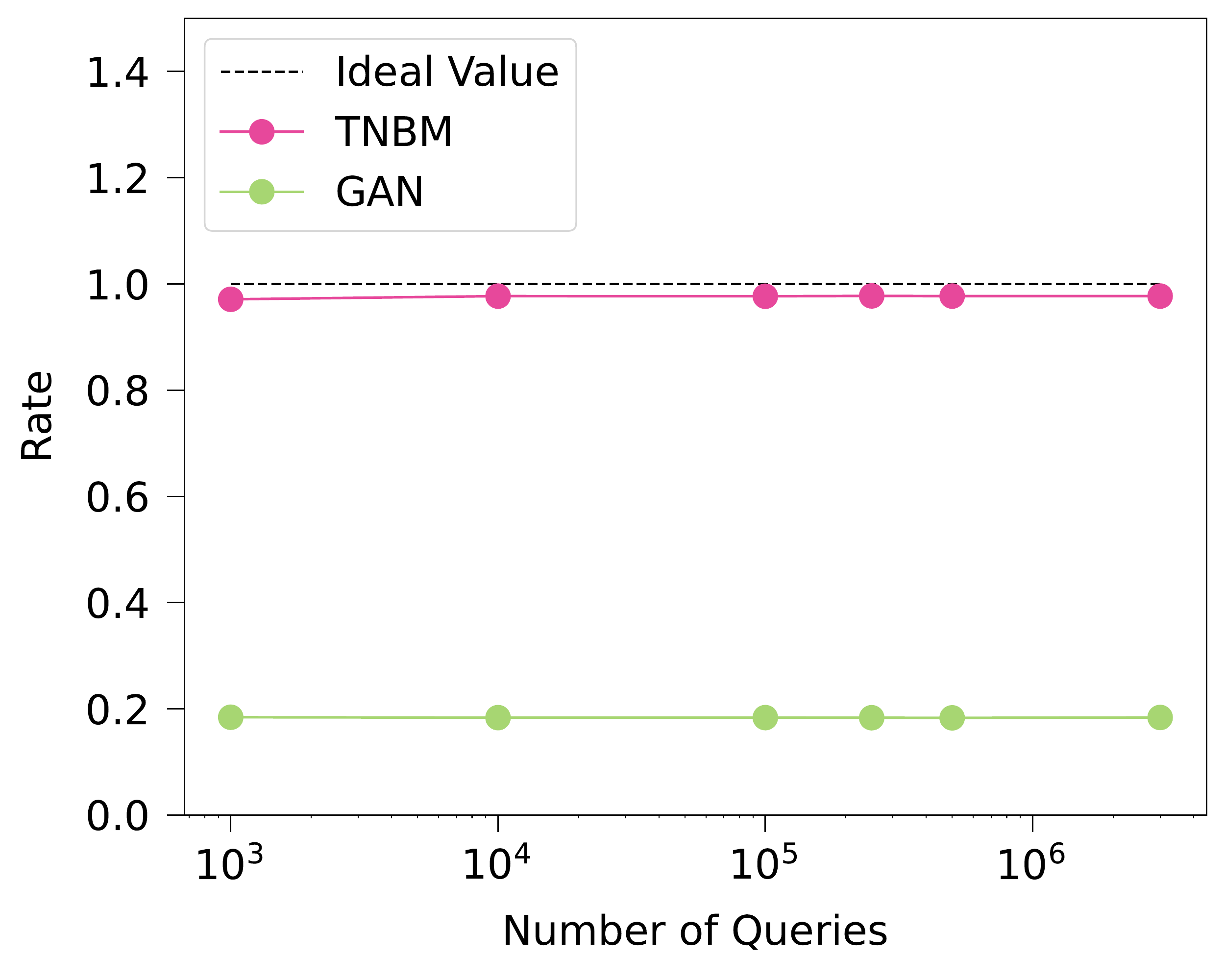}
\caption{\textbf{Rate trends for increasing number of queries.} The plot displays the constant behaviour of the rate $R$ for both TNBM (pink) and GAN (green) as we increase the number of queries $Q$ retrieved from the trained models. The dashed black line shows the ideal metric value of 1. In both models, the $R$ is independent of the number of generated queries.}
\label{fig:rate-trend}
\end{figure}
\begin{figure}
\includegraphics[width=\linewidth]{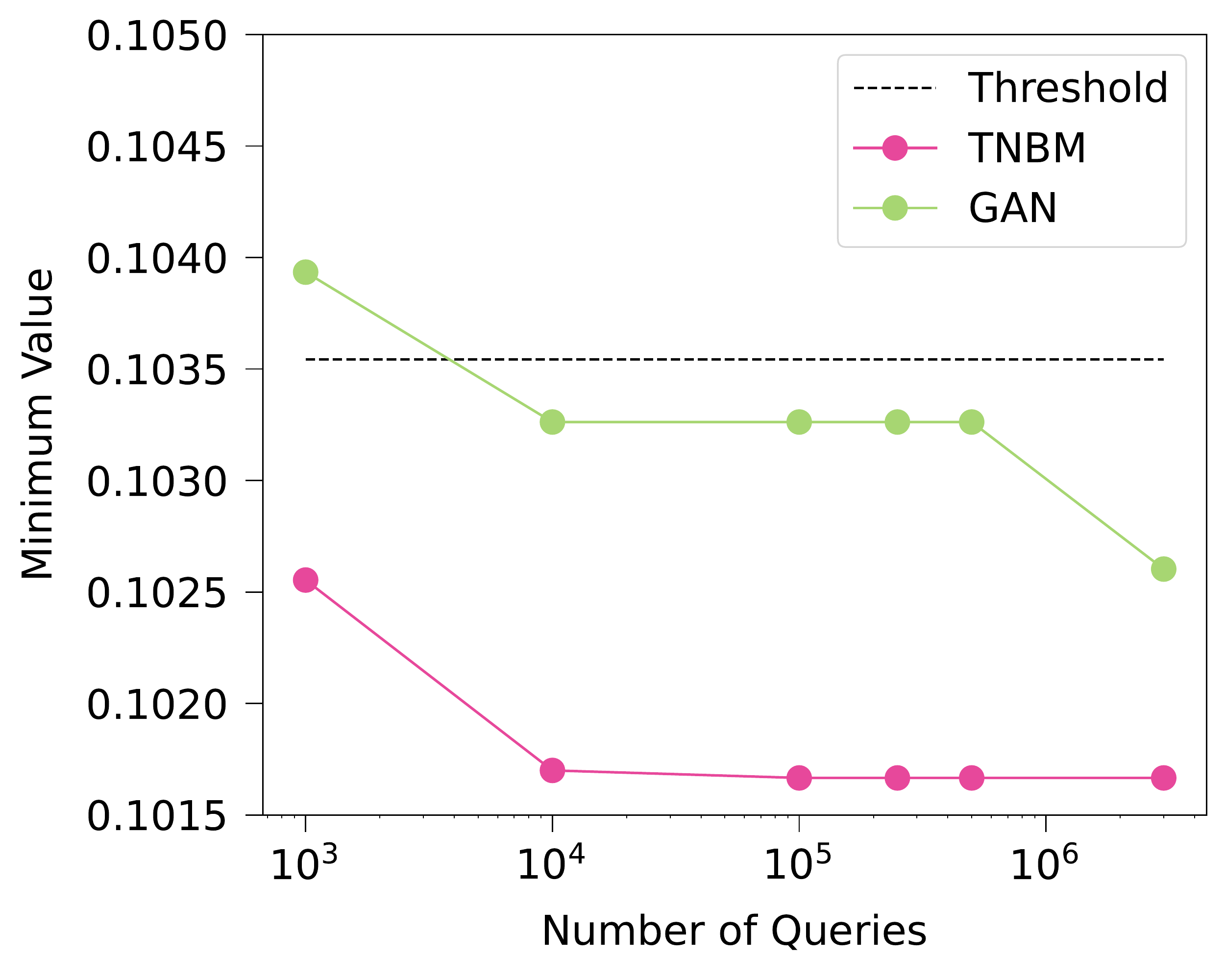}
\caption{\textbf{Minimum Value ($MV$) trends for increasing number of queries.} The plot displays the decreasing behaviour of the $MV$ metric for both TNBM (pink) and GAN (green) as we increase the number of queries $Q$ retrieved from the trained models. The dashed black line shows the minimum value of the training set $MV$. Note that both models not only produce unseen valid samples, but also samples with better quality than those in the training set. As we increase $Q$, both models are more likely to produce a query with a lower cost value, even if the GAN requires more samples than the TNBM to dip under the threshold.}
\label{fig:mv-trend}
\end{figure}
\begin{figure}[H]
\includegraphics[width=\linewidth]{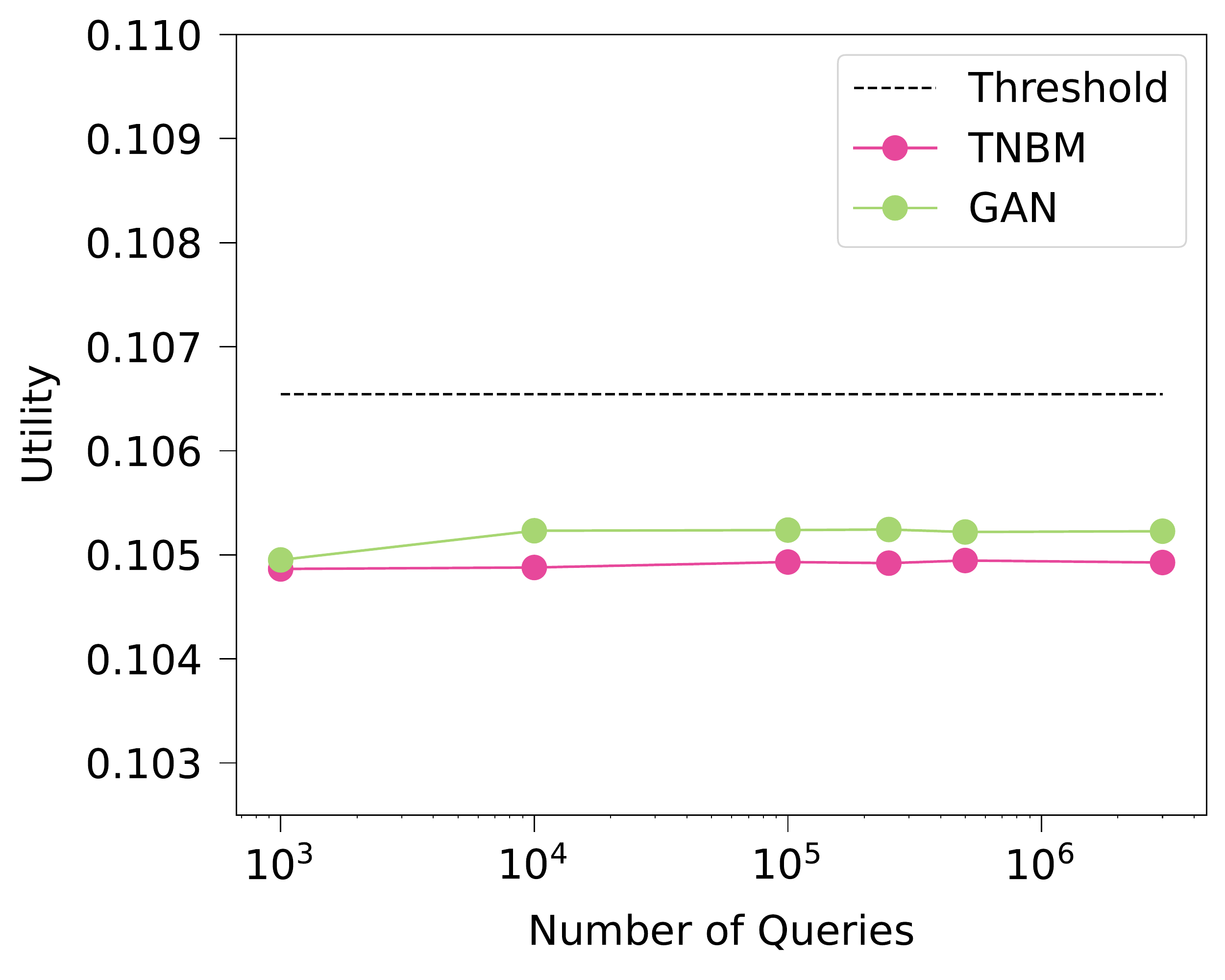}
\caption{\textbf{Utility trends for increasing number of queries.} The plot displays the constant behaviour of the utility $U$ for both TNBM (pink) and GAN (green) as we increase the number of queries $Q$ retrieved from the trained models. The dashed black line shows the threshold value of the training set $U$. Both the GAN and TNBM remain under the threshold, independent of the number of queries.}
\label{fig:uti-trend}
\end{figure}
We further propose an investigation on the stability of our approach across various training datasets $\mathcal{D}_{\text{Train}}$. Since a training dataset contains a subset of samples of size $T$ drawn from the solution space $\mathcal{S}$, it is possible to build different datasets from the same problem instance by randomizing this samples-drawing procedure.

We present the raw data of each of our metrics obtained using 10 distinct datasets built from the same fixed problem instance. Thus, all the datasets share the same asset universe,
cardinality, and seen portion $\epsilon$ as stated in \secref{sec:sim-det}, and they simply differ for the training bitstrings that get sampled from $P(x)$. Tables \ref{table:data_instances_validmps}-\ref{table:data_instances_qualgan} show the results we obtained for the different pre-generalization condition, validity-based and value-based generalization metrics across the 10 different datasets, where each line corresponds to one dataset. We see that the TNBM beats the GAN for all $(F, R, C)$ values. The relative percentage errors across the datasets for $(F, R, C)$ values are smaller for the TNBM $(0.5\%,0.5\%,0.5\%)$ than the GAN $(13\%,13\%,30\%)$, demonstrating that the TNBM produces more stable results across datasets. However, both standard deviations are small enough to show that our metrics produce similar results across various training data. 

\begin{table}
\centering
\renewcommand{\arraystretch}{1.3}
\begin{tabular}{||c|c|c|c||}
\hline
$E$ & $F$ & $R$ & $C$ \\
\hline
0.989 & 0.982 & 0.971 & 0.405\\
\hline 
0.989 & 0.978 & 0.968 & 0.406 \\
\hline 
0.989 & 0.971 & 0.961 & 0.401 \\
\hline
0.989 & 0.984 & 0.973 & 0.407 \\
\hline
0.989 & 0.983 & 0.973 & 0.406 \\
\hline 
0.989 & 0.985 & 0.975 & 0.407 \\
\hline
0.989 & 0.978 & 0.967 & 0.405 \\
\hline 
0.989 & 0.977 & 0.967 & 0.404 \\
\hline
0.989 & 0.987 & 0.977 & 0.406 \\
\hline 
0.989 & 0.987 & 0.977 & 0.409 \\
\hline 
\end{tabular} 
\caption{\textbf{TNBM pre-generalization and validity-based generalization metrics' values across multiple training datasets from the same problem instance}. We see that the metrics have similar values across the 10 datasets under examination with relative percentage errors $(0.5\%,0.5\%,0.5\%)$ for $(F, R, C)$ values respectively. Thus, our metrics produce similar values across multiple training datasets, demonstrating that they are independent of the portion of training samples selected from the valid space.}
\label{table:data_instances_validmps}

\end{table}
\begin{table}[h]
\centering
\renewcommand{\arraystretch}{1.3}
\begin{tabular}{||c|c|c|c||}
\hline
$E$ & $F$ & $R$ & $C$ \\
\hline
0.999 & 0.249 & 0.249 & 0.0062\\
\hline 
0.996 & 0.236 & 0.235 & 0.0062\\
\hline
0.996 & 0.309 & 0.307 & 0.0063\\
\hline
0.998 & 0.233 & 0.233 & 0.0042\\
\hline
0.995 & 0.181 & 0.179 & 0.0049\\
\hline 
0.999 & 0.232 & 0.232 & 0.0061\\
\hline 
0.997 & 0.274 & 0.274 & 0.0110\\
\hline 
0.997 & 0.276 & 0.275 & 0.0071\\
\hline
0.999 & 0.239 & 0.239 & 0.0066\\
\hline 
0.994 & 0.251 & 0.249 & 0.0077\\
\hline 
\end{tabular} 
\caption{\textbf{GAN pre-generalization and validity-based generalization metrics' values across multiple training datasets from the same problem instance}. We see that the metrics have similar values across the 10 datasets under examination with mean standard deviations $(13\%,13\%,30\%)$ for $(F, R, C)$ values respectively. Despite not being nearly as stable as the TNBM, we see that our metrics produce similar values across multiple training datasets, demonstrating that they independent of the portion of training samples selected from the valid space.} 
\vspace{1.5cm}
\label{table:data_instances_validgan}

\begin{tabular}{||c|c|c|c||}
\hline
$U$ & $U_T$ & $MV$ & $MV_T$ \\
\hline
0.1049  & 0.1064 & 0.1017 & 0.1018\\
\hline
0.1049  & 0.1065 & 0.1017 & 0.1034\\
\hline
0.1048 & 0.1067 & 0.1017 & 0.1018\\
\hline
0.1049  & 0.1064 & 0.1017 & 0.1031\\
\hline
0.1047  & 0.1062 & 0.1017 & 0.1033\\
\hline
0.1049  & 0.1065 & 0.1017 & 0.1027\\
\hline
0.1051  & 0.1068 & 0.1017 & 0.1036\\
\hline
0.1049  & 0.1065 & 0.1017 & 0.1029\\
\hline
0.1048  & 0.1064 & 0.1017 & 0.1039\\
\hline
0.1049  & 0.1062 & 0.1017 & 0.1021\\
\hline
\end{tabular} 
\label{table:data_instances_qualmps}
\caption{\textbf{TNBM quality-based metrics' values across various training datasets from the same problem instance.} The second and last columns display the values for the training set, defined as the $U$ and the $MV$ computed for the samples in $\mathcal{D}_{\text{Train}}$. We see that the TNBM's $U$ and $MV$ is always less than the training threshold. Additionally, the same low $MV$ value that exists in the fixed problem universe is generated independent of the training set.}
\end{table}

For the quality-based metrics, we see that the $MV$ for the TNBM is always either equal or less than that of the GAN. However, for $U$, the TNBM and the GAN trade-off in being the winner. This is not a surprise, as in \tableref{table:utility} the TNBM and the GAN produced very similar values for the utility. The same argument from \secref{sec:model_evaluation} holds such that both the TNBM and GAN are able to generate low cost samples. Simply, the TNBM contains more diversified high quality samples, which is not captured by the metric $U$. 
\begin{table}[H]
\centering
\renewcommand{\arraystretch}{1.3}
\begin{tabular}{||c|c|c|c||}
\hline
$U$ & $U_T$ & $MV$ & $MV_T$ \\
\hline
0.1041 & 0.1064 & 0.1032 & 0.1018\\
\hline
0.1040 & 0.1065 & 0.1021 & 0.1034\\
\hline
0.1042 & 0.1067 & 0.1019 & 0.1018\\
\hline
0.1038  & 0.1064 & 0.1019 & 0.1031\\
\hline
0.1029  & 0.1062 & 0.1017 & 0.1033\\
\hline
0.1044  & 0.1065 & 0.1018 & 0.1027\\
\hline
0.1043  & 0.1068 & 0.1028 & 0.1036\\
\hline
0.1048  & 0.1065 & 0.1024 & 0.1029\\
\hline
0.1044  & 0.1064 & 0.1017 & 0.1039\\
\hline
0.1056  & 0.1064 & 0.1038 & 0.1021\\
\hline
\end{tabular} 
\label{table:data_instances_qualgan}
\caption{\textbf{GAN quality-based metrics' values across various training datasets from the same problem instance.} The second and last columns display the values for the training set, defined as the $U$ and the $MV$ computed for the samples in $\mathcal{D}_{\text{Train}}$. We see that the GAN's $U$ is always less than the training threshold; however, this is not always true for $MV$, as the GAN has a lower $MV$ value only $70\%$ of the time.}
\end{table}
An additional analysis on the stability of the different generative models would be the investigation of their generalization capabilities across different problem instances, especially the ones characterized by larger asset universes, e.g. $N=500$ (which would correspond to all the assets in the S\&P500 index). We highlight here that our approach is not limited to the relatively small universe size considered in this work, i.e. $N=20$, that was chosen to allow for a practically feasible comparison with quantum generative models in the near term.
\newpage
\section{Supplementary Figures}\label{app:supp_figs}
We include supplementary figures to further demonstrate some of our results. Specifically, in \figref{sfig:2D-Viz} we provide 2D visualizations of the data distribution (\figref{sfig:2D-Viz-a}), the training distribution (\figref{sfig:2D-Viz-b}), and the output distributions of the trained TNBM (\figref{sfig:2D-Viz-c}) and GAN (\figref{sfig:2D-Viz-d}) for a $N=20,\, k=10$ problem instance. In the 2D image, every pixel is associated to one of the $2^N$ bitstrings in the search space $\mathcal{U}$, and its color encodes the associated probability value. We can see that the bi-dimensional representation of the data distribution displays a non-trivial pattern defined by the solution space $\mathcal{S}$. Remarkably, provided the small amount of samples that do not demonstrate a very clear pattern in the training distribution, the TNBM and GAN are able to learn the unknown correlations: in particular, the TNBM is able to almost perfectly infer the patterns in the data distribution from very little information. This result is in alignment with our findings in \secref{sec:metric_robustness} that suggest that TNBMs are able to infer the ground truth distribution from few training data, as indicated by the value of $KL_{\text{Target}}$.

In \figref{fig:GAN-value-based}, we provide a visualization of the GAN quality-based generalization metrics in analogy to \figref{fig:TNBM_utility}. By comparing the two plots, we can see that both models reach the low-risk section of the spectrum, but the TNBM samples exhibit more diversity than the GAN ones.

In \figref{fig:training-stability} we display a comparison of the training stability of TNBM and all the three GANs considered in this work, showing how good each of the model is in capturing the correct cardinality pattern encoded in the dataset. We can detect the higher instability affecting GAN models, as opposed to the MPS performance, which appears remarkable. We highlight that even if the TNBM produces only queries with a given cardinality, similar to the GAN-MC histograms, the quantum-inspired model is not exhibiting mode collapse onto an unseen and valid bitstring, as the coverage is not negligible as in the GAN-MC case (see \tableref{table:comparison}).

In \figref{f:stat_approach}, we showcase the full data from which values in \tableref{table:valid_learning_theory} are extracted. Here, we demonstrate the average $(F, R, C)$ values throughout five independent TNBM trainings of $\alpha = 7$. We show that with each training iteration, the metrics improve towards optimal $(F, R, C)$ values. To complement this information, we also plot the KL divergence of the model's output distribution relative to the unknown data distribution $(KL_{\text{Target}})$ as well as the model's output distribution relative to the training distribution $(KL_{\text{Train}})$ for $\epsilon \in \{0.01, 0.5, 1.0$\}. As discussed in \secref{review}, according to computational learning theory, a model is able to generalize well if it can successfully infer the ground truth data distribution given the amount of training data available: we encode this information in $KL_{\text{Target}}$, even though other metrics may be used for the same purpose, such as the Total Variation Distance~\cite{hinsche2022}. As we see that the $KL_{\text{Target}} < KL_{\text{Train}}$ across $\epsilon$ values, we can conclude that, in various training data regimes, the model is able to output data that is closer to the ground truth than the training data provided. We showcase the KL divergence alongside the $(F, R, C)$ values during training to emphasize that our metrics agree with proposed evaluation schemes in computational learning theory discussed in \secref{review}. 

\begin{figure*}[bht]
\subfloat[\label{sfig:2D-Viz-a}]{%
  \includegraphics[width = 0.5\linewidth]{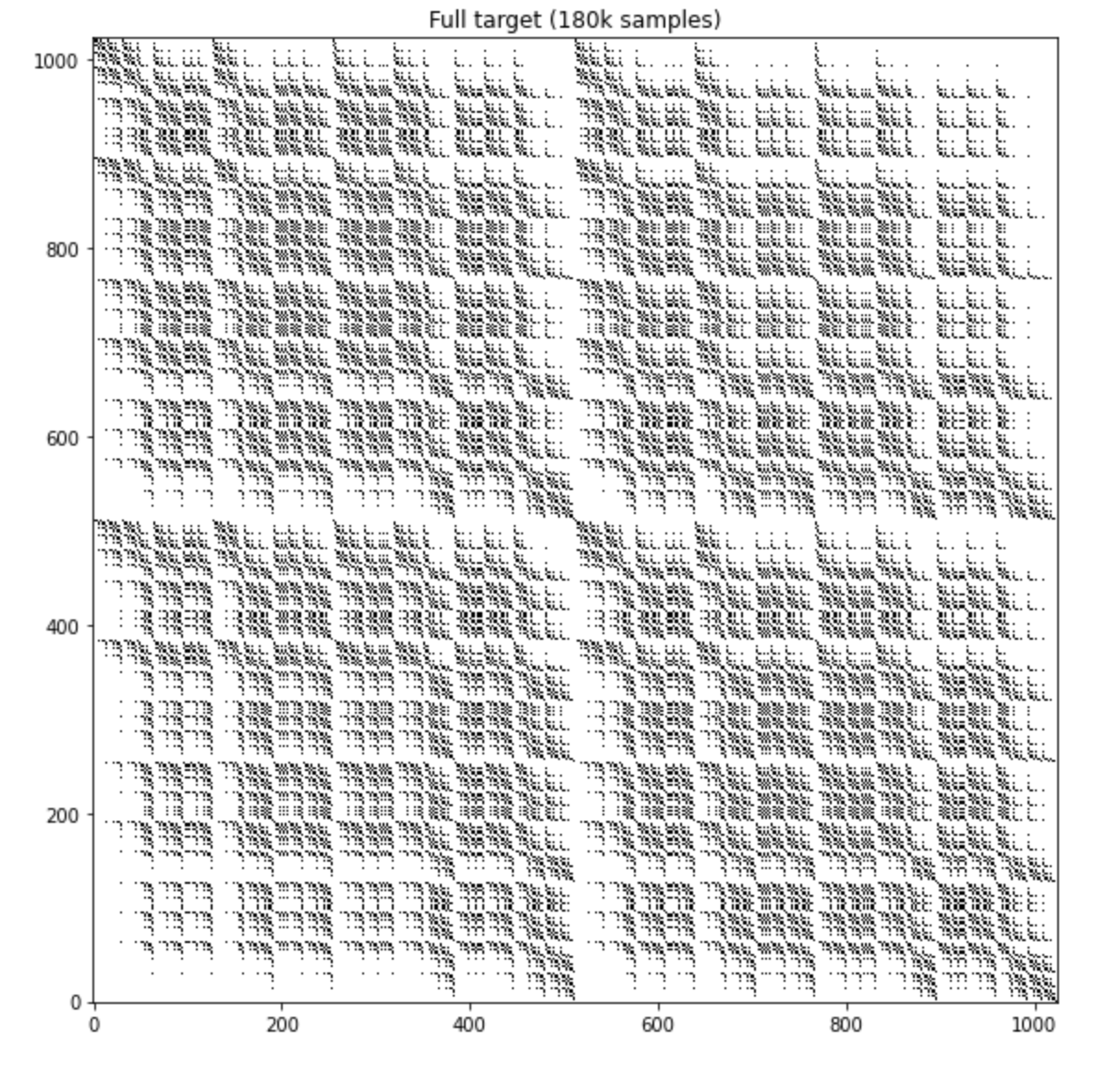}%
}
\subfloat[\label{sfig:2D-Viz-b}]{%
  \includegraphics[width = 0.5\linewidth]{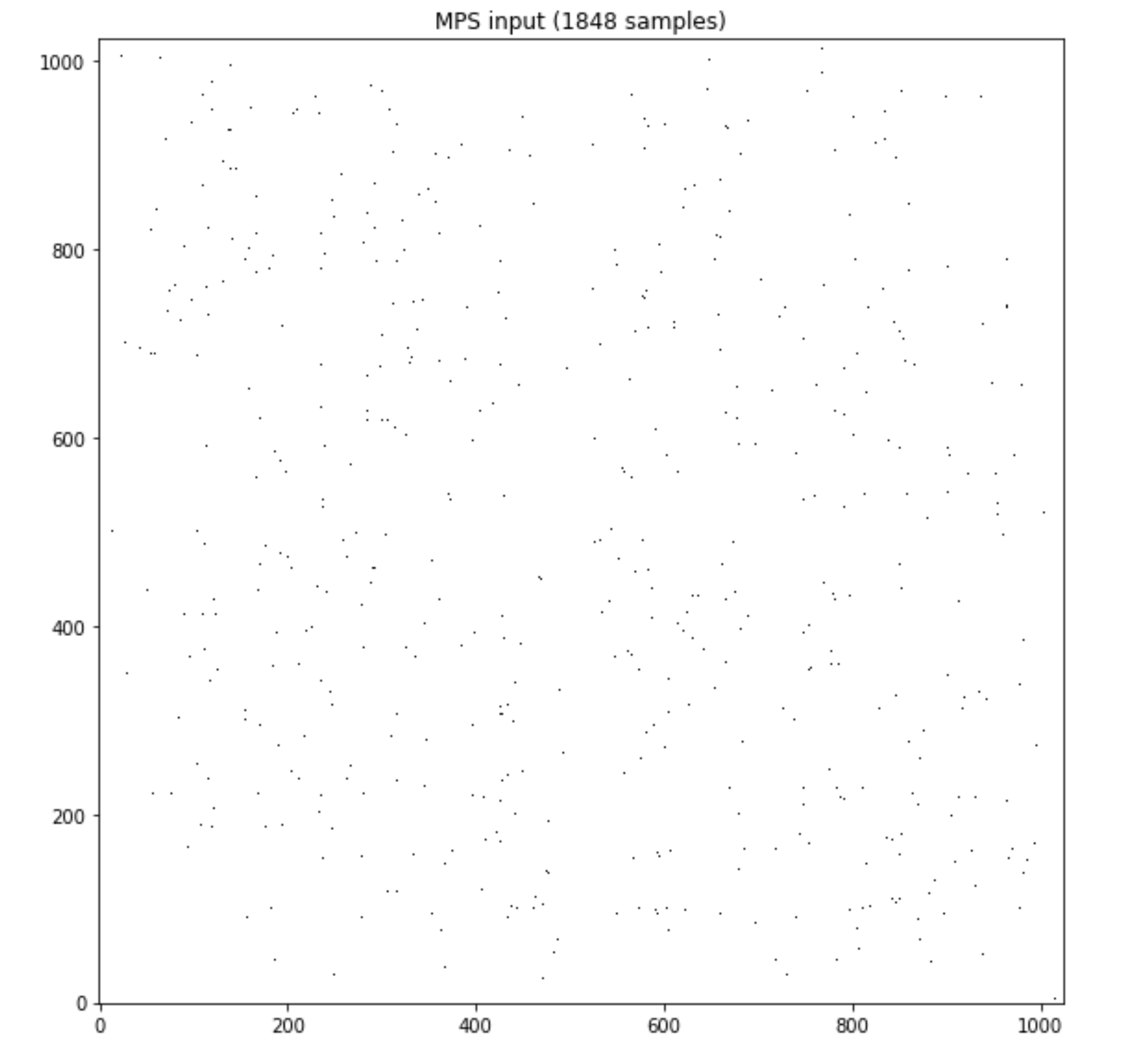}%
}\\
\subfloat[\label{sfig:2D-Viz-c}]{%
  \includegraphics[width = 0.5\linewidth]{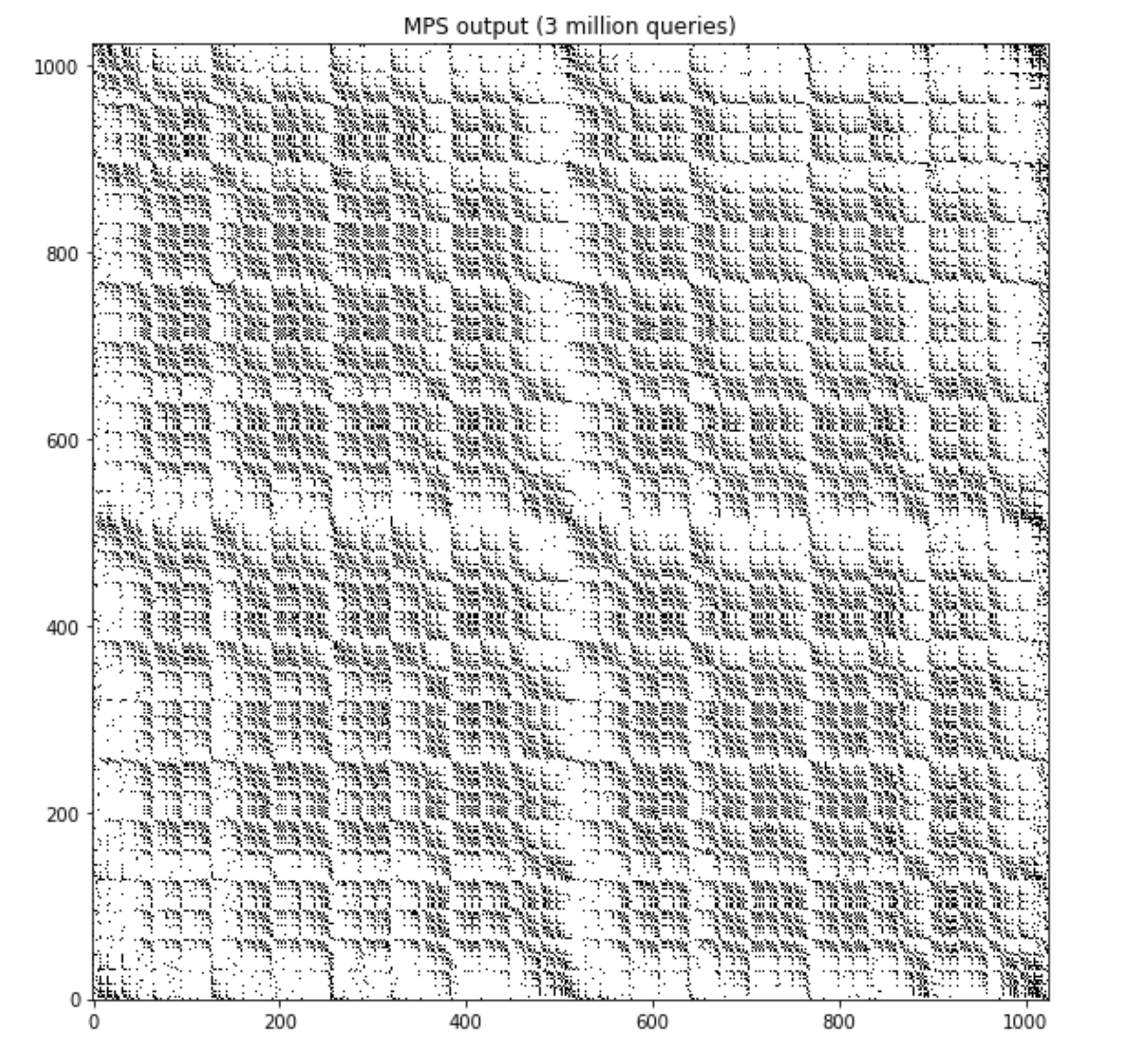}%
}
\subfloat[\label{sfig:2D-Viz-d}]{%
  \includegraphics[width = 0.5\linewidth]{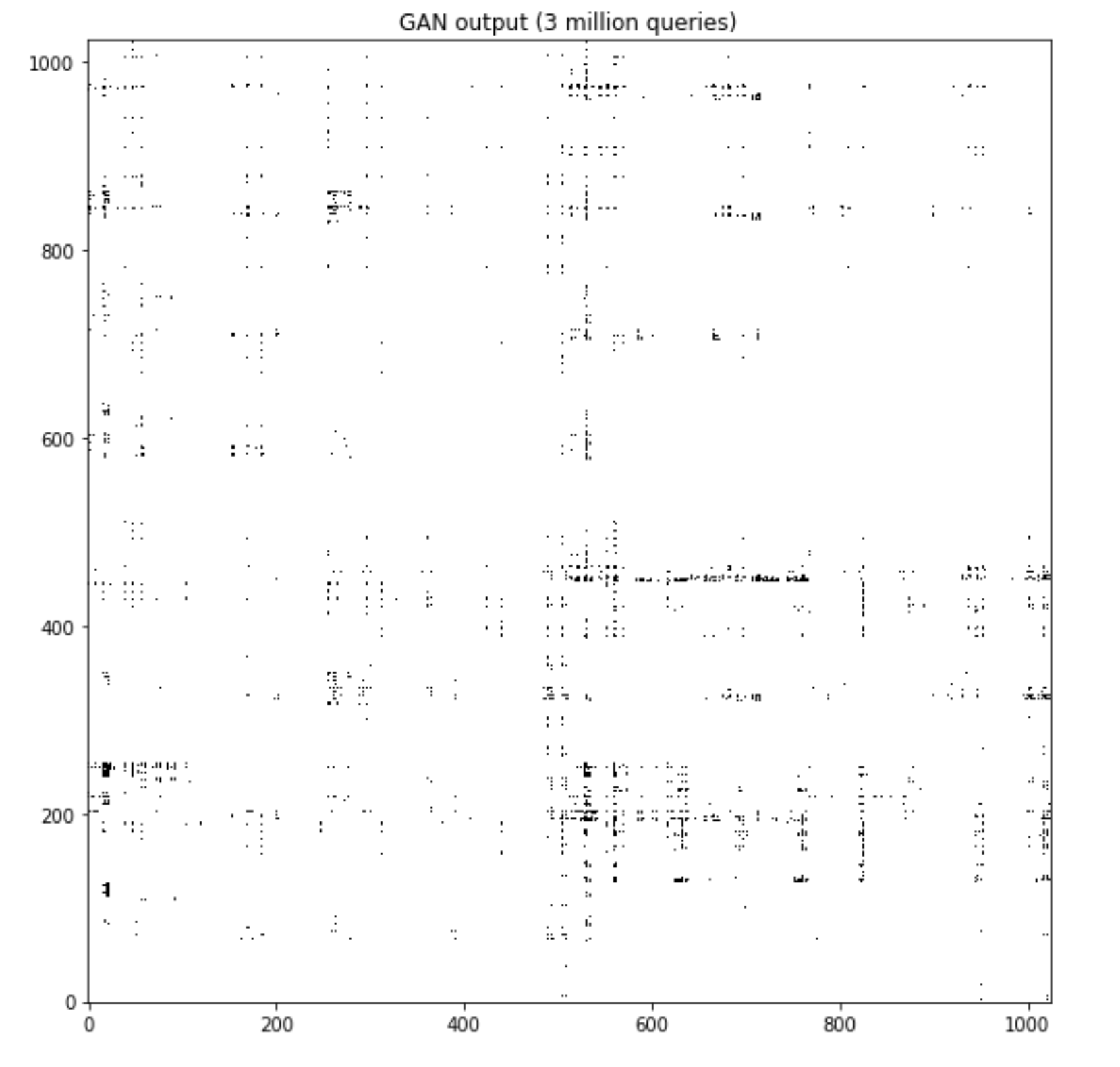}%
}
\caption{\textbf{2D Visualization of distributions.} \figref{sfig:2D-Viz-a} shows the 2D visualization of the exact data distribution defined by the solution space $\mathcal{S}$, where we see that a specific pattern emerges from the cardinality. In \figref{sfig:2D-Viz-b}, we display the 2D visualization for the training distribution, where the same distribution was given to both the TNBM and the GAN models. As shown in \figref{sfig:2D-Viz-c}, it is very remarkable that with this very limited number of training patterns provided to each model, the TNBM is able to generate the pattern from the data distribution almost exactly (as reflected in the metric values too). On the contrary, in \figref{sfig:2D-Viz-d} we see that while the GAN is able to learn portions of the pattern, it struggles to reproduce this data distribution.}
\label{sfig:2D-Viz}
\end{figure*}

\begin{figure}[h]
\includegraphics[width=\linewidth]{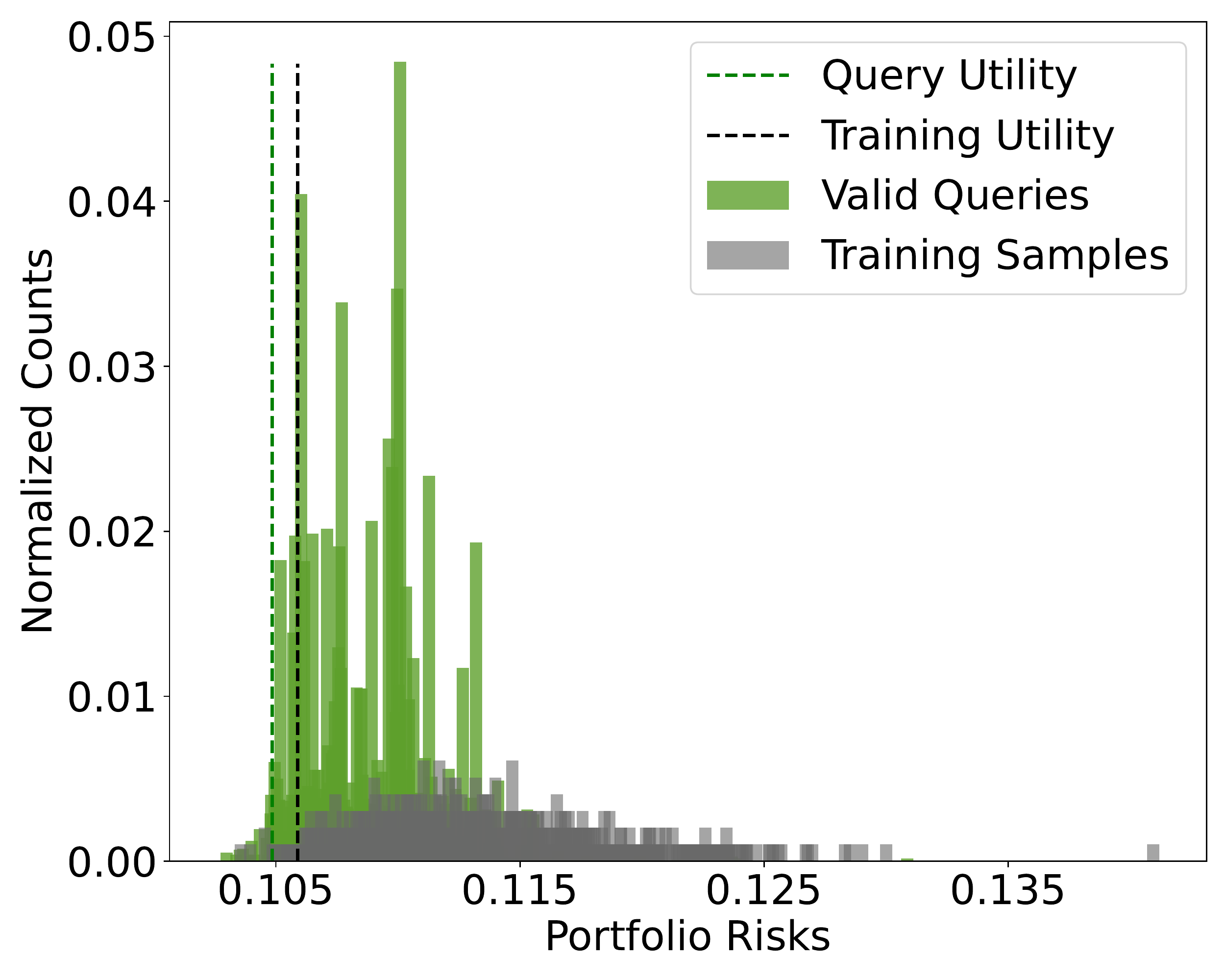}
\caption{\textbf{Visualization of quality-based metrics for GAN-generated queries.} The plot displays the number of portfolio counts associated to given risk values. The green spikes represent valid GAN queries, whereas the gray spikes represent the samples from the training set. Note that for calculating our metrics, we used $Q=10^5$ queries, but the training distribution only contains 1,848 samples, hence the need for normalizing the histograms. Here, we set the utility threshold to $t = 5\%$. Similar to the TNBM, the model distribution learns the low-risk `bias' encoded in the training set, and generates more values of low risk. However, unlike the TNBM, the model frequency counts per query are higher, and the sample diversity is quite low. The queries have a lower utility (green dashed) than the training set (black dashed), thus meeting the condition in \eqref{utility}. Ultimately, no matter the query count, we see that the GAN can reach low risk queries, but simply has less diversity among them in contrast to the TNBM.}
\label{fig:GAN-value-based}
\end{figure}

\begin{figure*}
\includegraphics[width=\linewidth]{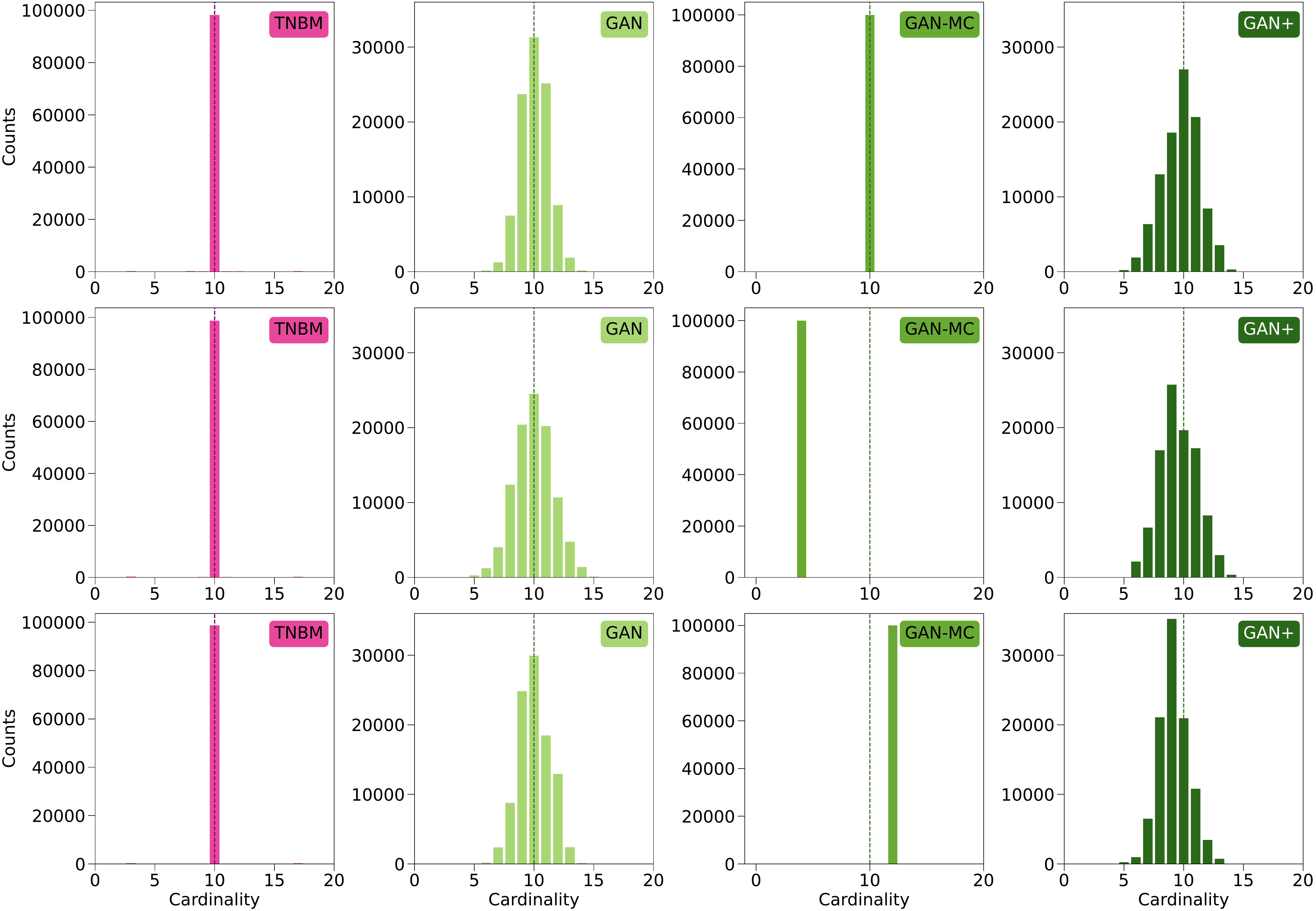}
\caption{\textbf{Cardinality distributions of queries generated by multiple models during independent trainings.} We represent cardinality histograms obtained when taking $Q=10^5$ queries from three independently trained instances of each model family (TNBM, GAN, GAN-MC, GAN+). Each plot displays the cardinality distribution of the retrieved queries, along with the desired cardinality $k=10$. We can see that the three TNBM models generate queries that always learn accurately the cardinality constraint, whereas the GAN models show less training stability, which is known to be one of the issues affecting this class of classical generative models. Specifically, for GAN and GAN+ we see that while each model always produces at least some valid queries, the centers and tails of the distributions vary greatly for each instance. For GAN-MC, distinct trainings collapse onto different cardinalities, implying that the model is not always guaranteed to generate valid queries.}
\label{fig:training-stability}
\end{figure*}

\begin{figure*}[bht]
\subfloat[\label{stat-kl_divergence}]{%
  \includegraphics[width = 0.5\linewidth] {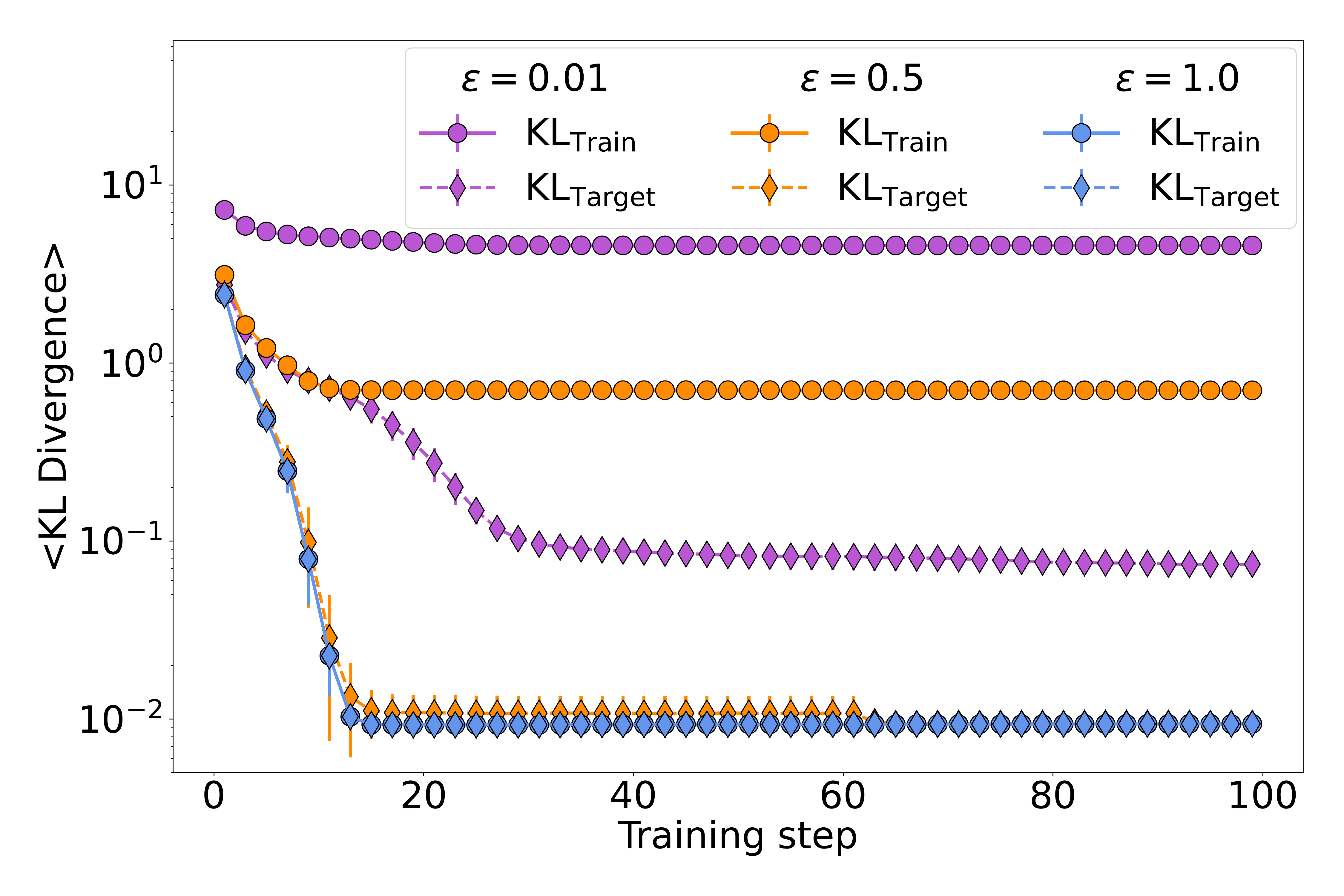}%
} 
\subfloat[\label{stat-coverage}]{%
  \includegraphics[width = 0.5\linewidth]{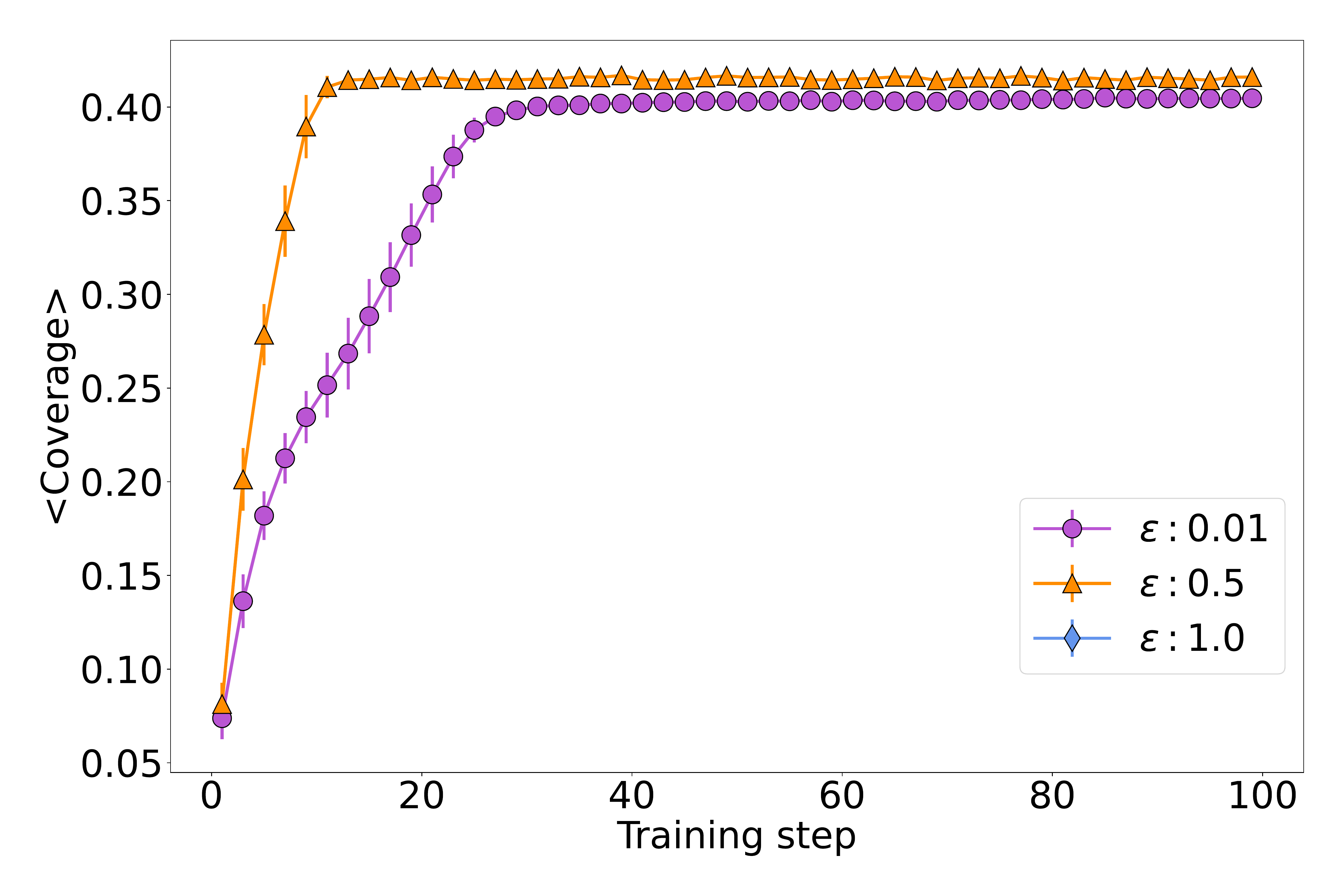}%
}
\\
\subfloat[\label{stat-fidelity}]{%
  \includegraphics[width = 0.5\linewidth]{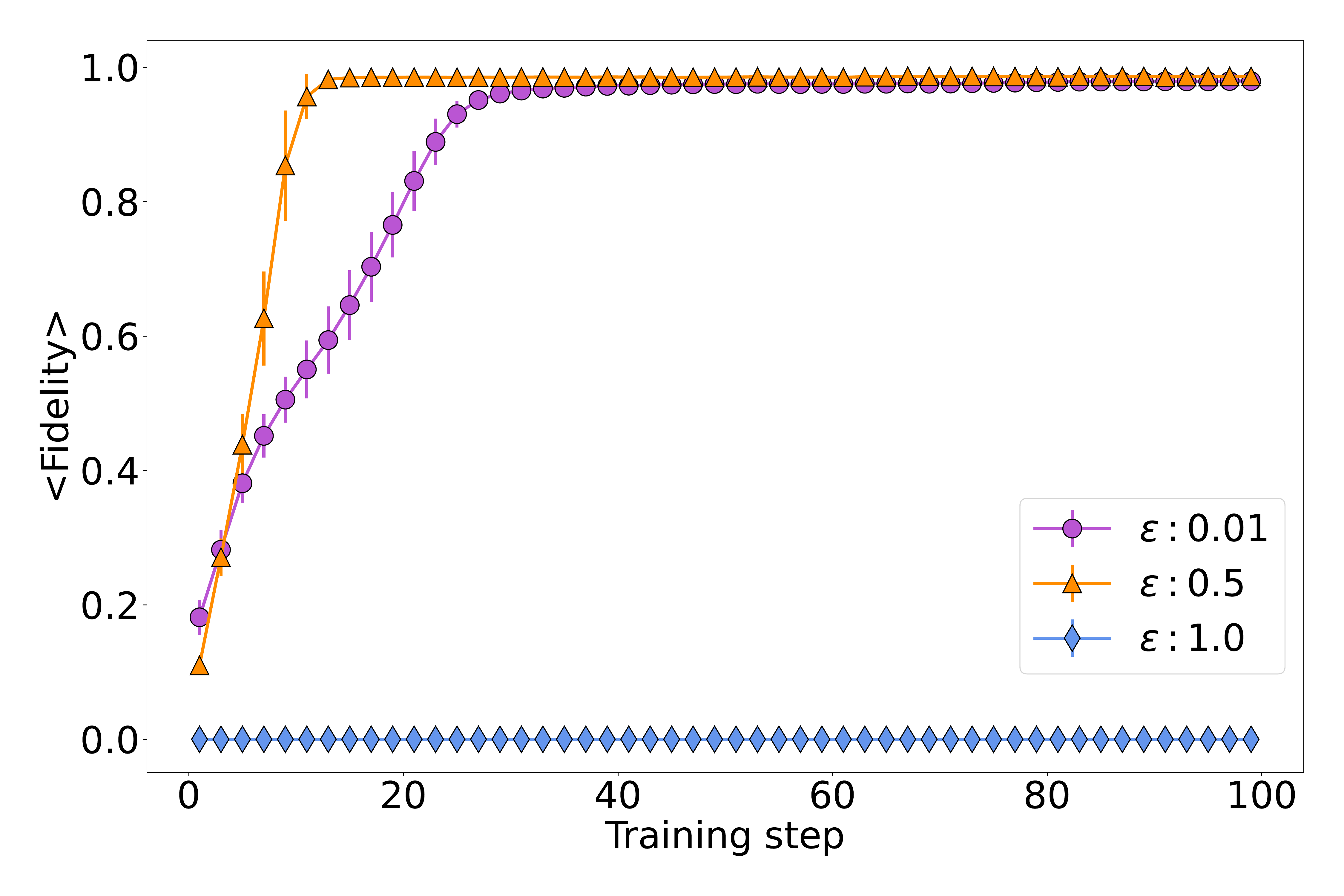}%
}
\subfloat[\label{stat-rate}]{%
  \includegraphics[width = 0.5\linewidth]{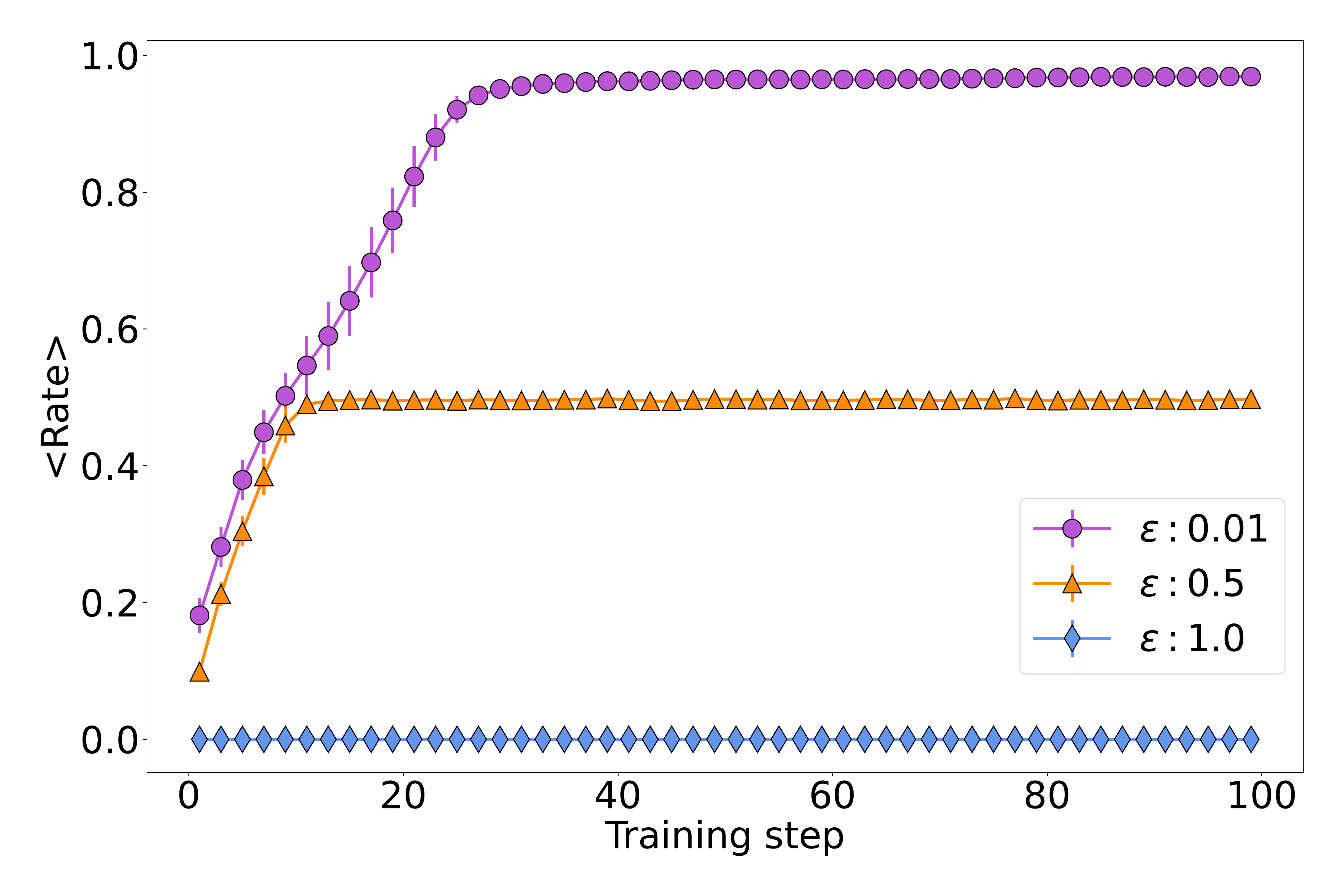}%
}

\caption{\textbf{TNBM's generalization performance throughout training across various $\epsilon$ values.} Here, we show the relationship between the model's ability to learn the ground truth distribution with access to a restricted portion $\epsilon$ of the solution space, and the $(F, R, C)$ values computed throughout the model's training. In panel (a), we see the $KL_\text{Train}$ is always higher than the $KL_\text{Target}$ across $\epsilon$ values - indicating good inference performance. Panels (b), (c), and (d) show that the model's $(F, R, C)$ values increase throughout training (note that coverage is not defined (nan) for $\epsilon=1$). The concurrent relationship between approximating the ground truth and obtaining high $(F, R, C)$ values suggests a positive correlation between our practical models' performance benchmarking approach and computational learning theory.}
\label{f:stat_approach}
\end{figure*}

\end{document}